\documentclass[10pt,journal]{IEEEtran}
\usepackage{amsmath,amsfonts}
\usepackage{array}
\usepackage{textcomp}
\usepackage{stfloats}
\usepackage{url}
\usepackage{verbatim}
\usepackage{graphicx}
\usepackage{cite}

\usepackage{enumerate}
\usepackage{algorithm}
\usepackage[noend]{algpseudocode}
\usepackage{bm}
\usepackage{tabularx}
\usepackage{multirow}
\usepackage{threeparttable}
\usepackage{caption}
\usepackage[font=small]{subfig}
\usepackage{psfrag}
\usepackage{hyperref}
\usepackage{multibib}
\newcites{appx}{References}
\usepackage{color}
\usepackage[dvipsnames]{xcolor}
\usepackage{xr}

\newcommand{\argmin}{\operatornamewithlimits{\mathrm{arg\,min}}}

\newcommand{\fracslant}[2]{
    \raisebox{0.4ex}{\small $#1$}
    \raisebox{0ex}{\large $/$}
    \raisebox{-0.2ex}{\small $#2$}
}

\begin{document}

\title{Accelerating Spherical K-Means Clustering\\
for Large-Scale Sparse Document Data}
\author{Kazuo~Aoyama~and~Kazumi~Saito %
\thanks{Kazuo Aoyama is with NTT Corporation. kazuo.aoyama@ntt.com}%
\thanks{Kazumi Saito is with Kanagawa University. k-saito@kanagawa-u.ac.jp}
}
\maketitle

\begin{abstract}
This paper presents an accelerated spherical K-means clustering algorithm 
for large-scale and high-dimensional sparse document data sets.
We design an algorithm working in an architecture-friendly manner (AFM), 
which is a procedure of suppressing performance-degradation factors 
such as the numbers of instructions, branch mispredictions, and cache misses
in CPUs of a modern computer system.
For the AFM operation, 
we leverage unique universal characteristics (UCs) of a data-object and a cluster's mean set,
which are skewed distributions on data relationships
such as Zipf's law and a feature-value concentration phenomenon.
The UCs indicate that 
the most part of the number of multiplications for similarity calculations is executed 
regarding terms with high document frequencies (df)  
and the most part of a similarity between an object- and a mean-feature vector
is obtained by the multiplications regarding a few high mean-feature values.
Our proposed algorithm applies an inverted-index data structure to a mean set, 
extracts the specific region with high-df terms and high mean-feature values
in the mean-inverted index by newly introduced two structural parameters,
and exploits the index divided into three parts for efficient pruning.
The algorithm determines the two structural parameters by 
minimizing the approximate number of multiplications related to that of instructions,
reduces the branch mispredictions 
by sharing the index structure including the two parameters with all the objects,
and suppressing the cache misses 
by keeping in the caches the frequently used data in the foregoing specific region, 
resulting in working in the AFM.
We experimentally demonstrate that 
our algorithm efficiently achieves superior speed performance in 
large-scale documents
compared with algorithms using the state-of-the-art techniques.
\end{abstract}

\section{Introduction}\label{sec:intro}
Large collections of text documents have been leveraged in a variety of fields.
Understanding their latent structures helps us effectively utilize such document collections.
Clustering is one useful method for revealing these structures.
Among clustering algorithms,
a $K$-means clustering called Lloyd's algorithm \cite{lloyd,macqueen}, 
which is classic yet simple and practical, has been widely used for over 50 years
and has been applied to a part of various algorithms 
such as spectral clustering \cite{ng,vonLuxburg} and 
neural-network compression \cite{wu,cho}.
It partitions an object data set into $K$ distinct clusters 
with a given number $K$ by locally minimizing an objective function
in an iterative greedy manner.
For document-data sets, its specialized variant is known as 
a spherical $K$-means clustering algorithm \cite{dhillon,dhillon_book}.

The spherical $K$-means clustering algorithm consists of two main steps, 
assignment and update, just like Lloyd's algorithm; 
it differs in its given object-feature vectors and a similarity definition.
Feature vector ${\bm x}_i$ $(i\!=\! 1,2,\cdots\!,N)$
is represented as a point on a unit hypersphere, i.e., 
$\|{\bm x}_i\|_2\!=\! 1$.
Mean-feature vector ${\bm \mu}_j^{[r]}$ of the $j$th cluster $(j\!=\! 1,2,\cdots\!,K)$,
which is calculated at the update step in the $r$th iteration, 
is also normalized by its $L_2$-norm, resulting in $\|{\bm \mu}_j^{[r]}\|_2\!=\! 1$.
A similarity between an object- and a mean-feature vector is defined by the cosine similarity, 
which is an inner product in this case.
Note that the $j$th cluster's mean can be referred to as the centroid 
to avoid confusion with a mathematical mean. 

In the spherical $K$-means setting, 
we solve a $K$-means clustering problem 
for large-scale and high-dimensional sparse document-data sets with a huge number of $K$.
In a typical data set, we deal with the following:
number of documents $N\!>\!1\!\times\!10^6$,
number of distinct terms in the data set 
(dimensionality) $D\!>\!1\!\times\!10^5$, 
average number of distinct terms per document $\hat{D}\sim 100$, 
and $K\!\sim\!N/100$.
For fine-grained analyses on a large-scale documents \cite{knittel}, 
a huge number of $K$ is required in the normal course of events.
Hereinafter, 
we respectively call data sets with $(\hat{D}/D)\!\ll 1$ and $(\hat{D}/D)\!\sim\! 1$ 
{\em sparse} and {\em dense}
\footnote{
Although the definitions of {\em sparse} and {\em dense} data are vague,
such language has been used in many studies in a similar manner.
For instance, the indicators $(\hat{D}/D)$ 
in the previous work \cite{huang,schubert} are 
from $2\!\times\!10^{-7}$ to $4.6\!\times\!10^{-3}$
and that in another \cite{li} is around $5\!\times\!10^{-2}$.
}.

Such data sets and their clustering results have interesting 
{\em universal characteristics} (UCs)
of skewed forms regarding some quantities such as Zipf's law (Section~\ref{sec:dchar}).
Among them, 
a {\em feature-value concentration phenomenon} is especially impressive, 
i.e., 
a cluster is annotated by one or a few dominant terms (words) with very large feature values.
Owing to them, a relationship between a similarity and its calculation cost becomes 
like the Pareto principle, i.e., 
a large fraction of the similarity is calculated at low cost (Section~\ref{sec:dchar}).
Furthermore,
a noteworthy phenomenon also appears, 
{\em initial-state independence}, i.e., 
an initial-state selection (seeding) for clustering does not affect algorithm's performance,
especially if a $K$ value is large as in our setting 
(Section~\ref{sec:dchar} and Appendix~\ref{app:init}).
Thus, 
the UCs play an important role in our algorithm design and evaluation as well as 
can be a scientific research subject from a viewpoint of knowledge discovery.

The Lloyd's algorithm has been improved so as to operate 
at high speed and the smallest possible memory size 
\cite{kaukoranta,elkan,hamerly,hamerly_book,drake,ding,rysavy,newling}.
Such {\em accelerations} are, however, for not sparse but dense data sets in a metric space.
Note that 
the term {\em acceleration} means only to speed up an algorithm 
while keeping the same solution as Lloyd's algorithm
if the algorithms start with identical initial states.
Then clustering accuracy is no longer a performance measure. 
The acceleration in the foregoing algorithms originates 
in reducing costly exact distance calculations 
between objects and centroids with pruning methods based on the triangle inequality 
in a metric space.

We briefly explain a typical usage of the triangle inequality
and point out its disadvantage.
Consider skipping a distance calculation between object ${\bm x}_i$ and 
current target centroid ${\bm \mu}_j^{[r]}$
by using a lower bound on the distance.
We can do it if the lower bound is larger than a distance 
between ${\bm x}_i$ and a centroid of the cluster which ${\bm x}_i$ is assigned to.
By applying the triangle inequality to three points of ${\bm x}_i$, ${\bm \mu}_j^{[r]}$,
and the previous centroid ${\bm \mu}_j^{[r-1]}$, the lower bound is calculated as
\begin{equation}
d_{LB}({\bm x}_i, {\bm \mu}_j^{[r]}) = 
|d({\bm x}_i,{\bm \mu}_j^{[r-1]})\!-\!\delta_j^{[r]}|\,,
\label{eq:lb}
\end{equation}
where $\delta_j^{[r]}$ denotes the moving distance between 
${\bm \mu}_j^{[r-1]}$ and ${\bm \mu}_j^{[r]}$ and
$d({\bm a},{\bm b})$ and $d_{LB}({\bm a},{\bm b})$ respectively denote
the distances of ${\bm a}$ to ${\bm b}$ and the lower bound
(detailed in Appendix~\ref{app:triangle}).
The lower bound tightens as the moving distance becomes smaller.
Then more centroids are pruned, causing acceleration.
However, such acceleration algorithms have an essential limit, 
that is, they become effective only around the last stage  
before the convergence where most of the centroids are invariant or slightly move.
Suppose that we prepare an algorithm by replacing the distance with the cosine similarity 
and utilizing a similar pruning strategy for the foregoing accelerations \cite{schubert}.
In our setting, 
it is not efficient as stated above (detailed in Section~\ref{sec:arch}).
For further improvement, it is desired that 
the acceleration goes through all the iterations, 
particularly from the early to the middle stage.

We challenge the efficient acceleration of the spherical $K$-means in our setting, 
under the condition of it operates in a modern computer system with CPUs.
Our idea is to make an algorithm to operate 
in an {\em architecture-friendly manner} (AFM) through all the iterations 
by leveraging the universal characteristics (UCs) 
for {\em data structures} and {\em pruning filters}. 
An algorithm operating in the AFM uses 
as few microcode instructions issued in the CPU as possible,
preventing pipeline hazards that degrade computational performance, 
i.e., suppressing performance-degradation factors 
including the multiplications for similarity calculations\footnote{
The number of multiplications is closely related to 
the number of microcode instructions regarding similarity calculations 
and can be directly monitored in real time.
} 
(Section~\ref{sec:arch}).

Regarding the data structure,
we employ an inverted-index data structure 
as in similarity search algorithms for large-scale documents \cite{zezula}, 
which utilize it for a database \cite{harman,knuth,samet,zobel,buttcher}.
Our inverted index is characterized by its usage and novel structure:
it is applied 
not to an object-feature-vector set but to a mean-feature-vector set
since it effects better performance \cite{aoyama_jdsa} and 
partitioned into three regions with two structural parameters (separators or thresholds) 
for an efficient pruning filter based on the UCs.
Furthermore, 
an object is represented with a tuple (term ID,\,object-feature value) 
called sparse expression, 
where the term IDs are given in ascending order of the document frequency.

We utilize two pruning filters; 
a novel main and a conventional auxiliary one.
The main filter exploits a unique upper bound on a similarity, 
which is calculated 
by utilizing both the {\em summable} property of the inner product (similarity) 
and an {\em estimated shared} (ES) threshold on mean-feature values, 
based on the foregoing data structures (Sections~\ref{sec:prop} and \ref{sec:strparam}).
Besides our upper-bound-based pruning (UBP) filter,
we can consider other UBP filters with the state-of-the-art techniques 
such as 
a threshold algorithm (TA) in the similarity-search field \cite{fagin01,li} 
and blockification using the Cauchy-Schwarz inequality (CS) \cite{bottesch}  
(Section~\ref{ssec:perfcomp}).
Our UBP filter differs from the TA and CS in ways that 
its threshold is {\em estimated} by minimizing an approximate number of multiplications
for similarity calculations and is {\em shared} with all the objects.
For this reason, we call our UBP filter ES.
Our auxiliary filter is called 
an invariant-centroid-based pruning (ICP) filter 
\cite{kaukoranta,bottesch,hattori}.
It becomes effective toward the last stage in the iterations.
It is a sort of accelerations that uses the moving distance (Section~\ref{ssec:icp}).
By combining both the filters,
our algorithm works in the AFM through all the iterations.

In a similar setting to ours, 
an algorithm has been reported which uses a mean-inverted index and 
pruning filters consisting of a UBP filter based on the Cauchy-Schwarz inequality 
and a simpler ICP filter \cite{knittel}.
Our proposed algorithm is evaluated compared with similar algorithms 
with UBP filters based on the TA and CS combined 
with the auxiliary ICP filter in Section~\ref{ssec:perfcomp}.

Note that 
the following strategies are out of our challenge 
while they are useful for speeding up: 
fast heuristics and approximations that don't give the same solution as Lloyd's algorithm 
\cite{broder14,dongmei_zhang}
and initial-state selections (seeding) that may lead to fast convergence 
\cite{arthur,bahmani,min_li,ji} but did not affect the performance in our preliminary experiments
(Section~\ref{sec:dchar} and Appendix~\ref{app:init}).
Since the strategies are orthogonal to our algorithm, 
they don't conflict with ours but rather can be merged into it.

Our contributions are threefold:
\begin{enumerate}
\item 
We propose an accelerated spherical $K$-means clustering algorithm
for large-scale and high-dimensional document-data sets with a huge $K$ value 
(Section~\ref{sec:prop}).
Our proposed algorithm drastically reduces the elapsed time required
at the assignment step in each iteration 
by suppressing the performance-degradation factors including the number of multiplications.
The algorithm is supported by the interrelated 
data structures and pruning filters 
based on the universal characteristics (UCs)  of the data sets (Section~\ref{sec:dchar}),
resulting in its operation in the architecture-friendly manner (AFM) (Section~\ref{sec:arch}).
\item
We develop an algorithm for determining two structural parameters 
that partition the mean-inverted index into three regions (Section~\ref{sec:strparam}).
Based on a newly introduced computational model, 
the algorithm simultaneously finds structural-parameter values at which 
the approximate number of multiplications is minimum.
\item
We experimentally demonstrate that our proposed algorithm 
achieves superior performance on speed and memory consumption
when it is applied to large-scale real document-data sets 
with large $K$ values 
and compare it with other similar algorithms (Section~\ref{sec:exp}).
\end{enumerate}

The remainder of this paper consists of the following eight sections.
Section \ref{sec:arch} describes the design of both the data structure and an algorithm 
operating in the AFM.
Section \ref{sec:dchar} shows 
the UCs of a large-scale high-dimensional sparse document-data set
and its clustering results.
Section \ref{sec:prop} explains our proposed algorithm in detail, and 
Section \ref{sec:strparam} describes how to estimate the two structural parameters 
that are separators of the mean-inverted index.
Section~\ref{sec:exp} shows our experimental settings 
and compares our algorithm's performance with others. 
Section \ref{sec:disc} discusses our proposed clustering algorithm 
based on experimental results, and 
Section \ref{sec:relate} reviews related work 
from viewpoints that clarify the distinct aspects of our work.
The final section provides a conclusion and future work.

For convenience, we list the notation in Table~\ref{table:nota}.
\begin{table}[t]
\small
\centering
\caption{Notation}
\begin{tabular}{|c|p{67mm}|}\hline
Symbol & Description and Definitions
\\ \hline\hline
$N$ & Number of given objects (feature vectors)\\ \hline
$K$ & Number of clusters (means or centroids)\\ \hline
\multirow{2}{*}{$D$} & 
Dimensionality\\
& \hspace*{2mm}Number of distinct terms in an object data set\\ \hline
\hline
\multirow{2}{*}{$\mathcal{X}$} & 
Set of given objects (feature vectors)\\
& \hspace*{2mm}${\mathcal{X}}\!=\!\{\bm{x}_1,\bm{x}_2,\cdots,\bm{x}_N\}$\\
\hline
\multirow{2}{*}{${\mathcal{C}}^{[r]}$} & 
Set of clusters at the $r$th iteration\\
& \hspace*{2mm}${\mathcal{C}}^{[r]}\!=\!\{ C_1^{[r]},\cdots,C_j^{[r]},\cdots,C_K^{[r]} \}$\\
\hline
$a(i)$ & D of the cluster which $\bm{x}_i$ is assigned to\\
\hline
\multirow{3}{*}{${\mathcal{M}}^{[r]}$} & 
Set of means calculated at the $r$th iteration\\
& \hspace*{2mm}${\mathcal{M}}^{[r]}=\{ \bm{\mu}_1^{[r]},\cdots,\bm{\mu}_j^{[r]},\cdots,\bm{\mu}_K^{[r]}\}$\\
& \hspace*{2mm}$\bm{\mu}_{a(i)}^{[r]}$: Mean of $C_{a(i)}^{[r]}$ which $\bm{x}_i$ is assigned to\\
\hline
\multirow{4}{*}{${\bm \rho}$} & 
Set of similarities between an object and centroids\\
& ${\bm \rho}=\{\rho_1,\rho_2,\cdots,\rho_K\},~\rho_{(max)}=\max_{1\leq j\leq K}(\rho_j)$\\
& \hspace*{2mm}$\rho_{a(i)}$: Similarity of $\bm{x}_i$ to $\bm{\mu}_{a(i)}$.\\
& \hspace*{2mm}Object ID is omitted from each symbol $\rho_j$.\\
\hline\hline
\multirow{2}{*}{$\hat{D}$} & 
Average number of distinct terms in sparse object\\
& data set $\hat{\mathcal{X}}$,\hskip.8em $\hat{D}= (1/N)\sum_{i=1}^N (nt)_i$\\
\hline
${\mathcal S}$ &
Set of term IDs, $s\in S$, $|{\mathcal S}|\!=\! D$\\
\hline
\multirow{7}{*}{${\hat{\mathcal{X}}}$} & 
Set of object-tuple arrays of $\hat{\bm{x}}_i$,~$i=1,2,\cdots,N$\\
& \hspace*{2mm}${\hat{\bm x}}_i=[(t_{(i,p)},u_{t_{(i,p)}})]_{p=1}^{(nt)_i}$\\
& \hspace*{4mm}$t_{(i,p)}$: $p$th term ID appeared in $\hat{\bm{x}}_i$\\ 
& \hspace*{12mm}Term ID's are sorted in ascending order\\
& \hspace*{12mm}of document frequency $(df)$\\
& \hspace*{4mm}$u_{t_{(i,p)}}$: $p$th feature value appeared in $\hat{\bm{x}}_i$\\ 
& \hspace*{4mm}$(nt)_i$: Number of distinct terms in $\hat{\bm{x}}_i$\\ 
\hline
\multirow{6}{*}{$\breve{\mathcal{M}}^{[r]}$} & 
Structured inverted index of means, each column\\
& of which is mean-tuple array ${\breve{\bm \xi}}_{s}^{[r]},~s=1,2,\cdots,D$\\
& \hspace*{2mm}${\breve{\bm \xi}}_s^{[r]}=[(c_{(s,q)},v_{c_{(s,q)}})]_{q=1}^{(mf)_s}$\\
& \hspace*{4mm}$c_{(s,q)}$: $q$th mean ID appeared in $\breve{\bm \xi}_s^{[r]}$\\ 
& \hspace*{4mm}$v_{c_{(s,q)}}$: $q$th feature value appeared in $\breve{\bm \xi}_s^{[r]}$\\ 
& \hspace*{4mm}$(mf)_s$: Mean frequency of $s$th term ID\\ 
\hline
\end{tabular}\label{table:nota}
\vspace*{-3mm}
\end{table}

\section{Architechture-Friendly Manner}\label{sec:arch}
We consider running an algorithm on a modern computer system with CPUs.
An architecture-friendly manner (AFM) is a procedure 
that suppresses the following three performance-degradation factors:
(1) the instructions, (2) the conditional branch mispredictions,
and (3) the cache misses. 
These factors impact the performance of algorithms
executed on a modern computer system, 
which contains CPUs and a hierarchical memory system as its main components.
A CPU has plural operating cores each of which has deep pipelines for instruction parallelism 
with superscalar out-of-order execution and multilevel cache hierarchy 
\cite{hennepat,intel_IA64}.
The memory system consists of registers, hierarchical caches 
from level 1 to the last level (e.g., level 3) \cite{intel_IA64}, 
and an external main memory.
To efficiently run an algorithm at high throughput on the system, 
we should not only reduce instructions 
but also avoid pipeline hazards that cause pipeline stalls.
In pipeline hazards, there are two serious threats:
a control hazard induced by branch mispredictions \cite{evers,eyerman} 
and a data hazard that needs access to the main memory 
by last-level-cache misses \cite{hennepat}.
Suppressing the three performance-degradation factors creates a high-performance algorithm.

\begin{algorithm}[t] 
\small
\newcommand{\algto}{\textbf{to}}
\algnewcommand{\LineComment}[1]{\Statex \(\hskip.8em\triangleright\) #1}
  \caption{\hskip.3em Assignment step of MIVI} 
  \label{algo:mivi_assign}
  \begin{algorithmic}[1]
    \Statex{\textbf{Input:}\hskip.8em $\hat{\mathcal{X}}$,~~$\breve{\mathcal{M}}^{[r-1]}$,~~
					$\left\{\rho_{a(i)}^{[r-1]}\right\}_{i=1}^N$,\hskip.3em
					$\forall j;\hskip.3em C_{j}^{[r]}\leftarrow\emptyset$
					}
	\Statex{\textbf{Output:}\hskip.8em $\mathcal{C}^{[r]}\!=\!\left\{ C_j^{[r]} \right\}_{j=1}^K$}

	\ForAll{~$\hat{\bm{x}}_i\!=\![(t_{(i,p)},u_{t_{(i,p)}})]_{p=1}^{(nt)_i}$, 
		\hskip.3em$|\hat{\mathcal X}|\!=\!N$\hskip.3em} \label{algo:mivi_outermost_start}
		\State{$\left\{ \rho_j \right\}_{j=1}^K \gets\! 0$, \hskip.8em 
				$\rho_{(max)}\!\gets\! \rho_{a(i)}^{[r-1]}$}

		\ForAll{~$[t_{(i,p)}]_{p=1}^{(nt)_i}$\hskip.3em} 
			\ForAll{~$[v_{c_{(s,q)}}]_{q=1}^{(mf)_s}\!\in \breve{\xi}_s^{[r-1]}$,\hskip.2em
			$s\!=\!t_{(i,p)}$} 
				\State{$\rho_{c_{(s,q)}}\gets \rho_{c_{(s,q)}}+u_s\cdot v_{c_{(s,q)}}$}
				\label{algo:mivi_outermost_end}
			\EndFor
		\EndFor

		\For{~$j\gets 1$ \algto\hskip.6em $K$\hskip.3em}
			\State{{\bf If}\hskip.3em$\rho_{j} > \rho_{(max)}$\hskip.3em{\bf then}
				\hskip.3em$\rho_{(max)}\leftarrow \rho_{j},\hskip.4em a(i)\leftarrow j$}
		\EndFor

		\State{$C_{a(i)}^{[r]}\leftarrow C_{a(i)}^{[r]}\cup\{\hat{\bm{x}}_i\}$}

	\EndFor
  \end{algorithmic}
\end{algorithm}

We describe the impact of branch mispredictions and last-level-cache misses,
comparing the following three algorithms. 
A baseline algorithm utilizes an inverted-index data structure for a mean (centroid) set
and adopts the term-at-a-time (TAAT) strategy \cite{fontoura} for similarity calculations, 
which is called a mean-inverted-index algorithm (MIVI) in Algorithm~\ref{algo:mivi_assign} 
\cite{aoyama_jdsa}.
The second differs from MIVI only in using an inverted index for a given data-object set,
which is called a data-inverted-index algorithm (DIVI).
The last is an algorithm Ding$^+$ \cite{ding},  
which is originally for dense data sets and works at high speed
using its pruning filters based on the triangle inequality.
To make Ding$^+$ available to a sparse data set in the spherical $K$-means setting,
we modified it without any inverted index as follows. 
A given data-object and a mean set were represented with sparse and full expression,
respectively.
In the full expression, 
all the term IDs were covered, where feature values of undefined term IDs were filled by zeros.
These representations enable us to simply and quickly access a mean-feature value 
by using a data-object term ID as a key for a similarity calculation\footnote{
Mean-feature vectors may also be represented with sparse expression.
However, these representations require solving a set-intersection problem of
finding terms shared by both a data-object- and a mean-feature vector. 
}.
As is obvious, Ding$^+$ used a cosine similarity 
as with an algorithm \cite{schubert} that is modified from Hamerly's algorithm \cite{hamerly}.

\begin{figure}[t]
\begin{center}
	\subfloat[Number of multiplications]{ 
		\hspace*{2mm}
		\psfrag{X}[c][c][0.85]{
			\begin{picture}(0,0)
				\put(0,0){\makebox(0,-2)[c]{Iterations}}
			\end{picture}
		}
		\psfrag{Y}[c][c][0.85]{
			\begin{picture}(0,0)
				\put(0,0){\makebox(0,32)[c]{\# multiplications}}
			\end{picture}
		}
		\psfrag{P}[r][r][0.75]{MIVI}
		\psfrag{Q}[r][r][0.75]{DIVI}
		\psfrag{R}[r][r][0.75]{Ding$^+$}
		\psfrag{A}[c][c][0.75]{$0$}
		\psfrag{B}[c][c][0.75]{$20$}
		\psfrag{C}[c][c][0.75]{$40$}
		\psfrag{D}[c][c][0.75]{$60$}
		\psfrag{H}[r][r][0.75]{$10^9$}
		\psfrag{I}[r][r][0.75]{$10^{10}$}
		\psfrag{J}[r][r][0.75]{$10^{11}$}
		\psfrag{K}[r][r][0.75]{$10^{12}$}
		\psfrag{L}[r][r][0.75]{$10^{13}$}
		\psfrag{M}[r][r][0.75]{$10^{14}$}
		\includegraphics[width=42mm]{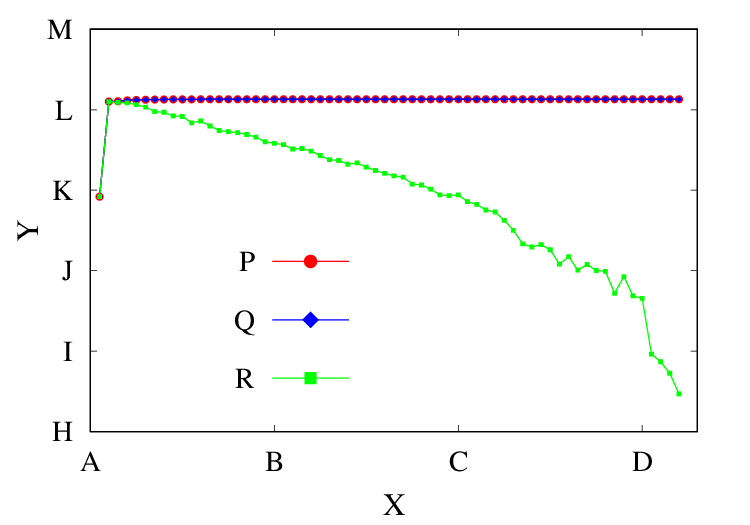}
	}\hspace*{-1mm}
	\subfloat[Elapsed time]{
		\psfrag{X}[c][c][0.85]{
			\begin{picture}(0,0)
				\put(0,0){\makebox(0,-2)[c]{Iterations}}
			\end{picture}
		}
		\psfrag{Y}[c][c][0.75]{
			\begin{picture}(0,0)
				\put(0,0){\makebox(-3,14)[c]{Elapsed time ($\times 10^4$sec)}}
			\end{picture}
		}
		\psfrag{P}[r][r][0.75]{MIVI}
		\psfrag{Q}[r][r][0.75]{DIVI}
		\psfrag{R}[r][r][0.75]{Ding$^+$}
		\psfrag{A}[c][c][0.75]{$0$}
		\psfrag{B}[c][c][0.75]{$20$}
		\psfrag{C}[c][c][0.75]{$40$}
		\psfrag{D}[c][c][0.75]{$60$}
		\psfrag{H}[r][r][0.75]{$0$}
		\psfrag{I}[r][r][0.75]{$1$}
		\psfrag{J}[r][r][0.75]{$2$}
		\psfrag{K}[r][r][0.75]{$3$}
		\psfrag{L}[r][r][0.75]{$4$}
		\psfrag{M}[r][r][0.75]{$5$}
		\includegraphics[width=42mm]{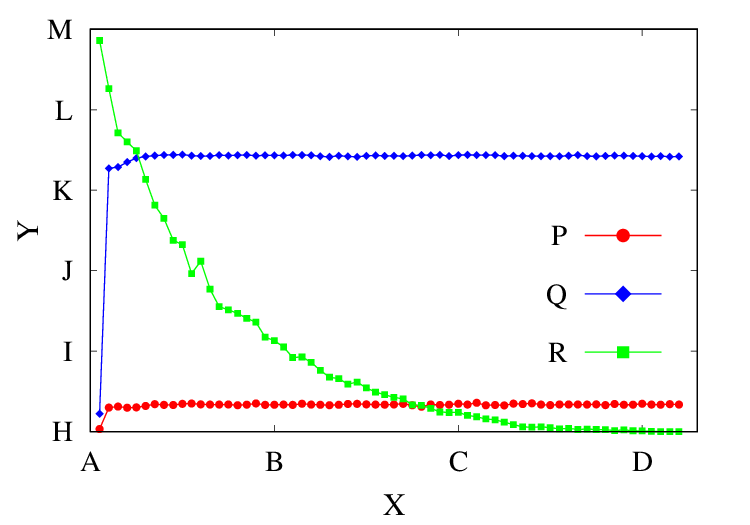}
	}
\end{center}
\vspace*{-2mm}
\caption{\small
Performance comparison of MIVI, DIVI, and Ding$^+$
in 8.2M-sized PubMed data set with $K$=80\,000:
(a) Number of multiplications and (b) Elapsed time 
along iterations.
}
\label{fig:ivfd}
\vspace*{-1mm}
\end{figure}

\begin{table}[t]
\centering
\caption{Performance rates of DIVI and Ding$^+$ to MIVI.}
\label{table:ivfd}
\begin{tabular}{|c|c|c||c|c|c|}\hline
\multirow{2}{*}{Algo} 
& Avg & Avg  &\multirow{2}{*}{Inst}&\multirow{2}{*}{BM}&\multirow{2}{*}{LLCM}\\
& Mult& time &                     &                   & \\ \hline
DIVI     & 1.000 & 10.21 & 0.9824 & 6.981 & 36.48\\ \hline
Ding$^+$ & 0.2284 & 2.892 & 0.6533 & 492.9 & 37.23\\ \hline
\end{tabular}
\vspace*{-2mm}
\end{table}

Figure \ref{fig:ivfd} and Table \ref{table:ivfd} show 
the performance comparisons of MIVI, DIVI, and Ding$^+$ in the 8.2M-sized PubMed data set
(Section \ref{ssec:data}) \cite{pubmed} with $K\!=\! 80\,000$ until the convergence.
The performance rates of DIVI and Ding$^+$ to MIVI are shown in Table \ref{table:ivfd}.
The columns from left to right show the algorithm (Algo), 
the average number of multiplications (Avg Mult), 
the average elapsed time (Avg time), 
the numbers of completed instructions (Inst), branch mispredictions (BM), and 
last-level-cache load misses (LLCM).
The entries in the last three columns were measured with Linux perf tools \cite{perf}.

Both MIVI and DIVI required 
an identical number of multiplications for similarity calculations.
However, their average elapsed times were critically different.
Surprisingly, DIVI needed almost 10 times more average elapsed time than MIVI. 
Ding$^+$ reduced the multiplications by its pruning method,
resulting in almost one-fourth the average number of MIVI's multiplications. 
However, its average elapsed time was about three times larger than MIVI's. 
These results reveal that only the number of multiplications or instructions 
is insufficient for the speed-performance criterion.

Table~\ref{table:ivfd} also shows 
that the numbers of branch mispredictions and cache misses of DIVI and Ding$^+$
were significantly larger than those of MIVI.
The branch misprediction and the cache miss cause many wasted clock-cycle times, 
resulting in slower operation.
This can be explained as follows.
When a target-data-object does not exist in the caches,
a last-level-cache miss occurs and 
the object is loaded from the main memory in the large memory latency.
Such a situation is generally caused by 
the poor temporal and spatial locality in data usage.
DIVI replaces the triple loop of MIVI 
at lines~\ref{algo:mivi_outermost_start} to \ref{algo:mivi_outermost_end}
in Algorithm~\ref{algo:mivi_assign} as follows:
the outermost loop is for mean-feature vectors, 
the middle loop is for the terms in a mean-feature vector 
whose number is much larger than $(nt)_i$ of MIVI, 
and the innermost loop is for object-inverted-index arrays 
whose length is much longer than $(mf)_s$ of $\breve{\xi}_s^{[r-1]}$ in MIVI 
(symbols in Table~\ref{table:nota}).
This replacement loses the locality of target data.
Ding$^+$, in a broad sense, replaces the middle and innermost loop 
with loops for mean-feature vectors and term IDs in $\hat{\bm x}_i$, respectively, 
although it is technically different from MIVI in the loop structure \cite{ding}.
Then, Ding$^+$ accesses a large array of the mean-feature vector in the innermost loop.
The large arrays in the middle and innermost loops diminish the foregoing locality.
Furthermore, 
Ding$^+$ uses many irregular conditional branches in several pruning filters
for reducing the multiplications.
When a conditional branch irregularly changes its judgment result, True or False, 
the branch prediction fails.
Then, a series of instructions executed under the prediction is flushed out 
and correct instructions are executed again. 
This procedure requires more clock-cycle times. 

From these facts, we know that it is important for an algorithm's AFM operation
to avoid using large arrays and irregular conditional branches 
in deeper loops.
Our proposed algorithm employs a mean-inverted index like MIVI and 
gives to the index a specific structure 
that is shared with all the objects to avoid irregular conditional branches.
The structure is designed so as to keep the locality of frequently used data. 
Furthermore, our algorithm exploits a novel pruning filter 
for skipping unnecessary multiplications for similarity calculations 
(Section~\ref{sec:prop}).
These lead to suppress the performance-degradation factors.

\section{Universal Characteristics}\label{sec:dchar}
Large-scale document data sets, which are generally high-dimensional and sparse, 
and their clustering results 
have universal characteristics of skewed forms regarding {\em some quantities}. 
This section describes the characteristics of such a data set
and a mean (centroid) set built by $K$-means clustering,
making connections with both the number of multiplications and 
the fractions of a similarity.
These play an important role in designing our proposed algorithm.

\begin{figure}[t]
\begin{center}\hspace*{1mm}
	\subfloat[{\normalsize Zipf's law}]{ 
		\psfrag{X}[c][c][0.90]{
			\begin{picture}(0,0)
				\put(0,0){\makebox(0,-6)[c]{Rank}}
			\end{picture}
		}
		\psfrag{Y}[c][c][0.90]{
			\begin{picture}(0,0)
				\put(0,0){\makebox(0,20)[c]{Frequency}}
			\end{picture}
		}
		\psfrag{P}[c][c][0.80]{$1$}
		\psfrag{Q}[c][c][0.80]{$10$}
		\psfrag{R}[c][c][0.80]{$10^{2}$}
		\psfrag{S}[c][c][0.80]{$10^{3}$}
		\psfrag{T}[c][c][0.80]{$10^{4}$}
		\psfrag{U}[c][c][0.80]{$10^{5}$}
		\psfrag{A}[r][r][0.80]{$10$}
		\psfrag{B}[r][r][0.80]{$10^{3}$}
		\psfrag{C}[r][r][0.80]{$10^{5}$}
		\psfrag{D}[r][r][0.80]{$10^{7}$}
		\psfrag{J}[r][r][0.84]{{\em tf}}
		\psfrag{K}[r][r][0.84]{{\em df}}
		\includegraphics[width=43mm]{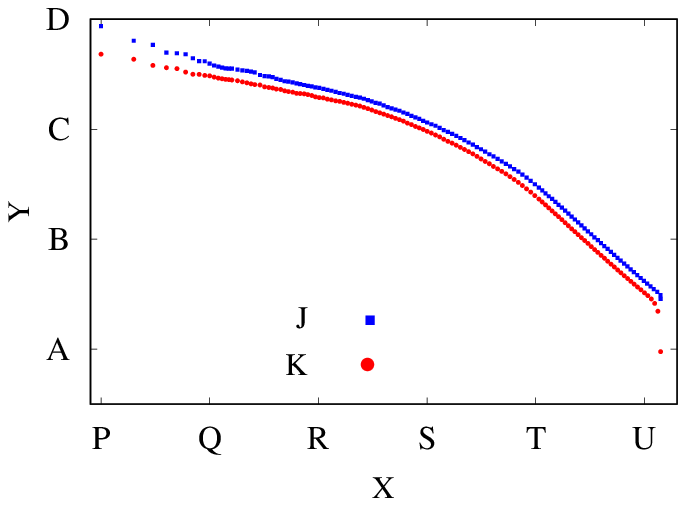}
	}\hspace*{1.0mm}
	\subfloat[{\normalsize Bounded Zipf's law}]{
		\psfrag{X}[c][c][0.90]{
			\begin{picture}(0,0)
				\put(0,0){\makebox(0,-6)[c]{Rank}}
			\end{picture}
		}
		\psfrag{Y}[c][c][0.90]{
			\begin{picture}(0,0)
				\put(0,0){\makebox(0,20)[c]{Frequency}}
			\end{picture}
		}
		\psfrag{P}[c][c][0.80]{$1$}
		\psfrag{Q}[c][c][0.80]{$10$}
		\psfrag{R}[c][c][0.80]{$10^{2}$}
		\psfrag{S}[c][c][0.80]{$10^{3}$}
		\psfrag{T}[c][c][0.80]{$10^{4}$}
		\psfrag{U}[c][c][0.80]{$10^{5}$}
		\psfrag{A}[r][r][0.80]{$10$}
		\psfrag{B}[r][r][0.80]{$10^{3}$}
		\psfrag{C}[r][r][0.80]{$10^{5}$}
		\psfrag{D}[r][r][0.80]{$10^{7}$}
		\psfrag{J}[r][r][0.72]{$K\!=\!8\!\times\!10^{4}$}
		\psfrag{K}[r][r][0.72]{$K\!=\!4\!\times\!10^{4}$}
		\psfrag{L}[r][r][0.72]{$K\!=\!2\!\times\!10^{4}$}
		\psfrag{M}[r][r][0.72]{$K\!=\!1\!\times\!10^{4}$}
		\includegraphics[width=43mm]{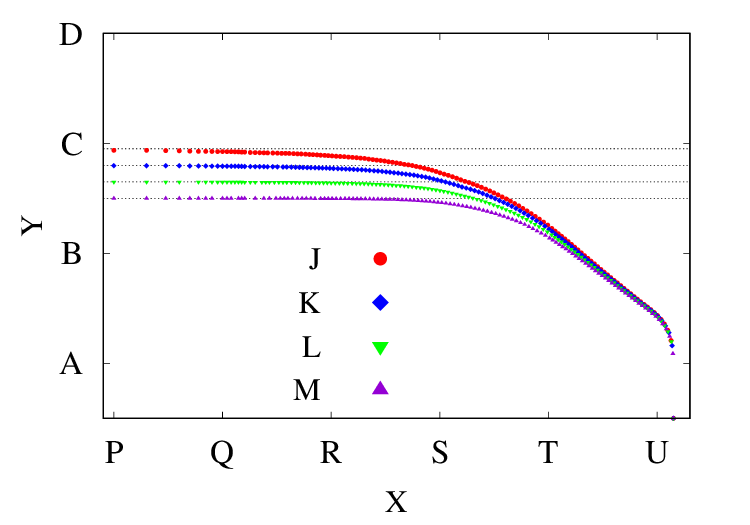}
	}
\end{center}
\vspace*{-2mm}
\caption{\small
Characteristics of 8.2M-sized PubMed data set:
(a) Zipf's law on term frequency ({\em tf}) and document frequency ({\em df}) and  
(b) Bounded Zipf's law on mean frequency ({\em mf}) with four $K$ values.
}
\label{fig:dchar}
\vspace*{-2mm}
\end{figure}

The Zipf's law is known as one of universal characteristics \cite{adamic,newman,clauset}.
It states that a relationship between 
an occurrence frequency of a term (or word) in a text corpus (document data set) 
and the term's rank in descending order of the frequency
is empirically represented in a rank range by 
\begin{equation}
\mbox{{\em Freq}(term)}\propto \mbox{{\em Rank}(term)}^{-\alpha},
\label{eq:zipf}
\end{equation}
where the functions, {\em Freq}(term) and {\em Rank}(term), 
are the frequency and rank of the term, and $\alpha$ is a positive exponent. 
In very large data sets, unlike the foregoing simple Zipf's law, 
it is known that a rank-frequency plot in log-log scale has two parts with different exponents 
\cite{montemurro}.

Figure~\ref{fig:dchar}(a) shows Zipf's law 
of the 8.2M-sized PubMed data set (Section~\ref{ssec:data}), where
{\em tf} and {\em df} respectively denote the total occurrence frequency of each term 
in all of the documents and the number of documents in which each term appeared.
We know that what follows Zipf's law  was not only the term frequency ({\em tf}\,), 
which is usually said, but also the document frequency ({\em df}\,).
Figure~\ref{fig:dchar}(b) shows 
the relationship between the mean frequency ({\em mf}) and {\em Rank}(term),  
where {\em mf} is the number of means (centroids) in which a distinct term appeared.
The log-log plots indicates that 
although the maximum mean frequencies were bounded by the corresponding $K$ values, 
the relationship in the mean set also followed Zipf's law, which we call a bounded Zipf's law.
The relationships in Figs.~\ref{fig:dchar}(a) and (b) are highly right-skewed forms 
expressed by power functions with negative exponents.

\begin{figure}[t]
\begin{center}\hspace*{2mm}
	\subfloat[{\normalsize {\em df}-$\overline{\mbox{\em mf}}$ scatter plot}]{
		\psfrag{X}[c][c][0.90]{
			\begin{picture}(0,0)
				\put(0,0){\makebox(0,-6)[c]{Document frequency}}
			\end{picture}
		}
		\psfrag{Y}[c][c][0.90]{
			\begin{picture}(0,0)
				\put(0,0){\makebox(0,20)[c]{Avg. mean frequency}}
			\end{picture}
		}
		\psfrag{P}[c][c][0.80]{$10$}
		\psfrag{Q}[c][c][0.80]{$10^{2}$}
		\psfrag{R}[c][c][0.80]{$10^{3}$}
		\psfrag{S}[c][c][0.80]{$10^{4}$}
		\psfrag{T}[c][c][0.80]{$10^{5}$}
		\psfrag{U}[c][c][0.80]{$10^{6}$}
		\psfrag{A}[r][r][0.80]{$10$}
		\psfrag{B}[r][r][0.80]{$10^{3}$}
		\psfrag{C}[r][r][0.80]{$10^{5}$}
		\psfrag{K}[r][r][0.80]{$\overline{\mbox{\em mf}}_2\!=\!K$}
		\psfrag{Z}[r][r][0.80]{$\overline{\mbox{\em mf}}_1\!=\!\mbox{\em df}$}
		\includegraphics[width=43mm]{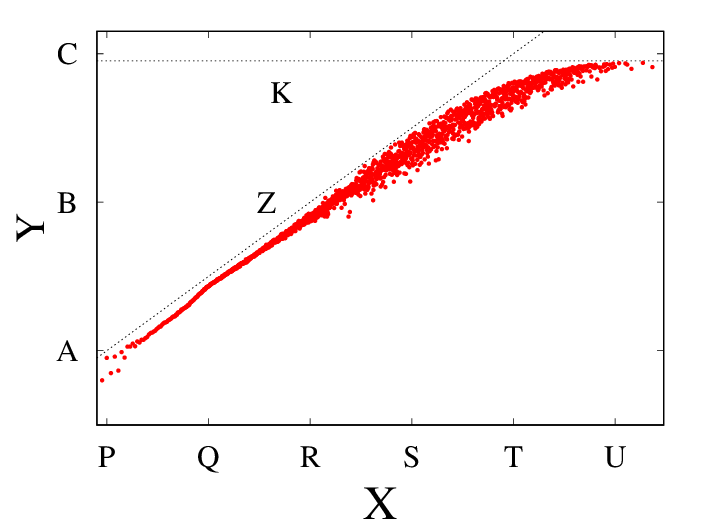}
	}\hspace*{-2.0mm}
	\subfloat[{\normalsize \mbox{\#} multiplications}]{
		\psfrag{K}[c][c][1.0]{$K$}
		\psfrag{N}[c][c][1.0]{$N$}
		\psfrag{D}[c][c][1.0]{$D$}
		\psfrag{M}[c][c][0.9]{{\em mf}}
		\psfrag{F}[c][c][0.9]{{\em df}}
		\psfrag{T}[c][c][0.9]{Term ID}
		\includegraphics[width=43mm]{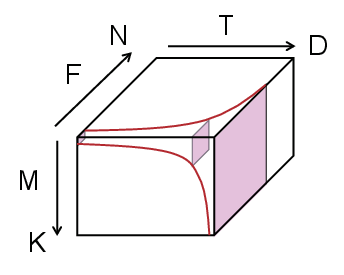}
	}
\end{center}
\vspace*{-1.5mm}
\caption{\small
Characteristics of 8.2M-sized PubMed data set:
(a) {\em df}-$\overline{\mbox{\em mf}}$ scatter plot in log-log scale and 
(b) Diagram of number of multiplications when MIVI is executed, 
which corresponds to volume surrounded by curves inside the rectangle.
}
\label{fig:mfdf}
\vspace*{-2mm}
\end{figure}

Figure~\ref{fig:mfdf}(a) shows 
a positive correlation between {\em df}\, and $\overline{\mbox{\em mf}}$\,
in the scatter plot in log-log scale when $K\!=\!8\!\times\! 10^4$ 
in Fig.~\ref{fig:dchar},  
where $\overline{\mbox{\em mf}}$ is the average mean frequency of terms having 
an identical document frequency, 
which is expressed by 
\begin{equation}
\overline{\mbox{\em mf}}=(1/|{\cal S}_{df}|) 
\textstyle{\sum_{s\in {\cal S}_{df}} \mbox{{\em mf}}_s},
\hspace{3mm}
{\cal S}_{df}=\{ s\,|\,\mbox{\em df}_s = \mbox{\em df}\,\}\, .
\label{eq:avg_mf}
\end{equation}
The slope of the diagonal straight line denotes 
$\overline{\mbox{\em mf}}_1\!=\!\mbox{\em df}$
and the horizontal line $\overline{\mbox{\em mf}}_2\!=\!K$.
This indicates that a high-{\em df}\, term has a high {\em mf}\,.
Figure~\ref{fig:mfdf}(b) shows a diagram representing the number of multiplications 
for similarity calculations 
($\sum_{s=1}^D \mbox{\em mf}_s\cdot \mbox{\em df}_s$) 
when MIVI in Section~\ref{sec:arch} is applied to the data in our setting.
The horizontal axis depicts the term ID given in ascending order of {\em df},
which is the opposite direction of the terms in Fig.~\ref{fig:dchar}(a),
the depth axis represents {\em df}, and the vertical axis {\em mf}. 
The volume surrounded by curves inside the rectangle corresponds to 
the number of multiplications.
This number is {\em quite unevenly distributed} in the large term-ID range, 
i.e., in the high-{\em df} region.

\begin{figure}
\begin{center}
	\subfloat[{\normalsize Skewed form}]{
		\psfrag{X}[c][c][0.86]{
			\begin{picture}(0,0)
				\put(0,0){\makebox(0,-4)[c]{Rank/$K$}}
			\end{picture}
		}
		\psfrag{Y}[c][c][0.86]{
			\begin{picture}(0,0)
				\put(0,0){\makebox(0,20)[c]{Mean-feature value}}
			\end{picture}
		}
		\psfrag{P}[c][c][0.80]{$10^{-4}$}
		\psfrag{Q}[c][c][0.80]{$10^{-2}$}
		\psfrag{R}[c][c][0.80]{$1$}
		\psfrag{S}[c][c][0.80]{$10^{2}$}
		\psfrag{T}[c][c][0.80]{$10^{4}$}
		\psfrag{A}[r][r][0.80]{$0$}
		\psfrag{B}[r][r][0.80]{$0.2$}
		\psfrag{C}[r][r][0.80]{$0.4$}
		\psfrag{D}[r][r][0.80]{$0.6$}
		\psfrag{E}[r][r][0.80]{$0.8$}
		\psfrag{F}[r][r][0.80]{$1.0$}
		\psfrag{W}[l][l][0.72]{$1/\sqrt{2}$}
		\psfrag{Z}[c][c][0.78]{$K\,(\times\!10^{4})$}
		\psfrag{J}[r][r][0.78]{$1$}
		\psfrag{K}[r][r][0.78]{$2$}
		\psfrag{L}[r][r][0.78]{$4$}
		\psfrag{M}[r][r][0.78]{$8$}
		\includegraphics[width=42mm]{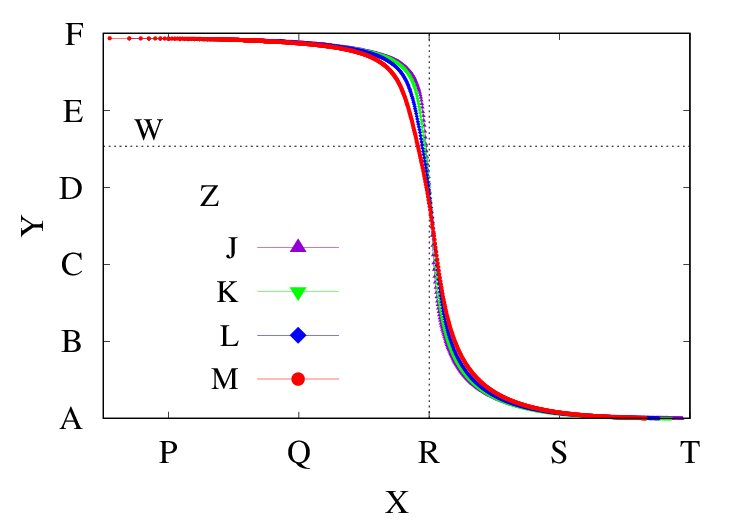}
	}
	\subfloat[{\normalsize Pareto principle}]{
		\psfrag{X}[c][c][0.86]{
			\begin{picture}(0,0)
				\put(0,0){\makebox(0,-4)[c]{Normalized rank}}
			\end{picture}
		}
		\psfrag{Y}[c][c][0.82]{
			\begin{picture}(0,0)
				\put(0,0){\makebox(0,22)[c]{CPS}}
			\end{picture}
		}
		\psfrag{A}[c][c][0.78]{$0$} \psfrag{B}[c][c][0.78]{$0.2$}
		\psfrag{C}[c][c][0.78]{$0.4$} \psfrag{D}[c][c][0.78]{$0.6$}
		\psfrag{E}[c][c][0.78]{$0.8$} \psfrag{F}[c][c][0.78]{$1$}
		\psfrag{G}[r][r][0.78]{$0$} \psfrag{H}[r][r][0.78]{$0.2$}
		\psfrag{I}[r][r][0.78]{$0.4$} \psfrag{J}[r][r][0.78]{$0.6$}
		\psfrag{K}[r][r][0.78]{$0.8$} \psfrag{L}[r][r][0.78]{$1$}
		\psfrag{U}[l][l][0.77]{(0.10,\,0.92)}
		\psfrag{V}[r][r][0.76]{2nd iteration}
		\psfrag{W}[r][r][0.76]{Convergence}
		\includegraphics[width=42mm]{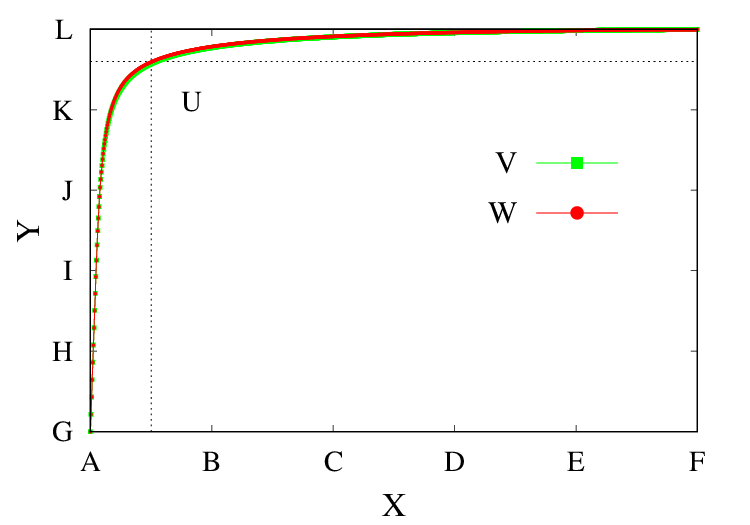}
	}
\end{center}
\vspace*{-2mm}
\caption{\small
(a) Skewed form of mean-feature values 
and (b) cumulative partial similarity (CPS) against 
normalized rank.
Both were built from 8.2M-sized PubMed data set and $K$=80\,000 was used for (b).
}
\label{fig:rankmean}
\vspace*{-3mm}
\end{figure}

Next, we describe skewed forms relating with the fractions of a similarity.
We observed that most of clusters' means had one or a few dominant terms 
with very large feature values, which is named 
a {\em feature-value concentration phenomenon} 
(related to Fig.~\ref{fig:order_mfv}).
Figure~\ref{fig:rankmean}(a) shows 
a skewed form on the feature values in all the centroids (bag-of-features), 
where each feature is expressed by {\em tf-idf} in Section~\ref{ssec:data}.
All the non-zero elements were sorted in descending order of their values, 
and the values were depicted along the rank normalized by $K$.
Qualitatively, there exist very large feature values, 
and many feature vectors have them.
Since no feature vector has plural elements whose values are larger than $(1/\sqrt{2})$,
the rank at this value corresponds to the number of 
centroids with the larger feature values.
This shows the feature-value concentration phenomenon 
that a very small number of features occupied a large part of feature values 
in each centroid.
Here, consider constructing a mean-inverted index whose array is sorted by its feature values
in descending order.
Then, another skewed form appears in the inverted-index array,
where the top to a few entries have very large feature values (Fig.~\ref{fig:order_mfv}).

The foregoing characteristics create a Pareto-principle-like phenomenon 
on a relationship between a cumulative partial similarity ({\em CPS}) and a normalized rank ({\em NR})
of the partial similarity, 
where the partial similarity is the product of an object- and a mean-feature value on a term
and the {\em NR} is the partial-similarity rank 
in descending order that is normalized by the corresponding similarity 
(detailed in Appendix~\ref{app:pareto}).
Figure~\ref{fig:rankmean}(b) shows the relationship 
in the 8.2M-sized PubMed with $K\!=\!80\,000$.
Only the $10\%$ partial-similarity calculations led to the $92\%$ {\em CPS}.
This phenomenon indicates that a large fraction of similarity is obtained by a few multiplications.
Based on this, we can make a tight upper bound on the similarity efficiently.

We call the skewed forms, which are represented as (1) Zipf's law, (2) bounded Zipf's law, 
(3) feature-value concentration phenomenon, and
(4) Pareto-principle-like phenomenon,
the universal characteristics of large-scale sparse data sets. 
We identify the special region in a mean-inverted index
by utilizing newly introduced two structural parameters 
in Section~\ref{sec:strparam} and structure the index.
By using the structured mean-inverted index, 
we can exploit the UCs to reduce the multiplications for similarity calculations
in Section~\ref{sec:prop}.

\section{Proposed Algorithm: ES-ICP}\label{sec:prop}
Our proposed algorithm ES-ICP prunes centroids 
that cannot be the most similar one
to reduce the multiplications for similarity calculations 
(more exactly, the multiply-add operations).
ES-ICP utilizes both a main and an auxiliary filter.
The former is a unique upper-bound-based pruning (UBP) filter named ES.
Our ES filter utilizes a tight upper bound obtained with low computational cost
and safely narrows down the target centroids for similarity calculations 
through all the iterations until convergence.
The auxiliary filter is an invariant-centroid-based pruning (ICP) filter
\cite{kaukoranta,lai,bottesch,hattori} 
that effectively works toward the end of the iterations,
which is a good match with generic algorithms using an inverted index.
This section explains our algorithm with both the filters 
from the viewpoint of how it exploits the universal characteristics 
(UCs) in Section~\ref{sec:dchar} 
and works in the architecture-friendly manner (AFM) in Section~\ref{sec:arch}.

\subsection{ES Filter Design}\label{ssec:frame}
\begin{table}[t]
\small
\centering
\caption{Notation in ES-ICP algorithm}
\label{table:nota_prop}
\begin{tabular}{|c|p{65mm}|}\hline
Symbol & Description and Definitions\\
\hline\hline
$t^{[th]}$ & Parameter: Threshold on term ID\\
\hline
$v^{[th]}$ & Parameter: Threshold on mean-feature-value\\
\hline
\multirow{2}{*}{$(mfH)_s$} 
& Mean frequency of $s$th term ID where\\
& $v_{(s,q)}\geq v_h^{[th]}$: \hskip.8em $(mfL)_s = (mf)_s -(mfH)_s$\\
\hline
\multirow{4}{*}{$y_{(i,j)}$} 
& Partial $L_1$-norm of object-feature-vectors whose\\
& tuple $(t_{(i,p)},u_{t_{(i,p)}})$ satisfies $t_{(i,p)}\!\geq t^{[th]}$ and that\\
& vary depending on the $j$th centroid's feature value\\
& $1\leq p\leq (nt)_i$, \hskip.5em $1\leq i\leq N$,\hskip.5em and \hskip.5em$1\leq j\leq K$\\
\hline
\multirow{2}{*}{$\mathcal{Z}_i$} 
& Set of candidate-mean ID's which is used at\\
& verification phase\\
\hline
\multirow{4}{*}{$\breve{\mathcal{M}}^{p[r]}$} 
& Inverted index of partial mean-feature-vectors,\\
& each column of which is a value array with full\\
& expression, $1\!\leq\! j\!\leq\! K$, denoted by $\breve{\bm \zeta}_s^{[r]}$,\\ 
& $t^{[th]}\!\leq\! s\!\leq\! D$\\
\multirow{2}{*}{$w_{(j,s)}$}
& Value $w_{(j,s)}\!\in\breve{\bm \zeta}_s^{[r]}$ is $v_{c_{(s,q)}}$ for $c_{s,q)}\!=\!j$ 
	if defined\\
& and $v_j\!<\!v^{[th]}$,~0 otherwise.\\
\hline\hline
\multirow{2}{*}{$xState$} 
& Boolean flag for each object: 1 if it satisfies\\ 
& the condition in Eq.~(\ref{eq:objstate}), otherwise 0.\\
\hline
$(nMv)$
& {Number of moving centroids}\\ 
\hline
\multirow{3}{*}{$(mfM)_s$} 
& Mean frequency of moving centroids (means)\\
& In Region 2, only centroids satisfying\\
& $v_{c_{(s,q)}}\!\geq\! v^{[th]}$ are counted.\\
\hline
\end{tabular}
\vspace*{-3mm}
\end{table}

\begin{figure}[t]
\begin{center}
	\psfrag{E}[c][c][.9]{1}
	\psfrag{M}[c][c][.9]{{\em mf}}
	\psfrag{U}[c][c][.88]{$(q)$}
	\psfrag{K}[l][l][.82]{$(mf)_s\!=\!(mfH)_s\!+\!(mfL)_s\leq K$}
	\psfrag{C}[r][r][.9]{$1$}
	\psfrag{T}[c][c][.9]{Term ID\,: $s$}
	\psfrag{D}[l][l][.9]{$D$}
	\psfrag{A}[c][c][.9]{$t^{[th]}$}
	\psfrag{B}[l][l][.9]{$v^{[th]}$}
	\psfrag{F}[r][r][.8]{$v^{[th]}$}
	\psfrag{H}[c][c][.9]{Inverted-index array}
	\psfrag{I}[c][c][.9]{$\breve{\xi}_s$ at $s$th term ID}
	\psfrag{L}[l][l][.82]{$(mfH)_s$}
	\psfrag{N}[l][l][.82]{$(mfL)_s$}
	\psfrag{a}[c][c][.80]{$c_{(s,1)},v_{c_{(s,1)}}$}
	\psfrag{b}[c][c][.80]{$c_{(s,2)},v_{c_{(s,2)}}$}
	\psfrag{c}[c][c][.80]{$c_{(s,3)},v_{c_{(s,3)}}$}
	\psfrag{d}[c][c][.80]{$c_{(s,q)},v_{c_{(s,q)}}$}
	\psfrag{e}[c][c][.78]{$c_{(s,(mf)_s)}$,}
	\psfrag{f}[c][c][.80]{$v_{c_{(s,(mf)_s)}}$}
	\psfrag{P}[r][r][.90]{Region~1}
	\psfrag{Q}[r][r][.90]{Region~2}
	\psfrag{R}[r][r][.90]{Region~3}
	\includegraphics[width=80mm]{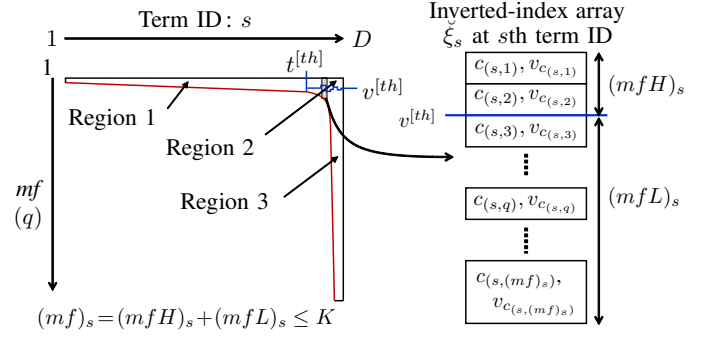}
\end{center}
\caption{\small
Diagram of three regions illustrated on plane of term ID 
and mean-inverted index. Right figure represents $s$th 
mean-inverted-index array composed of tuples $\bm{(c_{(s,q)},v_{c_{(s,q)}})}$.}
\label{fig:regions}
\vspace*{-2mm}
\end{figure}

Our ES filter does not use either the triangle inequality in a metric space or its variants;
it uses the summable property (or additivity) in the inner product of the cosine similarity.
To construct the ES filter, 
we first provide structures to a given data-object and a mean set.
In the data-object set, 
the term IDs ($1\!\leq\! s\!\leq\! D$) are sorted in ascending order 
of the term's document frequency ({\em df}\,).
An inverted-index data structure is applied to the mean set (Section~\ref{sec:arch})
and each array $\breve{\xi}_s$ in the mean-inverted index $\breve{\mathcal{M}}$ 
is aligned along term-ID $s$ (Fig.~\ref{fig:mfdf}(b)).
Then, from the correlation between {\em df}\, and {\em mf}\, in Fig.~\ref{fig:mfdf}(a),
frequently used mean-feature values collect in the arrays with the large term IDs.
To identify the arrays,
we introduce a structural parameter of $t^{[th]}$ on term IDs.
Furthermore, to exploiting the feature-value concentration phenomenon,
we incorporate another structural parameter of $v^{[th]}$ on mean-feature values.
The two structural parameters 
partition $\breve{\mathcal{M}}$ into the following three regions: 
\begin{enumerate}[\textrm{[Region}~1\textrm{]}]
\item $1\leq s < t^{[th]}$ with respect to term ID $(s)$
\item $t^{[th]}\!\leq\! s\!\leq\! D$ and $v_{c_{(s,q)}}\!\geq\!v^{[th]}$, 
where 
$v_{c_{(s,q)}}$ is the mean-feature value of the cluster ID of 
$c_{(s,q)}$\footnote{
$v_{c_{(s,q)}}$ is not sorted but only classified by a comparison to $v^{[th]}$.}
and $q$ is the local order ($q\!=\! 1,\!\cdots,\! (mfH)_s$) on $\breve{\xi}_s$ 
(symbols in Tables~\ref{table:nota} and \ref{table:nota_prop})
\item $t^{[th]}\leq s\leq D$ and $v_{c_{(s,q)}} < v^{[th]}$\, .
\end{enumerate}
By setting $t^{[th]}$ and $v^{[th]}$ at appropriate values 
in Section~\ref{sec:strparam}, 
we make a tight upper bound on a similarity of $\hat{\bm x}_i$ to mean-feature vectors,
which realizes the Pareto-principle-like phenomenon 
in Sections~\ref{sec:dchar} and \ref{sec:disc}.
Figure~\ref{fig:regions} shows a diagram of 
the mean-inverted index $\breve{\mathcal{M}}$ partitioned into the three regions 
and its array $\breve{\xi}_s$ in Regions 2 and 3\footnote{
The entries in Region 3 are not used. Instead, a partial mean-inverted index 
$\breve{\mathcal M}^{p[r]}$ in Table~\ref{table:nota_prop} is used 
for calculating an exact partial similarity.
}.

Next, we define a tight upper bound based on the data structure and 
the additivity of the inner product. 
Given the $i$th object, 
we calculate the exact partial similarities to the $j$th centroid;
$(\rho 1)_{(j;i)}$ and $(\rho 2)_{(j;i)}$ in Regions 1 and 2, respectively.
The upper-bound $(\rho 3)_{(j;i)}^{[ub]}$ in Region 3 is estimated 
as $\{ y_{(i,j)}\!\cdot\!v^{[th]}\}$ 
\footnote{
In our implementation, $\{ y_{(i,j)}\!\cdot\!v^{[th]} \}$ is not performed
for reducing the multiplications (Appendix~\ref{app:fullalgo}).
Instead, by $v^{[th]}$, 
object-feature values are multiplied and mean-feature values are divided, 
just after $t^{[th]}$ and $v^{[th]}$ are determined.
Then, we can simply use the scaled $\{ y_{(i,j)}\}$ keeping their other products identical.
This scaling becomes possible since $v^{[th]}$ is shared with all the objects.
\label{footnote:scaling}
},
where $y_{(i,j)}$ denotes the remaining $L_1$-norm of $\hat{\bm x}_i$
which is not used in Regions 1 and 2. 
The upper bound on similarity $\rho_{(j;i)}^{[ub]}$ is defined by 
\begin{equation}
\rho_{(j;i)}^{[ub]} = (\rho 1)_{(j;i)} + (\rho 2)_{(j;i)} + (\rho 3)_{(j;i)}^{[ub]} \, .
\label{eq:ub}
\end{equation}
Collecting frequently used mean-feature values in Region 2 leads to reducing cache misses.
Limiting the inexact part to the partial similarity in Region 3 
makes the upper bound tight and reduces the multiplications, i.e., the instructions.
Furthermore, sharing $t^{[th]}$ and $v^{[th]}$ in all the objects omits 
unnecessary conditional branches, resulting in reducing branch mispredictions.
At the verification phase,
only for the $j$th centroid passing through the ES filter with low probability, 
the exact partial similarity $(\rho 3)_{(j;i)}$ in Region 3 is calculated,
using additional partial mean-inverted index $\breve{\mathcal M}^{p[r]}$ 
in Table~\ref{table:nota_prop},
whose memory size of $\{ K(D -t^{[th]}\!+\!1)(\mbox{sizeof(double)}) \}$-byte
is not so large because $t^{[th]}$ is close to $D$.
Then, $(\rho_3)_{(j;i)}$ is added to $(\rho 1)_{(j;i)}\!+\!(\rho 2)_{(j;i)}$ 
for exact similarity.

Thus, we design the ES filter with a tight upper bound on the similarity, 
exploiting both the UCs and 
the summable property of the inner product 
and leveraging the data structures for the algorithm's operation in the AFM.

\subsection{ES-ICP Algorithm}\label{ssec:icp}
\begin{figure}[t]
\begin{center}
	\psfrag{H}[c][c][.90]{Inverted-index array}
	\psfrag{I}[c][c][.90]{in Region 1}
	\psfrag{J}[c][c][.90]{in Region 2}
	\psfrag{Q}[c][c][.90]{$q$}
	\psfrag{A}[c][c][.90]{$1$}
	\psfrag{B}[c][c][.90]{$(mf)_s$}
	\psfrag{C}[c][c][.90]{$(mfH)_s$}
	\psfrag{M}[l][l][.90]{$(mfM)_s$}
	\psfrag{V}[l][l][.85]{$v_{c_{(s,q)}}\!\geq\!v^{[th]}$}

	\psfrag{L}[l][l][.90]{Moving}
	\psfrag{P}[l][l][.90]{centroids}
	\psfrag{N}[l][l][.90]{Invariant}
	\psfrag{R}[l][l][.90]{centroids}

	\psfrag{a}[c][c][.81]{$c_{(s,1)},v_{c_{(s,1)}}$}
	\psfrag{b}[c][c][.81]{$c_{(s,2)},v_{c_{(s,2)}}$}
	\psfrag{c}[c][c][.81]{$c_{(s,3)},v_{c_{(s,3)}}$}
	\psfrag{d}[c][c][.81]{$c_{(s,4)},v_{c_{(s,4)}}$}
	\psfrag{e}[c][c][.81]{$c_{(s,(mf)_s)}$,}
	\psfrag{f}[c][c][.81]{$v_{c_{(s,(mf)_s)}}$}
	\psfrag{g}[c][c][.81]{$c_{(s,(mfH)_s)}$,}
	\psfrag{h}[c][c][.81]{$v_{c_{(s,(mfH)_s)}}$}
\hspace*{-4mm}\includegraphics[width=73mm]{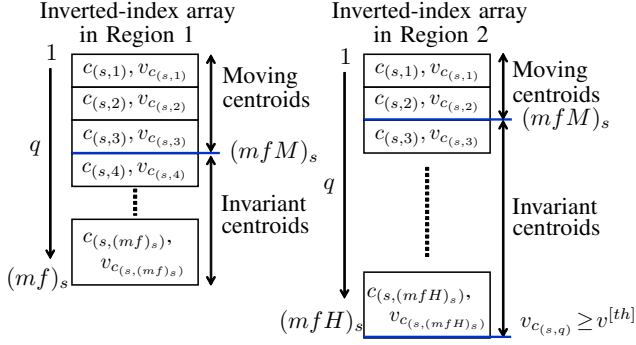}
\end{center}
\vspace*{-2mm}
\caption{\small
Structures of mean-inverted-index arrays: 
Left and right figures represent $s$th arrays in Regions 1 and 2. 
}
\label{fig:icp}
\end{figure}

Our ES-ICP algorithm simultaneously utilizes the ES and ICP filters.
The ICP filter omits the similarity calculations between the {\em more similar object} and 
the invariant centroids.
The $i$th object is {\em more similar} if its similarity to the centroid whose ID is $a(i)$ 
satisfies 
\begin{equation}
\rho_{(a(i);i)}^{[r-1]} \geq \rho_{(a(i);i)}^{[r-2]}\, .
\label{eq:objstate}
\end{equation}

To make the ICP filter operate in the AFM, 
we furthermore provide another structure for the mean-inverted index $\breve{\mathcal M}^{[r]}$ 
by dividing a mean-inverted-index array into two blocks: 
the first and second blocks consist of moving- and invariant-centroid's tuples, respectively.
Figure~\ref{fig:icp} shows mean-inverted-index arrays $\breve{\xi}_s$ with 
two blocks in Regions 1 and 2 on the left and right sides,
where $(mfM)_s$ denotes the number of moving centroids at the $s$th term ID.
The arrays in Region 2, in particular, have a double structure due to 
the two conditions of $v_{c_{(s,q)}}\!\geq\! v^{[th]}$ and that in Eq.~(\ref{eq:objstate}).
If the $i$th  object satisfies the condition of Eq.~(\ref{eq:objstate}),
we can limit the similarity calculations to only the moving centroids, 
which are the means of active clusters, 
$C_{j\neq a(i)}^{[r-1]}\neq C_{j\neq a(i)}^{[r-2]}$
and only in the first block 
identified by the top to the $(mfM)_s$ entries in $\breve{\xi}_s$.

\begin{algorithm}[t] 
\small
\newcommand{\mf}{\mbox{($mf$\hskip0.1em)}}
\newcommand{\mfH}{\mbox{($mfH$\hskip0.1em)}}
\newcommand{\algto}{\textbf{to}}
\algnewcommand{\LineComment}[1]{\Statex \(\hskip.8em\triangleright\) #1}

  \caption{\hskip.8em Assignment step in ES-ICP algorithm} 
  \label{algo:full}
  \begin{algorithmic}[1]
    \Statex{\textbf{Input:}\hskip.8em $\hat{\mathcal{X}}$,~~$\breve{\mathcal{M}}^{[r-1]}$,~~
					$\breve{\mathcal{M}}^{p[r-1]}$,~~$\left\{\rho_{a(i)}^{[r-1]}\right\}_{i=1}^N$,
					~~$t^{[th]},~v^{[th]}$}
	\Statex{\textbf{Output:}\hskip.8em $\mathcal{C}^{[r]}\!=\!\left\{ C_j^{[r]} \right\}_{j=1}^K$}

	\LineComment{Calculate similarities in parallel w.\!r.\!t $\hat{\mathcal{X}}$}
	\State{$C_{j}^{[r]}\leftarrow\emptyset$~,~~$j=1,2,\cdots,K$~}
	\ForAll{~$\hat{\bm{x}}_i\!=\![(t_{(i,p)},u_{t_{(i,p)}})]_{p=1}^{(nt)_i}\in \hat{\mathcal{X}}$~~}
	\label{full:out_start}
		\State{$\left\{ \rho_j \right\}_{j=1}^K \gets\! 0$,~
				\hskip.5em~$\rho_{(max)}\!\gets\! \rho_{a(i)}^{[r-1]}$,~
				\hskip.5em~$\mathcal{Z}_i\!\gets\!\emptyset$
		} \label{full:init_start}

		\For{$j\gets 1$ \algto\hskip.6em $K$\hskip.3em}
			\State{$y_{(i,j)}\gets \sum_{t_{(i,p)}\geq t^{[th]}}\hskip.3em (u_{t_{(i,p)}})$}
			\label{full:init_end}
			\Comment{Initializing $y_{(i,j)}$}
		\EndFor

		\LineComment{Gathering phase}
		\If{$xState = 1$ \label{full_icp_switch:g1_start}}
		\Comment{ICP filtering}
			\State{$\left( \mathcal{Z}_i,\hskip.5em \{\rho_j\}_{j=1}^K \right) = 
				G_1 (\mbox{args}_1)$} 
				\label{full_icp_switch:g1}
				\Comment{Algorithm~\ref{algo:icp_gather1}}
			\State{\hskip1.0em args$_1\supset\{\mathcal{Z}_i,\{(\rho_j,\,y_{(i,j)}\}_{j=1}^K,\,\rho_{(max)},(nMv))$\}}
		\Else \label{full_icp_switch:g0_start}
			\State{$\left( \mathcal{Z}_i,\hskip.5em \{\rho_j\}_{j=1}^K \right) = 
				G_0 (\mbox{args}_0)$}
			\State{\hskip1.0em args$_0\supset\{\mathcal{Z}_i,\{(\rho_j,\,y_{(i,j)}\}_{j=1}^K,\,\rho_{(max)}\}$
				\label{full_icp_switch:g0_end}
			}
		\EndIf

		\LineComment{Verification phase}
		\For{$(s\gets t_{(i,p)}) \geq t^{[th]}$} \label{full:r3_start} 
		\Comment{Region 3}
			\State{{\bf for}\hskip.3em~{\bf all}\hskip.3em~$j\in \mathcal{Z}_i$~\hskip.3em{\bf do}
				\hskip1.0em $\rho_j\gets \rho_j+u_s\cdot \underline{w_{(s,j)}}$
				\label{full:r3_end}
			}
		\EndFor
		\ForAll{$j\in \mathcal{Z}_i$}
		\label{full:verif_start}
			\State{{\bf If}\hskip.3em~$\rho_j > \rho_{(max)}$\hskip.3em~{\bf then}\hskip1.0em
				$\rho_{(max)}\!\leftarrow \rho_j\hskip.6em \mbox{and}\hskip.6em a(i)\leftarrow j$
			} \label{full:verif_end}
		\EndFor

		\State{$C_{a(i)}^{[r]}\leftarrow C_{a(i)}^{[r]}\cup\{\hat{\bm{x}}_i\}$}
		\label{full:out_end}

	\EndFor
  \end{algorithmic}
\end{algorithm}

\begin{algorithm}[th]
\small
\newcommand{\mf}{\mbox{($mf$\hskip0.1em)}}
\newcommand{\mfH}{\mbox{($mfH$\hskip0.1em)}}
\newcommand{\mfM}{\mbox{($mfM$\hskip0.1em)}}
\newcommand{\algto}{\textbf{to}}
\algnewcommand{\LineComment}[1]{\Statex \(\hskip.8em\triangleright\) #1}

  \caption{\hskip.8em Candidate-gathering function: $G_1$} 
  \label{algo:icp_gather1}
  \begin{algorithmic}[1]
    \Statex{\textbf{Input:}\hskip.8em $\hat{\bm{x}}_i$,~~$\breve{\mathcal{M}}^{[r-1]}$,~
					$t^{[th]}, v^{[th]}$, 
					$\mathcal{Z}_i$,~$\{\rho_j,\,y_{(i,j)}\}_{j=1}^K$,\par
					\hskip1.5em$\rho_{(max)}$,$(nMv)$  \Comment{(nMv): \# moving centroids} 
					}
	\Statex{\textbf{Output:}\hskip.8em $\mathcal{Z}_i$,~~$\{ \rho_j \}_{j=1}^K$}

	\LineComment{Exact partial similarity calculation}
	\For{$(s \leftarrow t_{(i,p)}) < t^{[th]}$ in all term IDs in $\hat{\bm x}_i$\hskip.3em}
	\label{icp:r1_start}\Comment{Region 1}
		\For{$1\leq q\leq \mfM_s$}
			\label{icp:innermost1}
			\State{$\rho_{c_{(s,q)}}\gets \rho_{c_{(s,q)}}+u_{s}\cdot v_{c_{(s,q)}}$}
			\label{icp:r1_end}
		\EndFor	
	\EndFor

	\For{$(s \leftarrow t_{(i,p)}) \geq t^{[th]}$ in all term IDs in $\hat{\bm x}_i$\hskip.3em}
	\label{icp:r2_start}\Comment{Region 2}
		\For{$1\leq q\leq \mfM_s$}
			\label{icp:innermost2}
			\State{$\rho_{c_{(s,q)}}\!\gets \rho_{c_{(s,q)}}\!+u_{s}\cdot v_{c_{(s,q)}}$,\par
				\hskip1.0em $y_{(i,c_{(s,q)})}\!\gets y_{(i,c_{(s,q)})}\! -u_{s}$
			} \label{icp:r2_end}
		\EndFor	
	\EndFor

	\LineComment{Upper-bound calculation}
	\For{$1\leq j'\leq (nMv)$}
	\label{icp:gather_start}
		\State{$j'$ is transformed to $j$.}
		\label{icp:trans}
		\State{$\rho_{j}^{[ub]}\leftarrow \rho_{j}$\, \underline{$+\, \{ y_{(i,j)}\cdot v^{[th]} \}$}
					\hskip.05em\footref{footnote:scaling} }
		\Comment{Region 3 (UB)}

	\LineComment{ES filtering}
		\State{{\bf if}\hskip.6em~$\rho_{j}^{[ub]} > \rho_{(max)}$\hskip.6em~{\bf then}
			\hskip1.2em~$\mathcal{Z}_i \gets \mathcal{Z}_i \cup \{j\}$
		} \label{icp:ubp}
		\label{icp:gather_end}
	\EndFor
  \end{algorithmic}
\end{algorithm}

Algorithms~\ref{algo:full} and \ref{algo:icp_gather1} show the pseudocodes of the 
ES-ICP algorithm and Table~\ref{table:nota_prop} shows the symbols 
(The complete pseudocodes are in Appendix~\ref{app:fullalgo}).
Algorithm~\ref{algo:full} differs from MIVI in Algorithm~\ref{algo:mivi_assign} 
in the following two phases: 
the gathering phase where the candidate centroids for the similarity calculations of the $i$th object 
are collected in $\mathcal{Z}_i$ 
using the ES and ICP filters at lines~\ref{full_icp_switch:g1_start}
to \ref{full_icp_switch:g0_start}
and the verification phase where exact similarities for the centroids passing through
the filters are calculated at lines~\ref{full:r3_start} to \ref{full:r3_end}
and compared with similarity threshold $\rho_{(max)}$ at line~\ref{full:verif_end}.
The two filters effectively prune unnecessary centroids and 
reduce the multiplications. 
This decreases the number of instructions, which is 
one of the performance-degradation factors.

Algorithm~\ref{algo:icp_gather1} shows the $G_1$ function at line~\ref{full_icp_switch:g1}
in Algorithm~\ref{algo:full}.
The function works for the $i$th object if it is more similar: 
it calculates upper-bound similarity $\rho_j^{[ub]}$ in Eq.~(\ref{eq:ub}) 
and returns both the candidate centroid-ID set $\mathcal{Z}_i$ and 
the sum of the partial similarities in Regions 1 and 2.
Owing to the structured mean-inverted index with the two structural parameters 
shared in all the objects, 
the function can avoid using irregular conditional branches for the similarity calculations
only by setting loop endpoints at $(mfM)_s$ in each mean-inverted array.
The $G_0$ function in Algorithm~\ref{algo:full} 
differs from the $G_1$ in the endpoints in Regions 1 and 2 and the number of centroids 
for calculating the upper bounds on similarities.
Their values in $G_1$ are $(mfM)_s$, $(mfM)_s$, and $(nMv)$ 
while those in $G_0$ are $(mf)_s$, $(mfH)_s$, and $K$.
The functions of $G_1$ and $G_0$ successfully work in the AFM, i.e., 
they avoid using irregular conditional branches that induce branch mispredictions 
and loading unnecessary large arrays that cause cache misses. 
Thus, our ES-ICP algorithm designed by exploiting the UCs works in the AFM, 
resulting in high-speed performance.

\section{Structural Parameters}\label{sec:strparam}
To make our ES-ICP more effective, 
we estimate structural parameters $t^{[th]}$ and $v^{[th]}$
depending on the given data set and $K$ value
rather than clamped at predetermined values.
Then, we has to perform the estimation in the ES-ICP algorithm 
with as low computational cost as possible,
e.g., much less than that required at the one iteration.

We propose an estimation algorithm for determining the parameters\footnote{
The estimation algorithm is integrated into the ES-ICP algorithm.
Its required elapsed time is merged with the elapsed time spent by the 
ES-ICP for a fair comparison.}.
Our proposed estimation algorithm minimizes objective function $J$ of 
the approximate number of the multiplications for similarity calculations 
$\tilde{\phi}$
(derived in Appendix~\ref{app:objfunct}), which is expressed by
\begin{equation}
( t^{[th]},v^{[th]} ) = 
\argmin_{\substack{v_h^{[th]}\in V^{[th]}\\ s_{(min)}\leq s'\leq D}}
\left(\, J(s',v_h^{[th]})\,\right) \label{eq:2params}
\end{equation}
\small
\begin{eqnarray}
J(s',v_h^{[th]}) &=& \tilde{\phi}_{(s',h)} \nonumber \\
&=& (\phi 1)_{s'} +(\phi 2)_{(s',h)} +(\tilde{\phi}3)_{(s',h)}\, , 
\label{eq:objfunct2}
\end{eqnarray}
\normalsize
\noindent where $s'$ is a candidate of $t^{[th]}$,
$s_{(min)}$ the predetermined minimum value of $s'$,
and $v_h^{[th]}$ the $h$th candidate of $v^{[th]}$ 
that is prepared before ES-ICP starts.
The numbers of multiplications in Regions 1 and 2,
$(\phi 1)_{s'}$ and $(\phi 2)_{(s',h)}$, are expressed by
\small
\begin{eqnarray}
&&(\phi 1)_{s'} = \sum_{s=1}^{s'-1} (df)_s\cdot (mf)_s \label{eq:cost1} \\
&&(\phi 2)_{(s',h)} = \sum_{s=s'}^D (df)_s\cdot (mfH)_{(s,v_h^{[th]})}\, ,
\label{eq:cost2}
\end{eqnarray}
\normalsize
where $(\phi 1)_{s'}\!+(\phi 2)_{(s',h)}$ is exactly determined, 
given the tentative $s'$ and $v_h^{[th]}$.
By contrast, 
it is difficult to exactly determine with low computational cost
the number of multiplications in Region 3.
To approximate this number so as to compute with low cost, 
we introduce a probability that a centroid passes through the ES filter,
which corresponds to $|{\mathcal Z}_i|/K$,
and express an approximate number as the expected value  $(\tilde{\phi} 3)_{(s',h)}$\,:
\small 
\begin{equation}
(\tilde{\phi} 3)_{(s',h)} = \sum_{i=1}^{N} (ntH)_{(i;s')}\! \cdot K\!\cdot 
\mbox{Prob}\left(\rho^{[ub]}(i)\!\geq\! \rho_{a(i)}\,;s',\!h \right),
\label{eq:cost3}
\end{equation}
\normalsize
where 
$(ntH)_{(i;s')}$ is the number of the terms whose IDs are more than or equal to $s'$, 
$\rho^{[ub]}(i)$ denotes the distribution function of the upper-bound similarity 
of the $i$th object (Appendix~\ref{app:est}),
and 
$\mbox{Prob}(\rho^{[ub]}(i)\!\geq\! \rho_{a(i)}\,;s',h)$
is the probability of $\rho^{[ub]}(i)\!\geq \rho_{a(i)}$, 
given $s'$ and $h$ (used instead of $v_h^{[th]}$).
The probability is represented by 
\begin{equation}
\mbox{Prob}\,\left( \rho^{[ub]}(i)\!\geq \rho_{a(i)}\,;s',h \right)
=\left( \frac{1}{K}\right)\,
\left( \frac{K}{e} \right)^{\frac{\Delta\bar{\rho}(i;s',h)}{\rho_{a(i)} -\bar{\rho}_i} }
\label{eq:prob_unpruned1}
\end{equation}
\begin{equation}
\Delta\bar{\rho}(i;s',h) = \bar{\rho}^{[ub]}(i\,;s',h) -\bar{\rho}_i \, , 
\label{eq:Deltabarrho}
\end{equation}
where $\bar{\rho}^{[ub]}(i\,;s',h)$ denotes
the average upper bound on the similarities of the $i$th object to all the centroids 
and $\bar{\rho}_i$ denotes the average similarity to them (derived in Appendix~\ref{app:est}).
Then Eq.~(\ref{eq:cost3}) is rewritten as 
\begin{equation}
(\tilde{\phi} 3)_{(s',h)} = \sum_{i=1}^{N} (ntH)_{(i,s')} \cdot
\left( \frac{K}{e} \right)^{\frac{\Delta\bar{\rho}(i;s',h)}{\rho_{a(i)} -\bar{\rho}_i} }\! .
\label{eq:simple_cost3}
\end{equation}
By substituting Eqs.~(\ref{eq:cost1}), (\ref{eq:cost2}), and (\ref{eq:simple_cost3})
into Eq.~(\ref{eq:objfunct2}),
the objective function is expressed by
{\small
\begin{equation}
J(s',v_h^{[th]}) = \sum_{s=1}^{s'-1} (df)_s\cdot (mf)_s 
+\!\sum_{s=s'}^D (df)_s (mfH)_{(s,v_h^{[th]})} \nonumber 
\end{equation}
}
\begin{equation}
+\!\sum_{i=1}^N\, (ntH)_{(i,s')}\cdot 
\left( \frac{K}{e} \right)^{\frac{ \Delta\bar{\rho}(i;s',h)}{\rho_{a(i)} -\bar{\rho}_i} } \! .
\label{eq:objfunct_all}
\end{equation}
We obtain the two structural parameters of $t^{[th]}$ and $v^{[th]}$ 
by efficiently solving Eq.~(\ref{eq:2params}) 
using the objective function in Eq.~(\ref{eq:objfunct_all})
(detailed in Appendix~\ref{app:est}).

\section{Experiments}\label{sec:exp}
We first describe the data sets and the extracted feature values, 
followed by a platform for the algorithm evaluation 
and performance measures.
We next compare our ES-ICP with the baseline MIVI 
and three algorithms, ICP, TA-ICP, and CS-ICP,
which were simply extended 
from existing search and clustering algorithms based on the state-of-the-art techniques
in our setting \cite{fagin01,li,bottesch,knittel}.
ICP employs only the auxiliary filter designed so as to operate in the AFM.
TA-ICP and CS-ICP incorporate their main UBP filters besides the ICP filter.
For its main filer, TA-ICP modifies the threshold algorithms (TA)  
in the search algorithms in Fagin$^+$ \cite{fagin01} and Li$^+$ \cite{li}.
CS-ICP utilizes the Cauchy-Schwarz inequality, 
which is widely used for making upper bounds on an inner product of the vectors
\cite{bottesch,knittel}.
TA-ICP and CS-ICP are described in Section~\ref{ssec:perfcomp} 
and detailed in Section~\ref{app:compalgo}.

\subsection{Data Sets}\label{ssec:data}
We employed two different types of large-scale and high-dimensional 
sparse real document data sets: 
{\em PubMed Abstracts} (PubMed) \cite{pubmed}
and {\em The New York Times Articles} (NYT).

PubMed contains 8\,200\,000 documents (data objects) 
each of which was represented by term (distinct word) counts, called 8.2M-sized PubMed.
The data set contained terms corresponding to the dimensionality of 141\,043.
The average number of non-zero elements in the objects was 58.96, 
and the sparsity indicator ($\hat{D}/D$) (Section~\ref{sec:intro})
was $4.18\!\times\! 10^{-4}$.
By contrast, the average number of non-zero elements in the centroids was 2094.94, 
which is 35.53 times larger than that in the objects, 
when $K\!=\!8\!\times\! 10^4$.
Note that mean-feature vectors are still sparse although they contain more terms
than object-feature vectors on average.

Regarding NYT,
we extracted 1\,285\,944 articles (data objects) from {\em The New York Times Articles} 
from 1994 to 2006 and counted the occurrence frequencies of the terms
after stemming and stop word removal.
The number of resultant terms in all the objects was 495\,126.
The average number of non-zero elements in the objects was 225.76, 
where $\hat{D}/D\!=\!4.56\!\times\!10^{-4}$.
The average number of non-zero elements in the centroids was 5105.73,
which is 22.62 times larger than that in the objects 
when $K\!=\!1\!\times\! 10^4$.

We made an object-feature vector from the occurrence frequencies of the terms in the data set.
An element of the feature vector was the value of the classic {\em tf-idf}
(term frequency-inverse document frequency),
normalized by the $L_2$-norm of the feature vector.
The {\em tf-idf}\, value of the $s$th term in the $i$th document 
was defined as 
\begin{equation}
\mbox{\em tf-idf}\,(s,i) = 
	\mbox{\em tf}\,(s,i)\times \log\left( \frac{N}{(df)_s} \right)\: ,
\label{eq:tf-idf}
\end{equation}
where {\em tf}\,$(s,i)$ denotes the raw counts of the $s$th term in the $i$th document 
and $(df)_s$ denotes the document frequency of the $s$th term.
Due to normalization, 
each feature vector was regarded as a point on a unit hypersphere.

\subsection{Platform and Performance Measures}\label{ssec:platform}
The algorithms were executed on a computer system 
that was equipped with two Xeon E5-2697v3 2.6-GHz CPUs 
with three-level caches from levels 1 to 3 (last level) 
and 256-GB main memory, 
by multithreading with OpenMP \cite{openmp} of 50 threads within the memory capacity.
The CPU performed  
the out-of-order superscalar execution with eight issue widths.
The algorithms were implemented in C and compiled with a GNU C compiler 
version 8.2.0 on the optimization level of {\sf -O3}. 

The performances were evaluated with the following measurements:
elapsed time until convergence, 
number of multiplications, 
performance-degradation factors,
and maximum size of the physical memory occupied through the iterations.
The accuracy of the clustering results was not our performance measure
because the algorithms were just accelerations of the spherical $K$-means.

\subsection{Compared Algorithms}\label{ssec:compalgo}
We evaluated our proposed algorithm ES-ICP by comparing it with the baseline algorithm MIVI 
and three algorithms using state-of-the-art techniques in our setting.
The first algorithm only uses the ICP filter, which is referred to as ICP.
The second algorithm combined a method inspired by Fagin$^+$'s TA \cite{fagin01} 
and Li$^+$'s cosine-threshold algorithm \cite{li} with ICP (TA-ICP). 
The third algorithm incorporated the Cauchy-Schwarz inequality as in 
Bottesch$^+$ \cite{bottesch} or Knittel$^+$ \cite{knittel} to ICP (CS-ICP).
We briefly describe TA-ICP and CS-ICP (detailed in Appendix~\ref{app:compalgo}).
Both algorithms used the identical framework as 
ES-ICP for a fair comparison, based on a three-region partition 
in their mean-inverted indexes and two phases for gathering candidates and their verification.
Incorporating parameter $t^{[th]}$ leads to save the required memory size of 
$\breve{\mathcal M}^{p[r]}$ at the verification phase in Section~\ref{ssec:icp}.
The $t^{[th]}$ was preset at $0.9\,D$, 
which was close to the value 
obtained by our estimation algorithm in Section~\ref{sec:strparam}.
TA-ICP and CS-ICP differed from ES-ICP mostly in the following four points: 
\begin{enumerate}
\item main upper-bound-based pruning (UBP) filters
\item structures of the partial mean-inverted indexes
\item usage of additional partial mean-inverted indexes
\item operations in the gathering phase and the exact similarity calculations.
\end{enumerate}

\subsubsection{TA-ICP}\label{sssec:taicp}
TA-ICP employs an individual structural parameter ({\em threshold}) on the mean-feature values 
for each object
while our algorithm uses the common structural parameter $v^{[th]}$, 
which is shared among all the objects.
The individual threshold $v_{(ta)i}^{[th]}$ for the $i$th object is defined by 
\begin{equation}
v_{(ta)i}^{[th]} = \rho_{(max)}/\| {\bm x}_i\|_1\, , \label{eq:ta_th}
\end{equation}
where $\rho_{(max)}$ denotes the similarity of the $i$th object to 
the centroid to which the object is assigned at the last iteration and 
$\|{\bm x}_i\|_1$ the $L_1$-norm of the object, 
i.e., $\|{\bm x}_i\|_1\!=\!\sum_{p=1}^{(nt)_i} u_{t_{(i,p)}}$.
Upper bound $\rho_{(j;i)}^{[ub]}$ on the similarity to the $j$th centroid,
which is used in the main filter, is expressed by
\begin{eqnarray}
&&\rho_{(j;i)}^{[ub]} = (\rho 1)_{(j;i)} +(\rho 2')_{(j;i)}
+v_{(ta)i}^{[th]}\cdot y_{(i,j)}\footnotemark
\label{eq:ta-ub}\\
&&\hskip1.2em y_{(i,j)}=\sum_{s\geq t^{[th]}}\!u_{(s,j)}\,,
\hskip1.2em s=t_{(i,p)}\,,
\end{eqnarray}
\footnotetext{To this multiplication,
the scaling like that in ES-ICP\,\footref{footnote:scaling} for reducing multiplications 
cannot be applied due to using the individual threshold.}
where $(\rho 2')_{(j;i)}$ denotes partial similarity in Region~2, 
determined by $v_{(ta)i}^{[th]}$, 
and $y_{(i,j)}$ is the partial $L_1$-norm of the $i$th object-feature vector,
which remains unused for $(\rho 1)_{(j:i)}\!+(\rho 2')_{(j;i)}$, 
and $u_{(s,j)}$ denotes the object-feature value with the $s$th term ID, 
given the $j$th centroid.
The upper bound $\rho_{(j;i)}^{[ub]}$ in Eq.~(\ref{eq:ta-ub}) works as the UBP filter,
compared with $\rho_{(max)}$.

TA-ICP uses a special {\em partial} mean-inverted index, 
each array of which is sorted in descending order of mean-feature values,
and calculates a part of $(\rho 2')_{(j;i)}$ 
while going down in the array from the top to the position of threshold $v_{(ta)i}^{[th]}$.

To combine the main filter TA with ICP,
an {\em additional} sorted mean-inverted index only for moving centroids was incorporated 
because TA-ICP needs a sorted mean-inverted index.
For calculating $(\rho 2')_{(j;i)}$, the additional sorted mean-inverted index is used 
if the $i$th object is more similar in Eq.~(\ref{eq:objstate}), 
otherwise, the sorted mean-inverted index is done.

TA-ICP prepares another partial mean-inverted index for calculating exact similarities of centroids
passing through the filters, which corresponds to $\breve{M}^{p[r]}$ in ES-ICP
(Table~\ref{table:nota_prop}) 
but differs in that it has all the mean-feature values 
due to using the individual threshold $v_{(ta)i}^{[th]}$.
When calculating the exact similarity, it has to skip with conditional branches 
the mean-feature values larger than or equal to $v_{(ta)i}^{[th]}$ 
that are already being utilized for $(\rho 2')_{(j;i)}$.

TA-ICP suffers from the disadvantages of 
more BMs and LLCMs in Table~\ref{table:comp}. 
These are attributed to properties of the TA algorithm itself,
that is, finding out a termination point of the gathering phase in its sorted inverted index
like Fagin$^+$ and Li$^+$.
The BMs were caused by
the conditional branches that irregularly return their judgments of 
whether or not a current point is the termination one.
The LLCMs were induced by simultaneously using the three distinct arrays, 
which were for the partial similarity $(\rho 1)_{(j:i)}\!+(\rho 2')_{(j;i)}$, 
the partial mean-inverted index, and the additional one.

\subsubsection{CS-ICP}\label{sssec:csicp}
CS-ICP employs upper bound $\rho_{(j;i)}^{[ub]}$ on the similarity to the $j$th centroid
based on the Cauchy-Schwarz inequality in Regions 2 and 3, 
which is expressed by
\begin{eqnarray}
&&\rho_{(j;i)}^{[ub]} = (\rho 1)_{(j;i)} +
\|{\bm x}_i^{p}\|_2 \times \sqrt{\|{\bm \mu}_{(j;i)}^{p}\|_2^2}\label{eq:cs-ub}\\ 
&&\hskip1.2em \|{\bm x}_i^{p}\|_2 = \sqrt{\sum_{t_{(i,p)}\geq t^{[th]}} u_{t_{(i,p)}}^2}\\
&&\hskip1.2em \|{\bm \mu}_{(j;i)}^{p}\|_2^2 = 
\sum_{\substack{t_{(i,p)}\geq t^{[th]}\\ s\gets t_{(i,p)}}} v_{c_{(s,q)}}^2 ,
\hskip1.2em j = c_{(s,q)}\, ,
\end{eqnarray}
where the second term on the right-hand side in Eq.~(\ref{eq:cs-ub}) is 
the upper bound on the partial similarity 
in the subspace spanned by the bases of the $i$th object's inherent dimensions 
of $t_{(i,p)}\!\geq \!t^{[th]}$.
The UBP filter compares the upper bound $\rho_{(j;i)}^{[ub]}$ in Eq.~(\ref{eq:cs-ub}) 
with $\rho_{(max)}$.

Applying the Cauchy-Schwarz inequality to the subspace 
requires the $L_2$-norms of the mean-feature vectors in the subspace.
CS-ICP calculates $\sqrt{\|{\bm \mu}_{(j;i)}^p\|_2^2}$ on the fly depending on a given object
and multiplies it with $\|{\bm x}_i^p\|_2$ $(i\!=\!1,2,\cdots,N)$ that is pre-calculated and stored
owing to the preset $t^{[th]}$.
To avoid the excessive sum of squares calculations of $\|{\bm \mu}_{(j;i)}^p\|_2^2$, 
CS-ICP prepares an additional partial squared-mean-inverted index each entry of which is a tuple of 
(mean ID, squared mean-feature value) in Regions 2 and 3 with a computational cost of
$\sum_{s=t^{[th]}}^D (mf)_s$.
We, however, still need to perform the square-root operation with high computational cost.

CS-ICP has the disadvantage of more LLCMs in Table~\ref{table:comp}.
This is caused by simultaneously using the three distinct arrays for calculating 
an upper bound on the similarity,
which are for the partial similarity $(\rho 1)_{(j;i)}$, object's partial $L_1$ norm $\|{\bm x}_i^p\|_2$,
and squared mean-feature values.
These are indispensable for the multiply-add operation for the upper-bound calculation.

\subsection{Performance Comparison}\label{ssec:perfcomp}
\begin{figure}[t]
\begin{center}
	\subfloat[{\normalsize Number of multiplications}]{ 
		\hspace*{2mm}
		\psfrag{X}[c][c][0.90]{
			\begin{picture}(0,0)
				\put(0,0){\makebox(0,-4)[c]{Iterations}}
			\end{picture}
		}
		\psfrag{Y}[c][c][0.88]{
			\begin{picture}(0,0)
				\put(0,0){\makebox(0,30)[c]{\# multiplications}}
			\end{picture}
		}
		\psfrag{A}[c][c][0.80]{$0$}
		\psfrag{B}[c][c][0.80]{$20$}
		\psfrag{C}[c][c][0.80]{$40$}
		\psfrag{D}[c][c][0.80]{$60$}
		\psfrag{H}[r][r][0.80]{$10^6$}
		\psfrag{I}[r][r][0.80]{$10^{8}$}
		\psfrag{J}[r][r][0.80]{$10^{10}$}
		\psfrag{K}[r][r][0.80]{$10^{12}$}
		\psfrag{L}[r][r][0.80]{$10^{14}$}
		\psfrag{P}[r][r][0.65]{MIVI}
		\psfrag{Q}[r][r][0.64]{ES-ICP}
		\psfrag{R}[r][r][0.65]{TA-ICP}
		\psfrag{S}[r][r][0.65]{CS-ICP}
		\psfrag{T}[r][r][0.65]{ICP}
		\includegraphics[width=41mm]{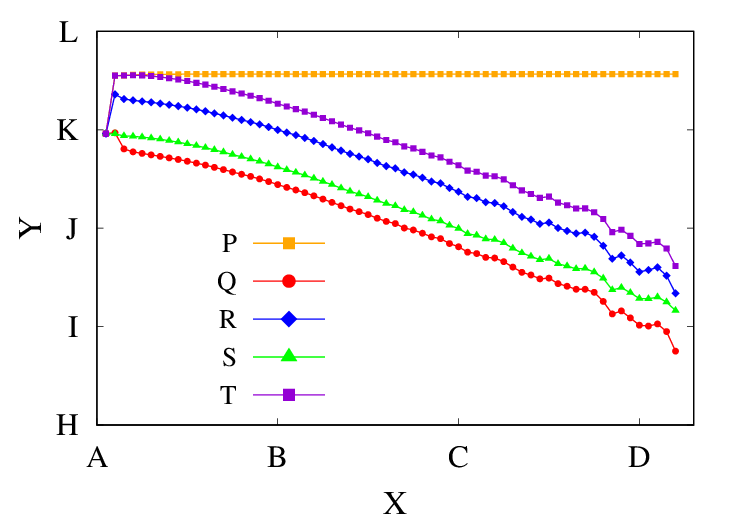}
	}\hspace*{1.0mm}
	\subfloat[{\normalsize Compl. pruning rate}]{
		\psfrag{X}[c][c][0.90]{
			\begin{picture}(0,0)
				\put(0,0){\makebox(0,-4)[c]{Iterations}}
			\end{picture}
		}
		\psfrag{Y}[c][c][0.82]{
			\begin{picture}(0,0)
				\put(0,0){\makebox(0,33)[c]{Compl. pruning rate}}
			\end{picture}
		}
		\psfrag{A}[c][c][0.80]{$0$}
		\psfrag{B}[c][c][0.80]{$20$}
		\psfrag{C}[c][c][0.80]{$40$}
		\psfrag{D}[c][c][0.80]{$60$}
		\psfrag{H}[r][r][0.78]{$10^{-8}$}
		\psfrag{I}[r][r][0.78]{$10^{-6}$}
		\psfrag{J}[r][r][0.78]{$10^{-4}$}
		\psfrag{K}[r][r][0.78]{$10^{-2}$}
		\psfrag{L}[r][r][0.78]{$1$}
		\psfrag{P}[r][r][0.63]{ICP}
		\psfrag{Q}[r][r][0.63]{TA-ICP}
		\psfrag{R}[r][r][0.63]{CS-ICP}
		\psfrag{S}[r][r][0.61]{ES-ICP}
		\includegraphics[width=41mm]{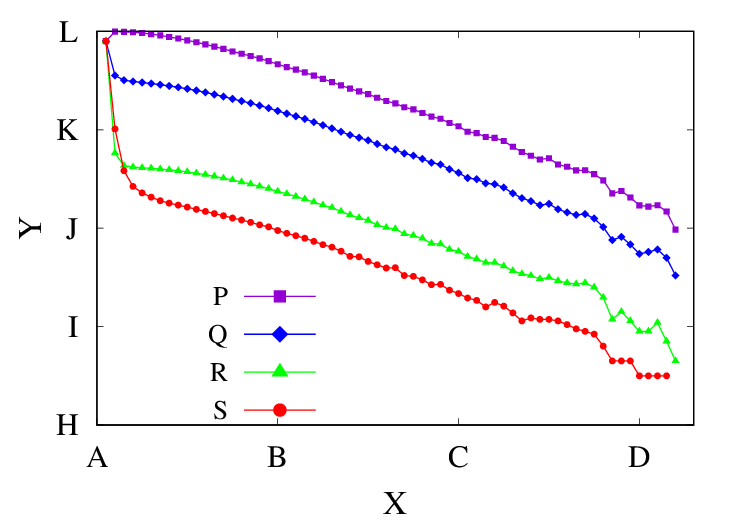}
}
\end{center}
\vspace*{-1mm}
\caption{\small
Algorithm performance in 8.2M-sized PubMed at K=80\,000:
(a) Number of multiplications and  
(b) Complementary pruning rate (CPR) along iterations until convergence.
}
\label{fig:comp}
\vspace*{-1mm}
\end{figure}

\begin{figure}[t]
\begin{center}
		\psfrag{X}[c][c][0.90]{
			\begin{picture}(0,0)
				\put(0,0){\makebox(0,-4)[c]{Iterations}}
			\end{picture}
		}
		\psfrag{Y}[c][c][0.86]{
			\begin{picture}(0,0)
				\put(0,0){\makebox(-2,8)[c]{Elapsed time ($\times 10^3$sec)}}
			\end{picture}
		}
		\psfrag{A}[c][c][0.80]{$0$}
		\psfrag{B}[c][c][0.80]{$20$}
		\psfrag{C}[c][c][0.80]{$40$}
		\psfrag{D}[c][c][0.80]{$60$}
		\psfrag{H}[r][r][0.80]{$0$}
		\psfrag{I}[r][r][0.80]{$1$}
		\psfrag{J}[r][r][0.80]{$2$}
		\psfrag{K}[r][r][0.80]{$3$}
		\psfrag{L}[r][r][0.80]{$4$}
		\psfrag{P}[r][r][0.80]{MIVI}
		\psfrag{Q}[r][r][0.80]{ES-ICP}
		\psfrag{R}[r][r][0.80]{TA-ICP}
		\psfrag{S}[r][r][0.80]{CS-ICP}
		\psfrag{T}[r][r][0.80]{ICP}
		\includegraphics[width=48mm]{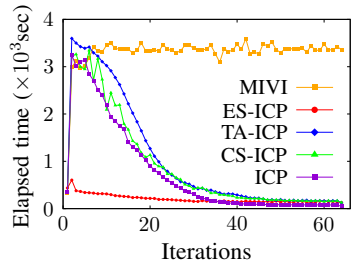}
\end{center}
\vspace*{-1mm}
\caption{\small
Elapsed time along iterations until convergence in 8.2M-sized PubMed at K=80\,000}
\label{fig:time_comp}
\vspace*{-3mm}
\end{figure}

Figures~\ref{fig:comp}(a) and (b) 
show the number of multiplications (Mult) 
for similarity calculations, including their upper-bound calculations,
and the complementary pruning rate (CPR) 
along the iterations until convergence when the algorithms were applied to 
8.2M-sized PubMed with $K\!=\! 80\,000$.
The CPR is defined by
\begin{equation}
\mbox{CPR}= \frac{1}{N}\sum_{i=1}^N \frac{|\mathcal{Z}_i|}{K}\,,
\end{equation}
where $|\mathcal{Z}_i|$ denotes the number of centroids 
passing through the filter\footnote{
$|\mathcal{Z}_i|$ approximately corresponds to the second term in the right-hand side 
in Eq.~(\ref{eq:prob_unpruned1}).
}.
An algorithm achieving a lower CPR value has better filter function. 
The Mult and CPR baselines are those of MIVI, 
whose CPR is $1.0$ because it calculates the exact similarities 
between all the objects and centroids.
The effects of the main filters in TA-ICP, CS-ICP, and 
ES-ICP are evaluated, 
compared with ICP 
that reduced both Mult and CPR with increasing the iterations.
Figure~\ref{fig:time_comp} shows the elapsed time required by  each algorithm 
in the same setting in Fig.~\ref{fig:comp}.
Table \ref{table:comp} shows the performance rates of the compared algorithms to 
ES-ICP in a similar format in Table~\ref{table:ivfd} 
where Max MEM in the last column is the rate of the maximum memory size through the iterations.

\begin{table}[t]
\small
\caption{\small
Performance comparison}
\label{table:comp}
\centering
\vspace*{-1mm}
\begin{tabular}{|c|c|c||c|c|c||c|}\hline
\multirow{2}{*}{Algo} 
& Avg & Avg & \multirow{2}{*}{Inst} &\multirow{2}{*}{BM} &\multirow{2}{*}{LLCM} & Max\\
& Mult & time &                     &                    &                      & MEM\\ \hline
MIVI & 141.2 &   16.13  & 16.53  & 4.082 & 10.91  & 0.4935 \\ \hline
ICP  &   31.52 &  3.709 &  4.641 & 2.905 &  2.759 & 0.4955 \\ \hline
CS-ICP & 1.845 &  4.404 &  3.785 & 3.249 &  4.956 & 1.095 \\ \hline
TA-ICP & 9.656 &  5.086 &  2.381 & 19.31 & 13.64  & 1.141 \\ \hline
\end{tabular}
\vspace*{-1mm}
\end{table}

ES-ICP significantly suppressed the elapsed time compared to the others 
from the early to the last stage in the iterations.
As a result, 
ES-ICP worked over 15 times faster than MIVI 
and at least 3.5 times faster than the others.
From the values in the three evaluation items of Inst, BM, and LLCM,
we know that ES-ICP was designed so as to operate in the AFM.
By contrast, TA-ICP and CS-ICP needed more elapsed time than ICP although they had better filters.
Their inferior speed-performances are attributed to the difficulty of designing the algorithms
in the AFM.
Compared with ICP, CS-ICP and TA-ICP caused many last-level-cache load misses (LLCM) 
because these algorithms needed additional data loaded to the limited-capacity caches.
CS-ICP stored the squared mean-feature values in the partial squared-mean-inverted index
whose term IDs ($s$) were in the range of $t^{[th]}\!\leq\! s\!\leq\! D$.
TA-ICP stored the mean-feature values of the moving centroids in the additional mean-inverted index
for all the term IDs.
TA-ICP also caused many branch mispredictions (BM)
due to the two additional conditional branches for 
terminating the exact partial similarity calculation in Region 2
and avoiding the double similarity calculations for the centroids passing through its main filter.

Thus, the proposed algorithm ES-ICP achieved higher performance
by reducing the performance-degradation factors of Mult, Inst, BM, and LLCM.
Regarding the memory usage in Table~\ref{table:comp}, 
the three algorithms of ES-ICP, CS-ICP, and TA-ICP needed 
around double the maximum memory size (Max MEM) of MIVI and ICP 
since they used partial mean-inverted index $\breve{\mathcal M}^{p[r]}$ 
for exact similarity calculations for centroids passing through their filters
at the verification phase, which is designed so as to work in the AFM.
To suppress the maximum memory size, we have to develop new techniques.

\section{Discussion}\label{sec:disc}
We discuss why the ES-ICP algorithm works efficiently
from the two viewpoints of the AFM in Section~\ref{sec:arch}  
and the effective use of the UCs
of large-scale and high-dimensional sparse document data sets 
in Section~\ref{sec:dchar}.
First, we classify the algorithms and highlight their characteristics.
Next, we describe the efficient operation of our ES filter,
relating with both the feature-value-concentrarion phenomenon and 
the estimated structural parameter (threshold) $v^{[th]}$. 
Last, we show that ES-ICP efficiently works 
in the AFM for not only the PubMed but also the NYT data set based on their UCs.

\subsection{Algorithm Classification}\label{ssec:class}
Regarding these two viewpoints, Table~\ref{table:algoclass} shows
a coarse classification of the algorithms using the mean-inverted index 
in Sections~\ref{ssec:compalgo}. 
The rows and columns respectively represent 
the two classes of whether the algorithms effectively used the UCs or not and
the three classes of how the algorithms were designed so as to operate in the AFM.
Note, however, that all the algorithms are designed to some level in the AFM 
since they adopt their structured mean-inverted index 
instead of the data-object-inverted index in Section~\ref{sec:arch}.

\begin{table}[t]
\small
\caption{\small Algorithm classification}
\label{table:algoclass}
\centering
\begin{tabular}{|c||c|c|c|}\hline
Effective use of& \multicolumn{3}{c|}{Architecture friendly}\\\cline{2-4}
UCs & High & Moderate & Low \\ \hline
Good  & ES-ICP & CS-ICP & TA-ICP\\ \hline
Poor  & - & ICP & MIVI\\ \hline
\end{tabular}
\end{table}

MIVI is classified into the class where UCs are not used and 
with a low-level design with respect to the AFM.
This is because 
MIVI without an ICP filter cannot reduce the instructions 
even at the last stage in the iterations.
Furthermore, continuing almost the constant number of the instructions  
causes that of cache misses to last.

ICP does not effectively use the UCs even though it is designed in a moderate-level AFM.
It suppresses the number of instructions from the middle and last stages in the iteration
owing to its filter and the number of the branch mispredictions 
by its structured mean-inverted-index 
without conditional branches to judge whether or not each centroid is invariant.

ES-ICP, CS-ICP, and TA-ICP effectively utilize the UCs 
by partitioning the mean-inverted index into three regions in Fig.~\ref{fig:regions}.
However, they are classified into the different classes on the AFM 
for their different main UBP filters.
ES-ICP is designed in a high-level AFM.
It needs neither additional data nor conditional branches for its main filter,
which leverages the common $v^{[th]}$ shared with all the objects
in Section~\ref{sec:prop}.
By contrast, 
since CS-ICP and TA-ICP respectively use their additional data 
for the squared mean-feature-value arrays
and the sorted invariant and moving mean-feature-value arrays, 
they suffer many cache misses.
Moreover, TA-ICP employs individual threshold $v_{(ta)i}^{[th]}$ for every object 
in Eq.~(\ref{eq:ta_th}).
It needs the additional conditional branch 
for judging the magnitude relationship between $v_{(ta)i}^{[th]}$ and a mean-feature value,
resulting in increasing the branch mispredictions.
For these reasons, CS-ICP and TA-ICP are classified into the moderate- and low-level classes
on the AFM.

\begin{figure}[t]
\begin{center}
	\psfrag{X}[c][c][0.9]{
		\begin{picture}(0,0)
			\put(0,0){\makebox(0,-6)[c]{Mean-feature value}}
		\end{picture}
	}
	\psfrag{Y}[c][c][0.9]{
		\begin{picture}(0,0)
			\put(0,0){\makebox(0,28)[c]{Probability}}
		\end{picture}
	}
	\psfrag{P}[c][c][0.80]{$0$}
	\psfrag{Q}[c][c][0.80]{$0.2$}
	\psfrag{R}[c][c][0.80]{$0.4$}
	\psfrag{S}[c][c][0.80]{$0.6$}
	\psfrag{T}[c][c][0.80]{$0.8$}
	\psfrag{U}[c][c][0.80]{$1.0$}
	\psfrag{A}[r][r][0.80]{$0$}
	\psfrag{B}[r][r][0.80]{$0.2$}
	\psfrag{C}[r][r][0.80]{$0.4$}
	\psfrag{D}[r][r][0.80]{$0.6$}
	\psfrag{E}[r][r][0.80]{$0.8$}
	\psfrag{F}[r][r][0.80]{$1.0$}
	\psfrag{Z}[c][c][0.80]{Order~~}
	\psfrag{JJJJ}[r][r][0.72]{100}
	\psfrag{KKKK}[r][r][0.72]{10}
	\psfrag{LLLL}[r][r][0.72]{3}
	\psfrag{MMMM}[r][r][0.72]{2}
	\psfrag{NNNN}[r][r][0.72]{1}
	\includegraphics[width=52mm]{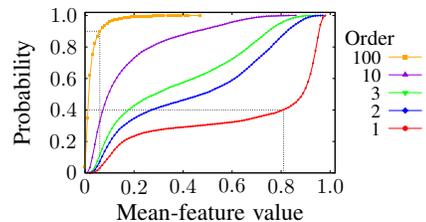}
\end{center}
\caption{\small
Probability less than or equal to mean-feature value 
for orders of 1, 2, 3, 10, 100 in mean-inverted-index arrays sorted in
descending order of their values in 8.2M-sized PubMed with K=80\,000
}
\label{fig:order_mfv}
\vspace*{-3mm}
\end{figure}

\subsection{ES Filter Using Feature-Value-Concentration Phenomenon}\label{ssec:es}
Our ES filter effectively exploits the UCs, in particular, 
the feature-value-concentration phenomenon
to make a tight upper bound on the similarity with low computational cost.
Figure~\ref{fig:order_mfv} shows the actual skewed form 
when ES-ICP with $K\!=\!80\,000$ was applied to 8.2M-sized PubMed.
Each curve represents the probability at which 
a mean-feature value at the order from top to $(mf)_s$ 
is less than or equal to the mean-feature value on the horizontal axis.
The mean-feature values in Regions 2 and 3 were used, i.e., 
term IDs ($s$) of the mean-inverted-index arrays satisfied $t^{[th]}\!\leq\!s\!\leq\!D$.
In Fig.~\ref{fig:order_mfv}, when the order is 1, 
the curve represents the probability distribution of 
the largest feature values in the mean-inverted-index arrays. 
The probability at a feature value of $0.81$ on the first-order curve was only 0.4
while even that at a feature value of $0.06$ on the 100th-order curve was 0.9. 
Considering that the maximum and average order were 75\,042 and 10\,341, respectively,
we know that very few elements in each array have large feature values and 
most of the remaining have very small values.
Based on this characteristic of the feature-value-concentration phenomenon, 
we can tighten the upper bounds on the similarity by few multiplications
in Section~\ref{ssec:frame}. 

\begin{figure}[t]
\begin{center}\hspace*{2mm}
	\subfloat[{\normalsize Mult before filtering }]{
		\psfrag{X}[c][c][0.90]{
			\begin{picture}(0,0)
				\put(0,0){\makebox(0,-5)[c]{Threshold}}
			\end{picture}
		}
		\psfrag{Y}[c][c][0.86]{
			\begin{picture}(0,0)
				\put(0,0){\makebox(0,28)[c]{\# multiplications}}
			\end{picture}
		}
		\psfrag{P}[c][c][0.78]{$10^{-3}$}
		\psfrag{Q}[c][c][0.78]{$10^{-2}$}
		\psfrag{R}[c][c][0.78]{$10^{-1}$}
		\psfrag{S}[c][c][0.78]{$1$}
		\psfrag{A}[r][r][0.78]{$10^{9}$}
		\psfrag{B}[r][r][0.78]{$10^{10}$}
		\psfrag{C}[r][r][0.78]{$10^{11}$}
		\psfrag{D}[r][r][0.78]{$10^{12}$}
		\psfrag{E}[r][r][0.78]{$10^{13}$}
		\psfrag{Z}[c][c][0.80]{$K$}
		\psfrag{J}[r][r][0.75]{$1\!\times\!10^{4}$}
		\psfrag{K}[r][r][0.75]{$2\!\times\!10^{4}$}
		\psfrag{L}[r][r][0.75]{$5\!\times\!10^{4}$}
		\psfrag{M}[r][r][0.75]{$8\!\times\!10^{4}$}
		\includegraphics[width=42mm]{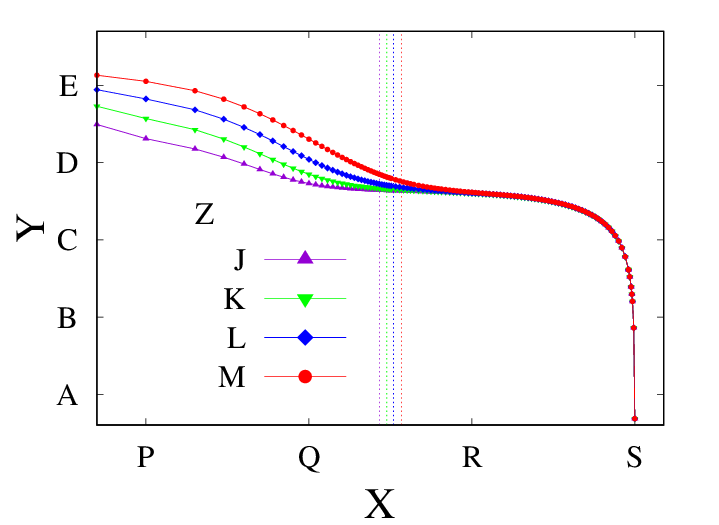}
	}\hspace*{1mm}
	\subfloat[{\normalsize Mult after filtering}]{ 
		\psfrag{X}[c][c][0.90]{
			\begin{picture}(0,0)
				\put(0,0){\makebox(0,-5)[c]{Threshold}}
			\end{picture}
		}
		\psfrag{Y}[c][c][0.86]{
			\begin{picture}(0,0)
				\put(0,0){\makebox(0,28)[c]{\# multiplications}}
			\end{picture}
		}
		\psfrag{P}[c][c][0.78]{$10^{-3}$}
		\psfrag{Q}[c][c][0.78]{$10^{-2}$}
		\psfrag{R}[c][c][0.78]{$10^{-1}$}
		\psfrag{S}[c][c][0.78]{$1$}
		\psfrag{A}[r][r][0.78]{$10^{9}$}
		\psfrag{B}[r][r][0.78]{$10^{10}$}
		\psfrag{C}[r][r][0.78]{$10^{11}$}
		\psfrag{D}[r][r][0.78]{$10^{12}$}
		\psfrag{E}[r][r][0.78]{$10^{13}$}
		\psfrag{Z}[c][c][0.78]{$K$}
		\psfrag{J}[r][r][0.75]{$1\!\times\!10^{4}$}
		\psfrag{K}[r][r][0.75]{$2\!\times\!10^{4}$}
		\psfrag{L}[r][r][0.75]{$5\!\times\!10^{4}$}
		\psfrag{M}[r][r][0.75]{$8\!\times\!10^{4}$}
		\includegraphics[width=42mm]{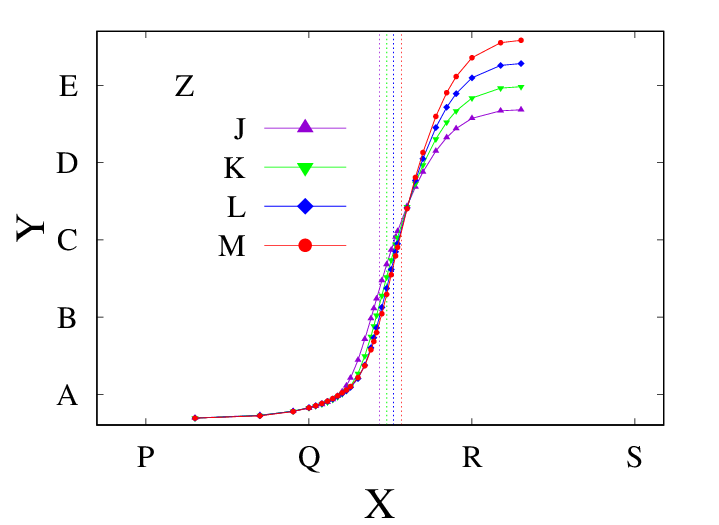}
	}
\end{center}
\vspace*{-2mm}
\caption{\small
Number of multiplications:
(a) before ES filtering with threshold on horizontal axis
and (b) for centroids passing through 
ES filter when algorithm was applied to 8.2M-sized PubMed.
Vertical dashed lines represent actual thresholds.
}
\label{fig:ths_mult}
\vspace*{-3mm}
\end{figure}

From this viewpoint, 
we confirmed that 
ES-ICP leverages the skewed form in the mean-inverted-index arrays.
Figure~\ref{fig:ths_mult} shows the number of multiplications (Mult) 
along threshold (structural parameter) $v^{[th]}$ 
when the algorithm with $t^{[th]}\!=\!1$ was applied to the 8.2M-sized PubMed,
where this setting was selected to be independent from our $t^{[th]}$.
Figure~\ref{fig:ths_mult}(a) depicts the Mult required before filtering, 
which is the cost for constructing the filter.
If $v^{[th]}\!=\!0$, Mult corresponds to the number of multiplications required by MIVI, 
and Mult is zero if $v^{[th]}\!=\!1$.
The vertical dashed lines represent the actual structural parameters ($v^{[th]}$'s) 
obtained by the estimation algorithm in Section~\ref{sec:strparam}.
Depending on $K$, Mult sharply increased in a range less than the threshold.
Figure~\ref{fig:ths_mult}(b) depicts Mult, including the zero multiplications, 
for the centroids passing through the filter.
The Mult represents the performance of the constructed filter, 
where a better filter shows a lower Mult.
From both the figures,
it is desirable to set the actual threshold at the mean-feature value so that
both the Mults are low.
The vertical dashed lines represent the actual estimated thresholds. 
We know that they were set at the desirable values.

\subsection{ES-ICP as General Algorithm}\label{ssec:general}
\begin{figure}[t]
\begin{center}\hspace*{2mm}
	\subfloat[{\normalsize Feature-value distribution}]{ 
		\psfrag{X}[c][c][0.85]{
			\begin{picture}(0,0)
				\put(0,0){\makebox(0,-5)[c]{Rank/$K$}}
			\end{picture}
		}
		\psfrag{Y}[c][c][0.85]{
			\begin{picture}(0,0)
				\put(0,0){\makebox(0,23)[c]{Mean-feature value}}
			\end{picture}
		}
		\psfrag{P}[c][c][0.80]{$10^{-4}$}
		\psfrag{Q}[c][c][0.80]{$10^{-2}$}
		\psfrag{R}[c][c][0.80]{$1$}
		\psfrag{S}[c][c][0.80]{$10^{2}$}
		\psfrag{T}[c][c][0.80]{$10^{4}$}
		\psfrag{A}[r][r][0.80]{$0$}
		\psfrag{B}[r][r][0.80]{$0.2$}
		\psfrag{C}[r][r][0.80]{$0.4$}
		\psfrag{D}[r][r][0.80]{$0.6$}
		\psfrag{E}[r][r][0.80]{$0.8$}
		\psfrag{F}[r][r][0.80]{$1.0$}
		\psfrag{W}[l][l][0.74]{$1/\sqrt{2}$}
		\psfrag{Z}[r][r][0.78]{$K$}
		\psfrag{J}[r][r][0.72]{$2\!\times\!10^{3}$}
		\psfrag{K}[r][r][0.72]{$5\!\times\!10^{3}$}
		\psfrag{L}[r][r][0.72]{$1\!\times\!10^{4}$}
		\includegraphics[width=42mm]{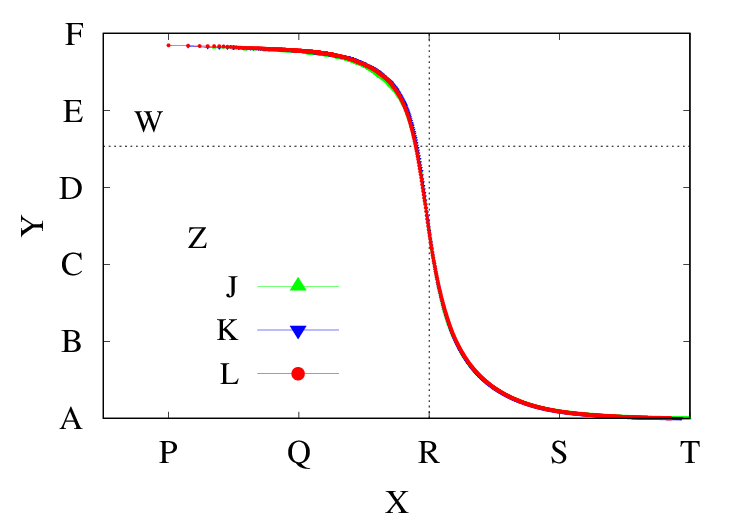}
	}\hspace*{.5mm}
	\subfloat[{\normalsize Probability}]{ 
		\psfrag{X}[c][c][0.85]{
			\begin{picture}(0,0)
				\put(0,0){\makebox(0,-5)[c]{Mean-feature value}}
			\end{picture}
		}
		\psfrag{Y}[c][c][0.85]{
			\begin{picture}(0,0)
				\put(0,0){\makebox(0,23)[c]{Probability}}
			\end{picture}
		}
		\psfrag{P}[c][c][0.80]{$0$}
		\psfrag{Q}[c][c][0.80]{$0.2$}
		\psfrag{R}[c][c][0.80]{$0.4$}
		\psfrag{S}[c][c][0.80]{$0.6$}
		\psfrag{T}[c][c][0.80]{$0.8$}
		\psfrag{U}[c][c][0.80]{$1.0$}
		\psfrag{A}[r][r][0.80]{$0$}
		\psfrag{B}[r][r][0.80]{$0.2$}
		\psfrag{C}[r][r][0.80]{$0.4$}
		\psfrag{D}[r][r][0.80]{$0.6$}
		\psfrag{E}[r][r][0.80]{$0.8$}
		\psfrag{F}[r][r][0.80]{$1.0$}
		\psfrag{Z}[r][r][0.80]{Order}
		\psfrag{J}[r][r][0.72]{100}
		\psfrag{K}[r][r][0.72]{10}
		\psfrag{L}[r][r][0.72]{3}
		\psfrag{M}[r][r][0.72]{2}
		\psfrag{N}[r][r][0.72]{1}
		\includegraphics[width=41mm]{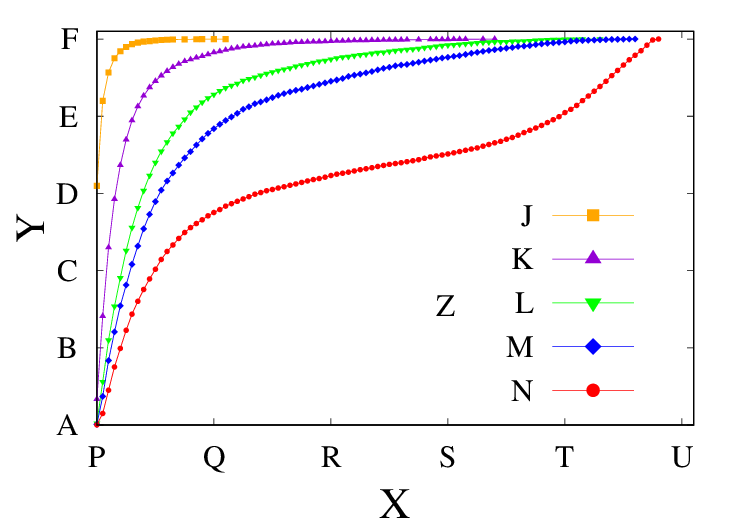}
	}
\end{center}
\vspace*{-2mm}
\caption{\small
(a) Skewed form on mean-feature values in 1M-sized NYT data set.
(b) Probability less than or equal to mean-feature value 
in mean-inverted-index arrays sorted in descending order of 
their values built with K=10\,000.
}
\label{fig:nyt_order_mfv}
\vspace*{-3mm}
\end{figure}

We show that ES-ICP is a general algorithm for large-scale document data sets 
by referring to the experimental results in the 1M-sized NYT data set 
in Section~\ref{ssec:data}.
Figures~\ref{fig:nyt_order_mfv}(a) and (b) respectively show 
the similar skewed form of the mean-feature values
to those in Figs.~\ref{fig:rankmean} and \ref{fig:order_mfv} in the 8.2M-sized PubMed.
That is, each of the many mean-feature vectors had a very large feature value.
These results show the feature-value concentration phenomenon 
and indicate that a cluster is annotated by one or a few dominant terms  
with very large feature values.
Compared with the 8.2M-sized PubMed, the rates of Rank/$K$ over $(1/\sqrt{2})$ and 
the maximum feature values were slightly small, 
and the probability curves also indicated a similar but slightly different tendency 
in the mean-feature-value bias.
This difference probably stems from both the terminology in the newspaper articles (NYT) 
and the corpus of the biomedical literature (PubMed) as well as 
the balance of the dataset size and the amount of vocabulary.

\begin{figure}
\begin{center}\hspace*{2mm}
	\subfloat[{\normalsize Mult before filtering}]{
		\psfrag{X}[c][c][0.85]{
			\begin{picture}(0,0)
				\put(0,0){\makebox(0,-5)[c]{Threshold}}
			\end{picture}
		}
		\psfrag{Y}[c][c][0.85]{
			\begin{picture}(0,0)
				\put(0,0){\makebox(0,28)[c]{\# multiplications}}
			\end{picture}
		}
		\psfrag{P}[c][c][0.78]{$10^{-3}$}
		\psfrag{Q}[c][c][0.78]{$10^{-2}$}
		\psfrag{R}[c][c][0.78]{$10^{-1}$}
		\psfrag{S}[c][c][0.78]{$1$}
		\psfrag{A}[r][r][0.78]{$10^{8}$}
		\psfrag{B}[r][r][0.78]{$10^{9}$}
		\psfrag{C}[r][r][0.78]{$10^{10}$}
		\psfrag{D}[r][r][0.78]{$10^{11}$}
		\psfrag{E}[r][r][0.78]{$10^{12}$}
		\psfrag{F}[r][r][0.78]{$10^{13}$}
		\psfrag{Z}[l][l][0.71]{$K$ $(\times 10^3)$}
		\psfrag{J}[r][r][0.72]{$2$}
		\psfrag{K}[r][r][0.72]{$5$}
		\psfrag{L}[r][r][0.72]{$10$}
		\includegraphics[width=41mm]{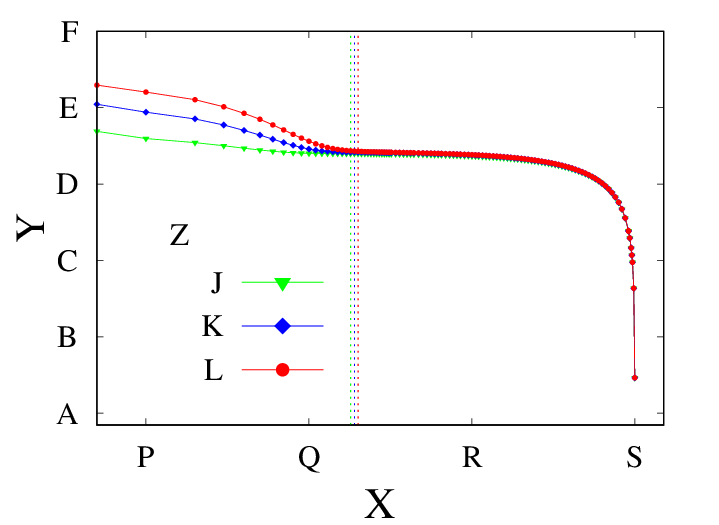}
	}\hspace*{.5mm}
	\subfloat[{\normalsize Mult after filtering}]{ 
		\psfrag{X}[c][c][0.85]{
			\begin{picture}(0,0)
				\put(0,0){\makebox(0,-5)[c]{Threshold}}
			\end{picture}
		}
		\psfrag{Y}[c][c][0.85]{
			\begin{picture}(0,0)
				\put(0,0){\makebox(0,28)[c]{\# multiplications}}
			\end{picture}
		}
		\psfrag{P}[c][c][0.78]{$10^{-3}$}
		\psfrag{Q}[c][c][0.78]{$10^{-2}$}
		\psfrag{R}[c][c][0.78]{$10^{-1}$}
		\psfrag{S}[c][c][0.78]{$1$}
		\psfrag{A}[r][r][0.78]{$10^{8}$}
		\psfrag{B}[r][r][0.78]{$10^{9}$}
		\psfrag{C}[r][r][0.78]{$10^{10}$}
		\psfrag{D}[r][r][0.78]{$10^{11}$}
		\psfrag{E}[r][r][0.78]{$10^{12}$}
		\psfrag{F}[r][r][0.78]{$10^{13}$}
		\psfrag{Z}[l][l][0.71]{$K$ $(\times 10^3)$}
		\psfrag{J}[r][r][0.72]{$2$}
		\psfrag{K}[r][r][0.72]{$5$}
		\psfrag{L}[r][r][0.72]{$10$}
		\includegraphics[width=42mm]{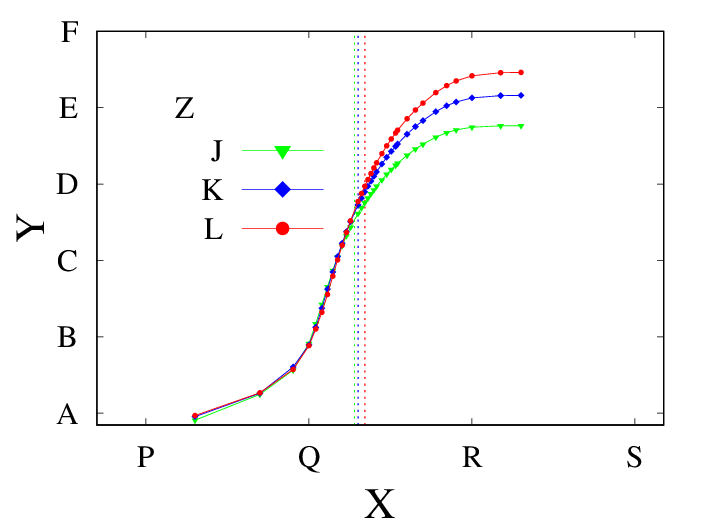}
	}
\end{center}
\vspace*{-2mm}
\caption{\small
Number of multiplications: 
(a) before ES filtering with threshold on horizontal axis
and (b) for centroids passing through 
ES filter when algorithm was applied to 1M-sized NYT.
Vertical dashed lines represent actual thresholds. 
}
\label{fig:ths_mult_nyt}
\vspace*{-3mm}
\end{figure}

\begin{table}[t]
\small
\caption{\small
Performance comparison in NYT}
\label{table:comp_nyt}
\centering
\begin{tabular}{|c|c|c||c|c|c||c|}\hline
\multirow{2}{*}{Algo.} 
& Avg & Avg & \multirow{2}{*}{Inst} &\multirow{2}{*}{BM} &\multirow{2}{*}{LLCM} & Max\\
& Mult & time &                     &                    &                      & MEM\\ \hline
MIVI   & 81.1 & 17.2 & 25.6 & 1.89 & 19.8 & 0.52 \\ \hline
ICP    & 16.4 & 4.30 & 5.77 & 1.38 & 3.99 & 0.52 \\ \hline
CS-ICP & 0.89 & 5.44 & 4.88 & 1.66 & 13.9 & 1.06 \\ \hline
TA-ICP & 12.1 & 6.80 & 6.06 & 10.6 & 20.0 & 1.09 \\ \hline
\end{tabular}
\vspace*{-3mm}
\end{table}

Figures~\ref{fig:ths_mult_nyt}(a) and (b) correspond to 
Figs.~\ref{fig:ths_mult}(a) and (b).
The numbers of multiplications required before and after filtering
for 1M-sized NYT depicted similar curves to those for 
the 8.2M-sized PubMed.
Table~\ref{table:comp_nyt} shows the performance rates of the compared algorithms to 
ES-ICP, which corresponds to Table~\ref{table:comp}.
Comparing the two tables, we see similar tendencies despite different data sets.
Thus, the proposed algorithm 
ES-ICP efficiently performed spherical $K$-means clustering 
for large-scale and high-dimensional sparse document data sets.

\section{Related Work}\label{sec:relate}
This section reviews the algorithms and techniques that are 
closely related to our proposed algorithm in two distinct topics,
Lloyd-type $K$-means clustering algorithms 
and search algorithms with inverted indexes, and 
describes their drawbacks and availability in our settings.

\subsection{K-means Clustering Algorithms}\label{ssec:kmeans}
Many Lloyd-type acceleration algorithms have been studied 
for dense object-feature vectors in a metric space 
\cite{pelleg,kaukoranta,elkan,hamerly_book,bottesch,hattori,xia}.
Their key strategies are using lower bounds 
on distances between an object and centroids  
for skipping unnecessary distance calculations.
Elkan's \cite{elkan} and Hamerly's algorithms \cite{hamerly}
and their variants \cite{drake,ding,rysavy,newling} are 
typical examples based on the former key strategy.
They leverage the distances calculated at the previous iteration and 
the moving distances between identical centroids at two consecutive iterations
(Appendix~\ref{app:triangle}).
However, using the moving distance limits 
the acceleration in the last stage at the iterations
where most centroids slightly move or become invariant.
Among them, 
Drake$^+$ \cite{drake} and Ding$^+$ \cite{ding} make the lower bounds tighter 
using distinct structural parameters proportional to $K$.
Given a large $K$ value like in our setting,
they need a large amount of memory and are not practical. 

Bottesch$^+$ algorithm \cite{bottesch} uses the Cauchy-Schwarz inequality 
to obtain the lower bound on a Euclidean distance 
between data-object- and mean-feature vectors.
To tighten the bound, the algorithm partitions a given feature space into 
disjoint subspaces whose number is an empirical parameter.
The algorithm is inefficient 
if it is na\"{i}vely applied to document data with small values of $(\hat{D}/D)$. 
This is because 
the set-intersection size of the terms in an object and a centroid 
is very small, and then the bound by this algorithm is loosened.
To avoid such inefficiency, 
a subspace might be used where an object-feature vector is contained 
instead of pre-defined subspaces.
This approach, however, is also inefficient 
because of the results of CS-ICP (Section~\ref{ssec:perfcomp}).

Newling$^+$ \cite{newling} and Xia$^+$ algorithms \cite{xia} structure
a centroid set.
Exponion in Newling$^+$ eliminates centroids outside a ball 
defined for each object from distance-calculation candidates. 
The ball's center is a centroid of a cluster 
whose radius is derived based on the triangle inequality.
Xia$^+$ algorithm regards a previous cluster as a ball
and narrows down the distance-calculation candidates 
using a geometric relationship of the balls.
A ball's center is a centroid, and its radius is the distance 
to the farthest object in the cluster.
These algorithms are unsuitable for high-dimensional data sets like document data
because they were originally developed for low-dimensional ones.
Furthermore, they need to store centroid-centroid distances with 
$\mathcal{O}(K^2)$ memory consumption, 
which is prohibited in our setting where large $K$ values are used.

Kaukoranta$^+$ algorithm \cite{kaukoranta} using its ICP filter 
deals with a special case where the moving distance is zero.
Due to its simplicity,
it is used as an auxiliary filter in several algorithms 
\cite{hattori,bottesch,knittel,xia}.
We also use it because it is easily implemented for an inverted index 
in the AFM.

\subsection{Search Algorithms with Inverted Indexes}\label{ssec:search}
A threshold algorithm (TA) \cite{fagin01}
is a well-known classic algorithm with guaranteed optimality.
TA's inverted-index consists of arrays called postings lists each entry 
of which is sorted in descending order of its feature value. 
TA collects document ID's in the postings 
until each of the moving cursors on the postings lists reaches a pre-defined position 
for a stopping condition of its gathering phase.
The strategy needs to execute many conditional branches.
If TA is applied to $K$-means clustering, it does not always reduce the elapsed time.
In fact, the TA-ICP algorithm, inspired by TA in Section~\ref{ssec:perfcomp}, 
required more elapsed time than the ICP algorithm 
without any UBP filter although TA-ICP reduced the multiplications.

A TA-based algorithm \cite{li} was recently reported that 
utilizes a complete and tight stopping condition 
and a traversal strategy with a near-optimal access cost over its inverted index.
Despite the near-optimal access cost,
the algorithm uses many conditional branches like TA 
and also needs additional computational costs to achieve near-optimality.
It does not take into account any negative impacts 
of the performance-degradation factors. 
This casts doubts on the algorithm's actual speed-performance.

Fast algorithms in the document-at-a-time (DAAT) query evaluation 
such as WAND \cite{broder03}\footnote{
The original algorithm is heuristic and out of our scope.}
 and its variants \cite{sding} 
dynamically skip unnecessary similarity calculations.
For the same reason as the TA-based algorithm, 
they suffer from such tail queries as long queries in our settings 
although they may achieve an expected performance for short queries.
This is attributed to their dynamic skipping techniques, 
which are far from the operation in the AFM in Section~\ref{sec:arch}.
Irregularly skipping postings by their conditional branches caused 
many branch mispredictions and cache misses \cite{crane}.

\section{Conclusions}
We proposed the accelerated spherical $K$-means clustering algorithm 
ES-ICP with a large $K$ value, 
which is suitable for large-scale and high-dimensional sparse document data sets
such as the 8.2M-sized PubMed and 1M-sized NYT data sets.
ES-ICP efficiently works with the structured mean-inverted index and 
the two filters of the novel ES and auxiliary ICP 
that are designed using the universal characteristics (UCs) of the document data sets
so as to operate in the architecture-friendly manner (AFM).
We presented the estimation algorithm for the two structural parameters
that partition the structured mean-inverted index into three regions,
resulting in a tight upper bound on the similarity between the object- and mean-feature vectors.
By using the tight upper bound, 
the ES filter achieved high pruning rates in all the iterations.
We confirmed that
ES-ICP performed over 15 times and at least three times faster than the baseline MIVI and 
the compared algorithms using the state-of-the-art techniques
for the 8.2M-sized PubMed, given $K\!=\!80\,000$, respectively.

The following issues remain for future work.
To extend the ranges of applications and platforms of our proposed algorithm,
we will investigate its availability to 
(1) various data sets and features 
and (2) computer systems with GPUs.

\bibliographystyle{IEEEtran}

\vfill

\newpage
\appendices
\section{Full Algorithm of ES-ICP}\label{app:fullalgo}
This section shows 
in Algorithms~\ref{algo:app_assign}, \ref{algo:app_icp_gather1}, and \ref{algo:app_update}
the pseudocode of the full algorithm of ES-ICP in Section~\ref{ssec:icp}.
First, we show the pseudocodes of the assignment step, 
which correspond to Algorithms~\ref{algo:full} and \ref{algo:icp_gather1}.
They are rewritten in the form with scaled feature values \footref{footnote:scaling}.
The scaling is performed to avoid the multiplications for calculating the upper bounds 
on similarities.
Next, we detail the update step at the $r$th iteration in ES-ICP 
in Algorithm~\ref{algo:app_update},
which is not described in Section~\ref{ssec:icp}.

\begin{algorithm}[ht] 
\small
\newcommand{\mf}{\mbox{($mf$\hskip0.1em)}}
\newcommand{\mfH}{\mbox{($mfH$\hskip0.1em)}}
\newcommand{\algto}{\textbf{to}}
\algnewcommand{\LineComment}[1]{\Statex \(\hskip.8em\triangleright\) #1}

  \caption{\hskip.8em Assignment step in ES-ICP algorithm} 
  \label{algo:app_assign}
  \begin{algorithmic}[1]
    \Statex{\textbf{Input:}\hskip.8em $\hat{\mathcal{X}}$,~~$\breve{\mathcal{M}}^{[r-1]}$,~~
					$\breve{\mathcal{M}}^{p[r-1]}$,~~$\left\{\rho_{a(i)}^{[r-1]}\right\}_{i=1}^N$,
					~~$t^{[th]},~v^{[th]}$}
	\Statex{\textbf{Output:}\hskip.8em $\mathcal{C}^{[r]}\!=\!\left\{ C_j^{[r]} \right\}_{j=1}^K$}

	\LineComment{Scale feature values before assignment step starts}
	\ForAll{$\hat{\bm{x}}_i\!=\![(t_{(i,p)},u_{t_{(i,p)}})]_{p=1}^{(nt)_i}\!\in\!\hat{\mathcal{X}}$}
		\label{app_assign:scaling_start}
		\State{$u_{t_{(i,p)}}\gets$ 
			$u_{t_{(i,p)}}\cdot v^{[th]}$}
		\label{app_assign:scaling_end}
		\Comment{Scaling}
	\EndFor

	\LineComment{Calculate similarities in parallel w.\!r.\!t $\hat{\mathcal{X}}$}
	\State{$C_{j}^{[r]}\leftarrow\emptyset$~,~~$j=1,2,\cdots,K$~}
	\ForAll{~$\hat{\bm{x}}_i\!=\![(t_{(i,p)},u_{t_{(i,p)}})]_{p=1}^{(nt)_i}\in \hat{\mathcal{X}}$~~}
	\label{app_assign:out_start}
		\State{$\left\{ \rho_j \right\}_{j=1}^K \gets\! 0$,~
				\hskip.5em~$\rho_{(max)}\!\gets\! \rho_{a(i)}^{[r-1]}$,~
				\hskip.5em~$\mathcal{Z}_i\!\gets\!\emptyset$
		} \label{app_assign:init_start}

		\For{$j\gets 1$ \algto\hskip.6em $K$\hskip.3em}
			\State{$y_{(i,j)}\gets \sum_{t_{(i,p)}\geq t^{[th]}}\hskip.3em (u_{t_{(i,p)}})$}
			\label{app_assign:init_end}
			\Comment{Initializing $y_{(i,j)}$}
		\EndFor

		\LineComment{Gathering phase}
		\If{$xState = 1$ \label{app_assign_icp_switch:g1_start}}
		\Comment{ICP filtering}
			\State{\hspace*{-3.5mm}\small{$\left( \mathcal{Z}_i,\{\rho_j\}_{j=1}^K \right)\!= 
				G_1 (\mathcal{Z}_i,\{(\rho_j,\,y_{(i,j)}\}_{j=1}^K,\,\rho_{(max)},(nMv))$}
			}
		\Else \label{app_assign_icp_switch:g0_start}
			\State{\hspace*{-3.5mm}\small{$\left( \mathcal{Z}_i,\hskip.5em \{\rho_j\}_{j=1}^K \right) = 
				G_0 (\mathcal{Z}_i,\{(\rho_j,\,y_{(i,j)}\}_{j=1}^K,\,\rho_{(max)})$}
				\label{app_assign_icp_switch:g0_end}
			}
		\EndIf

		\LineComment{Verification phase}
		\label{app_assign:r3_start} 
		\For{$(s\gets t_{(i,p)}) \geq t^{[th]}$}
		\Comment{Region 3}
			\State{{\bf for}\hskip.3em~{\bf all}\hskip.3em~$j\in \mathcal{Z}_i$~\hskip.3em{\bf do}
				\hskip1.0em $\rho_j\gets \rho_j+u_s\cdot w_{(s,j)}$
				\label{app_assign:r3_end}
			}
		\EndFor
		\ForAll{$j\in \mathcal{Z}_i$}
		\label{app_assign:verif_start}
			\State{{\bf If}\hskip.3em~$\rho_j > \rho_{(max)}$\hskip.3em~{\bf then}\hskip1.0em
				$\rho_{(max)}\!\leftarrow \rho_j\hskip.6em \mbox{and}\hskip.6em a(i)\leftarrow j$
			} \label{app_assign:verif_end}
		\EndFor

		\State{$C_{a(i)}^{[r]}\leftarrow C_{a(i)}^{[r]}\cup\{\hat{\bm{x}}_i\}$}
		\label{app_assign:out_end}

	\EndFor
  \end{algorithmic}
\end{algorithm}

\begin{algorithm}[ht]
\small
\newcommand{\mf}{\mbox{($mf$\hskip0.1em)}}
\newcommand{\mfH}{\mbox{($mfH$\hskip0.1em)}}
\newcommand{\mfM}{\mbox{($mfM$\hskip0.1em)}}
\newcommand{\algto}{\textbf{to}}
\algnewcommand{\LineComment}[1]{\Statex \(\hskip.8em\triangleright\) #1}

  \caption{\hskip.8em Candidate-gathering function: $G_1$} 
  \label{algo:app_icp_gather1}
  \begin{algorithmic}[1]
    \Statex{\textbf{Input:}\hskip.8em $\hat{\bm{x}}_i$,~~$\breve{\mathcal{M}}^{[r-1]}$,~~
					$t^{[th]},~v^{[th]}$,~~
					$\mathcal{Z}_i$,~~$\{\rho_j,\,y_{(i,j)}\}_{j=1}^K$,~~$\rho_{(max)}$,\par
					\hskip1.5em$(nMv)$  \Comment{(nMv): \# moving centroids}
					}
	\Statex{\textbf{Output:}\hskip.8em $\mathcal{Z}_i$,~~$\{ \rho_j \}_{j=1}^K$}

	\LineComment{Exact partial similarity calculation}
	\For{$(s \leftarrow t_{(i,p)}) < t^{[th]}$ in all term IDs in $\hat{\bm x}_i$\hskip.3em}
	\label{app_icp:r1_start}\Comment{Region 1}
		\For{
			$1\leq q\leq \mfM_s$ \hskip.2em}
			\label{app_icp:innermost1}
			\State{$\rho_{c_{(s,q)}}\gets \rho_{c_{(s,q)}}+u_{s}\cdot v_{c_{(s,q)}}$}
			\label{app_icp:r1_end}
		\EndFor	
	\EndFor

	\For{$(s \leftarrow t_{(i,p)}) \geq t^{[th]}$ in all term IDs in $\hat{\bm x}_i$\hskip.3em}
	\label{app_icp:r2_start}\Comment{Region 2}
		\For{\hskip.2em 
			$1\leq q\leq \mfM_s$ \hskip.2em}
			\label{app_icp:innermost2}
			\State{$\rho_{c_{(s,q)}}\!\gets \rho_{c_{(s,q)}}\!+u_{s}\cdot v_{c_{(s,q)}}$\par
				\hskip1.0em $y_{(i,c_{(s,q)})}\!\gets y_{(i,c_{(s,q)})}\! -u_{s}$
			} \label{app_icp:r2_end}
		\EndFor	
	\EndFor

	\LineComment{Upper-bound calculation}
	\For{\hskip.2em 
			$1\leq j'\leq (nMv)$ \hskip.2em}
	\label{app_icp:gather_start}
		\State{
			$j'$ is transformed to $j$.}
		\label{app_icp:trans}

		\State{$\rho_{j}^{[ub]}\leftarrow \rho_{j}\,+$\,\framebox[8mm][c]{$y_{(i,j)}$}}
		\label{app_icp:ub}
		\Comment{Region 3 (UB): Scaling}

	\LineComment{ES filtering}
		\State{{\bf if}\hskip.6em~$\rho_{j}^{[ub]} > \rho_{(max)}$\hskip.6em~{\bf then}
			\hskip1.2em~$\mathcal{Z}_i \gets \mathcal{Z}_i \cup \{j\}$
		} \label{app_icp:ubp}
		\label{app_icp:gather_end}
	\EndFor
  \end{algorithmic}
\end{algorithm}

Algorithm~\ref{algo:app_assign} shows the assignment-step pseudocode.
Scaling the data-object-feature values at lines \ref{app_assign:scaling_start} to 
\ref{app_assign:scaling_end} is performed once 
after the structural parameters $t^{[th]}$ and $v^{[th]}$ are determined 
in the ES-ICP algorithm.
Algorithm~\ref{algo:app_icp_gather1} shows $G_1$ function in Algorithm~\ref{algo:app_assign}.
Owing to scaling the data-object-feature values in Algorithm~\ref{algo:app_assign}
and mean-feature values at line~\ref{app_update:mean_end} in Algorithm~\ref{algo:app_update} 
(appearing later),
we can obtain the upper bound on similarities by only the addition 
at line~\ref{app_icp:ub} without any multiplications in Algorithm~\ref{algo:app_icp_gather1}.

\begin{algorithm}[t]
\small
\newcommand{\mf}{\mbox{($mf$\hskip0.15em)}}
\newcommand{\mfH}{\mbox{($mfH$\hskip0.15em)}}
\newcommand{\algto}{\textbf{to}}
\newcommand{\Break}{\hskip.5em\textbf{break}}
\algnewcommand{\LineComment}[1]{\Statex \(\hskip.8em\triangleright\) #1}
\algnewcommand{\JustLineComment}[1]{\State \(\triangleright\) #1}
\newcommand{\LineFor}[2]{%
	\State\algorithmicfor\ {#1}\ \algorithmicdo\ {#2} \algorithmicend\ \algorithmic
	}

  \caption{\hskip.8em Update step in ES-ICP algorithm} 
  \label{algo:app_update}
  \begin{algorithmic}[1]
	\Statex{\textbf{Input:}\hskip.8em $\hat{\mathcal{X}}$,\hskip.5em
					$\mathcal{C}^{[r]}\!=\!\left\{ C_j^{[r]}\right\}_{j=1}^K$\hskip.5em,
					\hskip.4em $t^{[th]}$}
    \Statex{\textbf{Output:}\hskip.8em $\breve{\mathcal{M}}^{[r]}$,~~$\breve{\mathcal{M}}^{p[r]}$,~~
					\hskip.3em$\left\{\rho_{a(i)}^{[r]}\right\}_{i=1}^N$,\hskip.8em 
					$[\mfH_s]_{s=t^{[th]}}^D$}

	\ForAll{~$C_j^{[r]}\in \mathcal{C}^{[r]}$~} \label{app_update:outloop}
		\LineComment{(1) Making tentative mean-feature vector ${\bm \lambda}$}
		\State{$\bm{\lambda}\!=\!(\lambda_s)_{s=1}^D\gets \bm{0}$}
		\Comment{Tentative mean-feature vector} \label{app_update:mean_start}

		\ForAll{~$\hat{\bm{x}}_i\!=\!\left[ (t_{(i,p)},u_{t_{(i,p)}})\right]_{p=1}^{(nt)_i}\in C_j^{[r]}$~~}
			\State{$\lambda_{t_{(i,p)}}\gets \lambda_{t_{(i,p)}}+u_{t_{(i,p)}}$}
		\EndFor
		\State{$\bm{\lambda}\gets \bm{\lambda}/|C_j^{[r]}|$,
			\hskip.6em $\bm{\lambda}\gets$ 
			\framebox[25mm][l]{$\bm{\lambda}/(v^{[th]}\cdot \|\bm{\lambda}\|_2)$}}
			\Comment{Scaling}
			\label{app_update:mean_end}

		\LineComment{(2) Calculating similarities $\rho_{a(i)}^{[r]}$}
		\ForAll{~$\hat{\bm{x}}_i\!=\!\left[ (t_{(i,p)},u_{t_{(i,p)}})\right]_{p=1}^{(nt)_i}\in C_j^{[r]}$~~}
			\label{app_update:asssim_start}
			\State{$\rho_{a(i)}^{[r]}\gets \sum_{p=1}^{(nt)_i} u_{t_{(i,p)}}\cdot \lambda_{t_{(i,p)}}$}
			\label{app_update:asssim_end}
		\EndFor

		\LineComment{(3) Constructing mean-inverted indexes}
		\ForAll{~term ID $s$ of $\lambda_s > 0$~~} \label{app_update:inv_start}
			\Comment{Initialize $q\gets 1$}
			\State{$(c_{(s,q)},v_{c_{(s,q)}})\in\breve{\xi}_{s}^{[r]}\subset\breve{\mathcal{M}}^{[r]}
					\gets (j,\lambda_s)$} \Comment{$q\!+\!\!+$}
			\label{app_update:inv_end}
			\If{$s\geq t^{[th]}$ and $\lambda_s < v^{[th]}$}
				\State{$w_{(s,j)}\,(\in \breve{\bm \zeta}_s^{[r]}\subset\breve{\mathcal{M}}^{p[r]})\,\gets \lambda_s$}
				\label{app_update:pinv_end}
			\EndIf
		\EndFor
	\EndFor

	\LineComment{(4) and (5) Structuring $\breve{\xi}_s^{[r]}\in \breve{\mathcal{M}}^{[r]}$}
	\ForAll{~~$[\breve{\xi}_s^{[r]}]_{s=1}^{t^{[th]}-1}\subset\breve{\mathcal{M}}^{[r]}$~~}
	\label{app_update:strR1_start}
		\State{Divide $[(c_{(s,q)},v_{c_{(s,q)}})]_{q=1}^{(mf)_s}$ into two blocks:\par
		\hskip.3em $[(c_{(s,q)},v_{c_{(s,q)}})]_{q=1}^{(mfM)_s}$ and 
				$[(c_{(s,q)},v_{c_{(s,q)}})]_{q=(mfM)_s+1}^{(mf)_s}$\,.
		}
		\label{app_update:strR1_end}
	\EndFor
	\ForAll{~~$[\breve{\xi}_s^{[r]}]_{s=t^{[th]}}^D\subset\breve{\mathcal{M}}^{[r]}$~~}
	\label{app_update:strR2_start}
		\State{Pick up tuples from $[(c_{(s,q)},v_{c_{(s,q)}})]_{q=1}^{(mf)_s}\!\in\breve{\xi}_s^{[r]}$\par
				\hskip.5em that satisfy $v_{c_{(s,q)}}\!\geq v^{[th]}$ and\par
				\hskip.5em  make $[(c_{(s,q)},v_{c_{(s,q)}})]_{q=1}^{(mfH)_s}$ from them.}
		\label{app_update:vth_end}
		\State{Divide $[(c_{(s,q)},v_{c_{(s,q)}})]_{q=1}^{(mfH)_s}$ into two blocks:\par
			\hskip.3em $[(c_{(s,q)},v_{c_{(s,q)}})]_{q=1}^{(mfM)_s}$ and 
				$[(c_{(s,q)},v_{c_{(s,q)}})]_{q=(mfM)_s+1}^{(mfH)_s}$.}
		\label{app_update:strR2_end}
	\EndFor

	\LineComment{Estimating structural parameters (Appendices~\ref{app:objfunct} and \ref{app:est})}
	\If{$r\in\{1,2\}\;$} \label{app_update:est_start}
		\State{$(t^{[th]},v^{[th]})\gets$ {\it EstParams}\,
			($\breve{\mathcal{M}}^{[r]},\hat{\mathcal{X}}$, other arguments) \par
			\hskip1.2em(detailed in Appendix~\ref{app:est})
		}
		\State{Reconstruct $\breve{\mathcal{M}}^{[r]}$ using new $(t^{[th]},v^{[th]})$}
		\label{app_update:est_end}
	\EndIf	
  \end{algorithmic}
\end{algorithm}

Algorithm~\ref{algo:app_update} shows the pseudocode of the update step 
at the $r$th iteration.
The algorithm performs the following five processes 
at lines~\ref{app_update:outloop} to \ref{app_update:strR2_end}.
\begin{enumerate}
\item Making tentative mean-feature vector $\bm{\lambda}$ of
the cluster $C_j^{[r]}$ at lines~\ref{app_update:mean_start} to \ref{app_update:mean_end}.
In particular, at line~\ref{app_update:mean_end}, 
scaling the mean-feature values by $1/v^{[th]}$ is performed.
The $\bm{\lambda}$ with full expression consists of $D$ feature values $\lambda_s$ $(s\!=\!1,2,\cdots,D)$.
\item Calculating the similarity $\rho_{a(i)}^{[r]}$ of the $i$th object to
the centroid of cluster $C_{a(i)}^{[r]}$ to which the object belongs 
at lines~\ref{app_update:asssim_start} and \ref{app_update:asssim_end}.\hskip.5em
The $\rho_{a(i)}^{[r]}$ is utilized as the similarity threshold in the gathering phase
at the assignment step at the next $(r\!+\! 1)$ iteration, 
which is for determining whether the $i$th object satisfies the condition of 
$\rho_{j}^{[ub]} > \rho_{(max)}$ ($\rho_{(max)}\gets \rho_{a(i)}^{[r]}$ at the initialization)
for the ES filter, e.g.,
at line~\ref{app_icp:ubp} in Algorithm~\ref{algo:app_icp_gather1}.
\item Constructing mean-inverted index $\breve{\mathcal M}^{[r]}$ 
and partial mean-inverted index $\breve{\mathcal M}^{p[r]}$
using the obtained $\bm{\lambda}$ at lines~\ref{app_update:inv_start} to \ref{app_update:pinv_end}.
$\breve{\mathcal M}^{p[r]}$, whose element is $w_{(s,j)}$, is utilized for exact similarity calculations 
in Region 3 at the assignment step in Algorithm~\ref{algo:app_assign}.
$\breve{\mathcal M}^{p[r]}$ consists of $(D -\!t^{[th]} +\!1)$ inverted-index arrays $\breve{\zeta}_s^{[r]}$
whose lengths are $K$.
The $j$th element in $\breve{\zeta}_s$ is mean-feature values 
$w_{(s,j)}$ if $w_{(s,j)}\!<\!v^{[th]}$, otherwise 0.
Due to this full expression, 
the $j$th element can be directly accessed by using the centroid ID $j$ as the key.
\item Structuring inverted-index arrays for the ES filter, which are 
$\breve{\xi}_s^{[r]}\in \breve{\mathcal M}^{[r]}$ $(t^{[th]}\!\leq\! s\! \leq\! D)$
at lines~\ref{app_update:strR2_start} to \ref{app_update:vth_end}.
Since mean-feature values of $v_{c_{(s,q)}}\!\geq\! v^{[th]}$ are used 
for calculating exact partial similarities in Region 2, 
the tuples containing such values are placed from the top of $\breve{\xi}_s^{[r]}$ 
to the $(mfH)_s$th entry.
\item Structuring inverted-index arrays $\breve{\xi}_s^{[r]}$ for the ICP filter 
at lines~\ref{app_update:strR1_start} to \ref{app_update:strR1_end} in Region 1
and at line~\ref{app_update:strR2_end} in Region 2.
\end{enumerate}

Furthermore, at the first and second iterations,
two structural parameters $t^{[th]}$ and $v^{[th]}$ are estimated by using {\em EstParams} function 
in Algorithm~\ref{algo:2params} (appearing later) 
and described in Appendices~\ref{app:objfunct} and \ref{app:est}.
\hskip.3em$\breve{M}^{p[r]}$ is constructed after the structural-parameter estimation 
and then $\breve{M}^{[r]}$ is updated.

\section{Objective Function in Section~\ref{sec:strparam}}\label{app:objfunct}
The objective function is expressed by Eq.~(\ref{eq:objfunct_all}) in Section~\ref{sec:strparam}.
Its key term is the probability 
that the upper bound $\rho^{[ub]}(i)$ is more than or equal to
the similarity threshold $\rho_{a(i)}$, given structural-parameter values of $v_h^{[th]}$ and
$s'$ for $t^{[th]}$.
To deal with the discrete similairties $\rho_j$ $(j\!=\!1,\cdots,K)$ 
of $\bm{x}_i$ to the $j$th centroid, 
with a continuous relaxation, 
we introduce the probability density function of $\rho$, $f(\rho)$, 
as the similarity distribution w.r.t. $\rho$.
Then, we derive the probability of
$\mbox{Prob}\,(\rho^{[ub]}(i)\!\geq \rho_{a(i)}\,;s',h)$
in Eq.~(\ref{eq:prob_unpruned1})
where $s'$ denotes a candidate of $t^{[th]}$ and $h$ represents 
$v_h^{[th]}$ that is the $h$th candidate of $v^{[th]}$.
To do this, we assume the following three. 
\begin{enumerate}
%
\item 
The probability density function $f(\rho)$ is approximated 
by a distribution function in an exponential family: 
\begin{eqnarray}
&&f(\rho)=\lambda_{0}\cdot e^{-\lambda_{0}(\rho -\bar{\rho}_i)} \label{eq:asm1}\\
&&\hspace*{8mm}(1/\lambda_0)=\bar{\rho}_i\, ,\nonumber
\end{eqnarray}
where $\bar{\rho}_i$ denotes the average of similarities of ${\bm x}_i$ 
to all the centroids.
We apply the condition of $\rho\! \geq\! \bar{\rho}_i$ to $f(\rho)$.
Then, the following equation holds:
\begin{equation}
\mbox{Prob}(\rho\geq\bar{\rho}_i) = \int_{\bar{\rho}_i}^{\infty}f(\rho)d\rho =\frac{1}{e}\, .
\label{eq:prob-e}
\end{equation}
%
\item We precisely model a probability density function
of $f(\rho,\rho\geq\bar{\rho}_i)$ in the range of
$\rho\geq\bar{\rho}_i$ through the following expression:
\begin{equation}
f(\rho,\rho\!\geq\!\bar{\rho}_i) = \mbox{Prob}(\rho\geq \bar{\rho}_i)
\cdot f(\rho\,|\,\rho\geq\bar{\rho}_i)\,  \label{eq:asm2-1}
\end{equation}
\begin{equation}
f(\rho\,|\, \rho\geq\bar{\rho}_i) =\lambda\cdot e^{-\lambda(\rho -\bar{\rho}_i)}\, ,
\label{eq:asm2-2}
\end{equation}
where $(1/\lambda)$ denotes the expected value of $f(\rho,\rho\geq\bar{\rho}_i)$ 
and $\lambda \neq \lambda_0$.
%
\item Similarity-upper-bound distribution $f^{[ub]}(\rho,\rho\!\geq\!\bar{\rho}_i)$ 
is expressed by the parallel translation of $f(\rho,\rho\!\geq\!\bar{\rho}_i)$ 
by $\Delta\bar{\rho}(i;s',h)$ as 
\begin{align}
&f^{[ub]}(\rho-\Delta\bar{\rho}(i;s',h),\rho\geq\bar{\rho}_i)
=f(\rho,\rho\geq\bar{\rho}_i) \label{eq:asm3} \\
&\Delta\bar{\rho}(i;s',h) = \bar{\rho}^{[ub]}(i\,;s',h) -\bar{\rho}_i \notag\, , 
\end{align}
where the second equation is the same as Eq.~(\ref{eq:Deltabarrho}).
\end{enumerate}

\vskip\baselineskip
Since an object belongs to a distinct cluster, i.e., hard clustering,
the centroid with a similarity more than or equal to $\rho_{a(i)}$ is 
only one of the cluster to which the object belongs.
Then, the following equation holds:
\begin{eqnarray}
\mbox{Prob}(\rho > \rho_{a(i)})
&=& \int_{\rho_{a(i)}}^{\infty} f(\rho,\rho\!\geq\!\bar{\rho}_i)\,d\rho \nonumber\\
&=& \mbox{Prob}(\rho\!\geq\!\bar{\rho}_i)\cdot\, e^{-\lambda (\rho_{a(i)} -\bar{\rho}_i)} 
\nonumber\\
&=& \frac{1}{K}\, , 
\label{eq:singleclst}
\end{eqnarray}
where $\rho_{a(i)}$ denotes the similarity of the $i$th object 
to the centroid of the cluster which the object belongs to.
For simplicity, the superscript of $[r]$ for $\rho_{a(i)}^{[r]}$,
which represents the $r$th iteration, is omitted.
Using Eqs.~(\ref{eq:asm2-1}), (\ref{eq:asm2-2}), and (\ref{eq:singleclst}), 
$\lambda$ is expressed as 
\begin{equation}
\lambda = \frac{\log (K/e)}{\rho_{a(i)} -\bar{\rho}_i}\, .
\label{eq:ab}
\end{equation}
The probability in Eqs.~(\ref{eq:cost3}) and (\ref{eq:prob_unpruned1}) 
is expressed by
\begin{eqnarray}
&&\mbox{Prob}\,(\rho^{[ub]}(i)\!\geq \rho_{a(i)}\,;s',h)\\ \nonumber
&&\quad = \int_{\rho_{a(i)}-\Delta\bar{\rho}(i;s',h)}^{\infty} 
f(\rho,\rho\!\geq\!\bar{\rho}_i)\,d\rho \\ \nonumber
&&\quad = \left( \frac{1}{K}\right)\,
\left( \frac{K}{e} \right)^{\frac{\Delta\bar{\rho}(i;s',h)}{\rho_{a(i)} -\bar{\rho}_i}} \, .
\label{eq:prob}
\end{eqnarray}
Then, the expected value of the partial-similarity upper bound 
in Region~3 in Eq.~(\ref{eq:cost3}) 
is rewritten as 
\begin{equation}
(\tilde{\varphi}3)_{(s',h)} = \sum_{i=1}^N\, (ntH)_{(i;s')}\cdot 
\left( \frac{K}{e} \right)^{\frac{ \Delta\bar{\rho}(i;s',h)}{\rho_{a(i)} -\bar{\rho}_i} } \, .
\label{eq:re_cost3_h}
\end{equation}
$\Delta\bar{\rho}(i;s',h)$ is obtained by substituting 
$\bar{\rho}_i$ and $\bar{\rho}^{[ub]}(i;s',h)$ in Eqs.~(\ref{eq:avgrho}) and (\ref{eq:avgrhoub})
into Eq.~(\ref{eq:asm3}).
\begin{equation}
\bar{\rho}_i = \sum_{p=1}^{(nt)_i}\,\left( u_{t_{(i,p)}}
\cdot\frac{1}{K}\sum_{q=1}^{(mf)_s}\, v_{c_{(s,q)}}\right)\, ,\quad
s = t_{(i,p)}\, ,\label{eq:avgrho}
\end{equation}
where $t_{(i,p)}$ and $c_{(s,q)}$ respectively denote 
the term ID appeared at the $p$th position in $\hat{\bm{x}}_i$ and 
the mean ID (cluster ID) appeared at the $q$th position in 
the inverted-index-array with the $s$th term ID $\breve{\xi}_s$
(see Table~\ref{table:nota}).
\begin{align}
&\bar{\rho}^{[ub]}(i;s',h) =
\sum_{s=1}^{s'-1}\left( u_s\cdot \frac{1}{K}
\sum_{q=1}^{(mf)_s}\, v_{c_{(s,q)}}\right) \qquad\qquad\notag \\
&+
\sum_{s= s'}^D\! u_s\cdot\frac{1}{K}
\left( \sum_{q=1}^{(mfH)_s}
v_{c_{(s,q)}}\! +v_h^{[th]}\cdot (mfL)_s \right)
\label{eq:avgrhoub}
\end{align}
where $(mfL)_s = (mf)_s\!-\!(mfH)_{(s,v_h^{[th]})}$ and
$u_s$ is $u_{t_{(i,p)}}$ if $u_{t_{(i,p)}}$ exits for $t_{(i,p)} \geq t^{[th]}$, 
$0$ otherwise.

We can obtain the objective function by 
substituting Eqs.~(\ref{eq:cost1}), (\ref{eq:cost2}), and (\ref{eq:simple_cost3})
into Eq.~(\ref{eq:objfunct2}) as
\begin{eqnarray*}
J(s',v_h^{[th]}) &=& \sum_{s=1}^{s'-1} (df)_s\cdot (mf)_s \nonumber \\
&+&\!\sum_{s=s'}^D (df)_s\cdot (mfH)_{(s,v_h^{[th]})} \nonumber\\
&+&\!\sum_{i=1}^N\, (ntH)_{(i,s')}\cdot 
\left( \frac{K}{e} \right)^{\frac{ \Delta\bar{\rho}(i;s',h)}{\rho_{a(i)} -\bar{\rho}_i} } \, .
\end{eqnarray*}

\section{Parameter Estimation in Section~\ref{sec:strparam}}\label{app:est}
This section describes our proposed efficient and practical algorithm for 
simultaneously estimating $t^{[th]}$ and $v^{[th]}$.
For the parameter estimation,
we suppose that an appropriate $t^{[th]}$ exists near $D$ 
on the grounds of the universal characteristics (UCs) in Section~\ref{sec:dchar}, i.e., 
the skewed forms of both the document frequency ($df$) and mean frequency ($mf$) 
with respect to term ID sorted in ascending order of $df$ 
in Figs.~\ref{fig:dchar} and \ref{fig:mfdf},
and $v^{[th]}$ exists in the short range from the top of each inverted-index array
$\breve{\xi}_s$ due to the skewed form of the mean-feature-values in the array.
This allows us to narrow down a search space for the two parameters.

The estimation algorithm minimizes the objective function  
in Eq.~(\ref{eq:objfunct2}) that is rewritten as 
\begin{eqnarray*}
&& ( t^{[th]},v^{[th]} ) = 
\argmin_{\substack{v_h^{[th]}\in V^{[th]}\\ s_{(min)}\leq s'\leq D}}
\left(\, J(s',v_h^{[th]})\,\right)\\
&& J(s',v_h^{[th]}) = 
\tilde{\phi}_{(s',h)} = (\phi 1)_{s'} +(\phi 2)_{(s',h)} +(\tilde{\phi}3)_{(s',h)}\, ,\qquad
\end{eqnarray*}
where $s'$ denotes the term ID that is a candidate of $t^{[th]}$ and 
decremented one by one from $D$ to the pre-determined minimum term ID of $s_{(min)}$
and $h$ represents the $h$th candidate of $v^{[th]}$, i.e., $v_h^{[th]}$.
Based on the presumption that $t^{[th]}$ exists near $D$,
the estimation algorithm efficiently calculates $\tilde{\phi}_{(s',h)}$ 
for each $v_h^{[th]}$\footnote{
This process for each $v_h^{[th]}$ is executed in parallel processing.
}
by exploiting a recurrence relation with respect to $s'$ expressed by
\begin{eqnarray}
\tilde{\phi}_{(s';h)} &=& \tilde{\phi}_{(s'+1;h)} 
-(df)_{s'}\cdot (mfL)_{(s';v_h^{[th]})}\nonumber\\
&&\hskip 3.5em +(\tilde{\phi}3)_{(s';h)} -(\tilde{\phi}3)_{(s'+1;h)}
\label{eq:recurrent}\\
&&\hspace*{-8mm}\tilde{\phi}_{(D+1;h)} = \sum_{s=1}^D (df)_s\cdot (mf)_s = \phi
\nonumber \\
&&\hspace*{-8mm}(\tilde{\phi}3)_{(D+1;h)} = 0 \, ,\nonumber
\end{eqnarray}
where $(mfL)_{(s';v_h^{[th]})}\!=\!(mf)_{s'} -(mfH)_{(s';v_h^{[th]})}$, i.e.,
$(mfL)_{(s';v_h^{[th]})}$ denotes 
the number of centroids whose feature values are lower than $v_h^{[th]}$.

Moreover, $(\tilde{\phi}3)_{(s';h)}$ itself in Eq.~(\ref{eq:recurrent}) is
represented with a recurrence relation.
We rewrite the approximate number of multiplications with respect to the $i$th object 
in Eq.~(\ref{eq:simple_cost3}) as   
\begin{equation}
(\bar{\phi}3)_{(i,s';h)} = (ntH)_{(i,s')}\cdot\left( \frac{K}{e} \right)^{
 \frac{\bar{\rho}^{[ub]}(i,s';h) -\bar{\rho}_i}
	{\rho_{a(i)} -\bar{\rho}_i} } \, .
\label{eq:app_simple_cost3}
\end{equation}

We consider two cases of the $i$th object contains the term whose ID is $s'$,
i.e., $s'\!\in\! \{t_{(i,p)}\}_{p=1}^{(nt)_i}$, and the other.
In the latter case, 
$(\bar{\phi}3)_{(i,s';h)}$ is invariant from $(\bar{\phi}3)_{(i,s'+1;h)}$ as
\begin{equation}
(\bar{\phi}3)_{(i,s';h)} = (\bar{\phi}3)_{(i,s'+1;h)} \, . \label{eq:cost3i_same}
\end{equation}
In the former case, we derive the recurrence relation while focusing on 
$(ntH)_{(i,s')}$ and $\bar{\rho}^{[ub]}(i,s';h)$ in Eq.~\ref{eq:app_simple_cost3}.
First, the following equation with respect to $(ntH)_{(i,s')}$ holds: 
\begin{equation}
(ntH)_{(i,s')} = (ntH)_{(i,s'+1)}+1 \label{eq:nthi}\, .
\end{equation}
Next, $\bar{\rho}^{[ub]}(i,s';h)$ is expressed by
\begin{equation}
\bar{\rho}^{[ub]}(i,s') = \bar{\rho}^{[ub]}(i,s'\!+\!1)
+ \Delta\bar{v}_{(i,s')}\cdot u_{s'}
\label{eq:avgrhoub_recurrent}
\end{equation}
\begin{eqnarray}
\Delta\bar{v}_{(i,s')}
&=&\frac{1}{K}\sum_{q=(mfH)_{s'}+1}^{(mf)_{s'}}\hspace*{-2mm}(v_h^{[th]}\! -v_{c_{(s,q)}})
\\ \nonumber
&+&\! \left( 1-\frac{(mf)_{s'}}{K} \right) v_h^{[th]}\, ,
\label{eq:avgrhoub_recurrent2}
\end{eqnarray}
where $u_{s'}$ denotes $u_{t_{(i,p)}=s'}$
and $\Delta\bar{v}_{(i,s')}$ is the average difference between $v_h^{[th]}$ and 
the exact mean-feature values in Region 3 at the $s'$th term ID
and $h$ is omitted for simplicity.
Then, $\bar{\rho}^{[ub]}(i,s';h)$ is expressed by 
\begin{equation}
(\tilde{\phi} 3)_{(i,s';h)} = (ntH)_{(i,s')}\cdot 
\left(\frac{K}{e}\right)^{
\frac{\bar{\rho}^{[ub]}(i,s')-\bar{\rho}_i}{\rho_{a(i)}-\bar{\rho}_i} }
\end{equation}
\begin{eqnarray}
(\tilde{\phi} 3)_{(i,s';h)}
&=&\left(\frac{K}{e}\right)^{\gamma(i,s')}\\ \nonumber
&\times&
\left\{
(\tilde{\phi} 3)_{(i,s'+1;h)}
+\left(\frac{K}{e}\right)^{\frac{\bar{\rho}^{[ub]}(i,s'+1)-\bar{\rho}_i}{\rho_{a(i)}-\bar{\rho}_i}}
\right\}
\label{eq:cost3i_recurrence}\\
\end{eqnarray}
\begin{equation}
\gamma(i,s') = \frac{\Delta\bar{v}_{(i,s')}\cdot u_{s'}}{\rho_{a(i)}-\bar{\rho}_i}\, ,
\label{eq:gamma_i}
\end{equation}
where the values in the square bracket are already obtained at the $(s'\!+\!1)$th term ID.
The boundary conditions are set as 
\begin{eqnarray}
&&(\tilde{\phi}3)_{(i,D+1;h)}\! = 0 \\ 
&&(ntH)_{(i,D+1)}\! = 0 \\
&&\bar{\rho}^{[ub]}(i,D\!+\!1) = \bar{\rho}_i\, .
\end{eqnarray}
By applying Eqs.~(\ref{eq:cost3i_same}) and (\ref{eq:cost3i_recurrence}) 
to Eq.~(\ref{eq:recurrent}), we can make the recurrence relation.

\begin{algorithm}[t]
\small
\newcommand{\mf}{\mbox{($mf$\hskip0.1em)}}
\newcommand{\mfH}{\mbox{($mfH$\hskip0.1em)}}
\newcommand{\algto}{\textbf{to}}
\algnewcommand{\LineComment}[1]{\Statex \(\hskip.8em\triangleright\) #1}

  \caption{\hskip.8em Parameter-estimation function: {\it EstParams}} 
  \label{algo:2params}
  \begin{algorithmic}[1]
    \Statex{\textbf{Input:}\hskip.5em $\breve{\mathcal{M}}$,\;$\hat{\mathcal{X}}$,
			\;$\breve{\mathcal{X}}^p$,
			\;$V^{[th]}=\left\{ v_1^{[th]},\cdots, v_{|V^{[th]}|}^{[th]}\right\}$,
			\;$s_{(min)}$
	}
	\Statex{\textbf{Output:}\hskip.8em $\left( t^{[th]},\hskip.4em v^{[th]} \right)$}

	\LineComment{Initialization}
	\State{$\phi\!=\!\sum_{s=1}^{D} (df)_s\cdot (mf)_s$}
	\ForAll{$\hat{\bm{x}}_i\in \hat{\mathcal{X}}$}
		\State{ $\bar{\rho}_i \gets (1/K)\sum_{p=1}^{(ntH)_i}\sum_{q=1}^{(mf)_{s=t_{(i,p)}}}
		(v_{c_{(s,q)}}\cdot u_{t_{(i,p)}})$}
	\EndFor
	
	\LineComment{Parallel processing}
	\ForAll{$v_h^{[th]}\in V^{[th]}$}
	\label{algo:app_est_outerstart}
		\State{$\tilde{\phi}_{(D+1,h)}\gets \phi$}
		\State{$(ntH)_{(i,D+1)}\gets 0,\hskip.5em
			\Delta v_{(i,D+1)}\gets 0\hskip 1.2em \mbox{for}\hskip.8em 1\leq i\leq N$} 

		\For{$s'\gets D$\hskip.5em \algto\hskip.5em $s_{(min)}$}
		\label{algo:app_est_midstart}
			\Comment{$s'$: $t^{[th]}$ candidate}

			\ForAll{$i=o_{(s',q')}\in \breve{\bm{\eta}}_{s'}\subset \breve{\mathcal{X}}^p$}
			\label{algo:app_est_innerstart}
			\State{$\tilde{\phi}_{(s',h)} \gets \tilde{\phi}_{(s'+1,h)}
					-(ntH)_{(i,s'+1)}(K/e)^{\gamma_{(i,s'+1)}}$}
			\State{$\Delta v_{(i,s')} \gets \Delta v_{(i,s'+1)}$\par
					\hskip\algorithmicindent\hspace*{18mm} 
					$+\sum_{q=(mfH)_s'+1}^{(mf)_{s'}}(v_h^{[th]}-v_{c_{(s',q)}})$\par
					\hskip\algorithmicindent\hspace*{18mm} 
					$+(K\!-\!(mf)_{s'})\cdot v_h^{[th]} $}
			\State{$\Delta \bar{v}_{(i,s')} \gets\Delta v_{(i,s')}/K$}
			\State{$\gamma_{(i,s')}\gets
				(\Delta\bar{v}_{(i,s')}\cdot \breve{u}_i)/(\rho_{a(i)}\!-\!\bar{\rho}_i)$}

			\State{$\tilde{\phi}_{(s',h)} \gets \tilde{\phi}_{(s',h)}
					+(ntH)_{(i,s')}(K/e)^{\gamma_{(i,s')}}$}
			\label{algo:app_est_innerend}
			\EndFor 

			\State{$\tilde{\phi}_{(s',h)}\gets \tilde{\phi}_{(s',h)} 
					-(df)_{s'}\cdot (mfL)_{(s';v_h^{[th]})}$}
			\State{$J(s',v_h^{[th]}) = \tilde{\phi}_{(s',h)}$}
			\label{algo:app_est_midend}
		\EndFor

		\State{$(t_h^{[th]},v_h^{[th]})\gets \argmin_{s_{(min)}\leq s' \leq D}~
			\left( J(s',v_h^{[th]}) \right)$}
		\label{algo:app_est_outerend}
	\EndFor

	\State{$(t^{[th]},v^{[th]})\gets \argmin_{1\leq h \leq |V^{[th]}|}~
		\left( J(t_h^{[th]},v_h^{[th]}) \right)$}
	\label{algo:app_est_end}
  \end{algorithmic}
\end{algorithm}

\begin{table}[t]
\small
\centering
\caption{Notation for {\it EstParams} function}\label{table:2params}
\begin{tabular}{|c|p{58mm}|}\hline
Symbol & \qquad Description and Definitions \\ \hline\hline
\multirow{5}{*}{$J(s',v_h^{[th]})$}
& Function that returns the approximation\\ 
& number of the multiplications\\
& \hspace*{1mm}{$s'$: Term ID as $t^{[th]}$ candidate}\\
& \hspace*{1mm}{$v_h^{[th]}$: Mean-feature-value as $v^{[th]}$ candidate}\\
& \hspace*{3mm}{$v_h^{[th]}$ can be represented with $h$.}\\
\hline
$V^{[th]}$
& Set of $v_h^{[th]}$, $h=1,\cdots,|V^{[th]}|$\\
\hline
\multirow{2}{*}{$(\tilde{\phi})_{(s',h)}$}
& Approximate number of multiplications for\\
& similarity calculations, given $s'$ and $h$\\
\hline
$(df)_s$ & Document frequency of $s$th term ID\\
\hline
$(mf)_s$ & Mean frequency of $s$th term ID\\
\hline
\multirow{2}{*}{$s_{(min)}$}
& Minimum term ID of candidates ($s'$) of $t^{[th]}$,\\
& $s_{(min)}\leq s' \leq D$\\
\hline
\multirow{6}{*}{$\breve{\mathcal{X}}^p$}
& Partial inverted-index of objects, whose\\
& column is object-tuple array ${\breve{\bm \eta}}_s,~s\geq s_{(min)}$\\
& \hspace*{1mm}{${\breve{\bm \eta}}_s=[(o_{(s,q')},\breve{u}_{o_{(s,q')}})]_{q'=1}^{(df)_s}$}\\
& \hspace*{3mm}{$o_{(s,q')}$: $q'$th object ID appeared in $\breve{\bm \eta}_s$}\\ 
& \hspace*{3mm}{$\breve{u}_{o_{(s,q')}}$: $q'$th feature value appeared in $\breve{\bm \eta}_s$}\\ 
& \hspace*{3mm}{$(df)_s$: Document frequency of $s$th term ID}\\ 
\hline
\hline
$\bar{\rho}_i$ 
& Average similarity of $i$th object to centroids\\
\hline
\multirow{2}{*}{$(ntH)_{(i,t^{th})}$}
& Number of terms in $i$th object  whose ID is\\
& higher than or equal to given $t^{[th]}$\\ 
\hline
\multirow{3}{*}{$(mfH)_{(s;v_h^{[th]})}$} 
& Mean frequency of $s$th term ID where\\
& $v_{(s,q)}\geq v_h^{[th]}$,~given $v_h^{[th]}$,\\
& where $q$ is a position in $\breve{\xi}_s$\\
\hline
$(mfL)_{(s;v_h^{[th]})}$
& {\small $(mfL)_{(s;v_h^{[th]})} = (mf)_s -(mfH)_{(s;v_h^{[th]})}$}\\
\hline
\end{tabular}
\end{table}

\vskip\baselineskip
When using the recurrence relation,
we need to access a data-object- and a mean-feature value at the $s$th term ID.
Although the mean-feature value is easily accessed through the mean-inverted-index array 
$\breve{\xi}_s$ in the inverted index $\breve{\mathcal{M}}$,
it is difficult to selectively access only the data-object feature value at the $s$th term ID 
through the current data-object structure of $\hat{\mathcal{X}}$ at low computational cost.
To overcome the difficulty in the architecture-friendly manner (AFM) in Section~2,
we introduce a partial object-inverted-index $\breve{\mathcal{X}}^{p}$
for the term IDs from $s_{(min)}$ to $D$ with low memory consumption.
By using the partial object-inverted-index, 
we can access both the feature values at the $s$th term ID
without using conditional branches causing many mispredictions  
or loading the large array of the object data set, 
resulting in efficiently calculating $(\tilde{\phi}3)_{(i,s';h)}$.

Algorithm~\ref{algo:2params} and Table~\ref{table:2params} 
show a pseudocode of the practical algorithm {\em EstParams} and its notation, respectively.
Given the $v^{[th]}$ candidate $v_h^{[th]}\!\in\! V^{[th]}$,
at lines~\ref{algo:app_est_outerstart} to \ref{algo:app_est_outerend}, 
the tuple of $(t_h^{[th]},v_h^{[th]})$ is obtained for each $v_h^{[th]}$, 
where $t_h^{[th]}$ denotes the $s'$ that minimizes $J(s',v_h^{[th]})$.
At line~\ref{algo:app_est_end},
$(t^{[th]},v^{[th]})$ is determined as the tuple that minimizes $J(t_h^{[th]},v_h^{[th]})$
among all the tuples.
At lines~\ref{algo:app_est_midstart} to \ref{algo:app_est_midend},
the recurrent relation is used with the initial state of $s'\!=\!D$.
By using the inverted-index-array $\breve{\eta}_{s'}$ in the data-object inverted index 
$\breve{\mathcal{X}}^{p}$,
we can efficiently access the object-feature value $\breve{u}_i$ having the term ID of $s'$
and calculate the approximate number of the multiplications $\tilde{\phi}_{(s',h)}$
at lines~\ref{algo:app_est_innerstart} to \ref{algo:app_est_midend}.

To confirm the effectiveness of the parameter-estimation algorithm {\em EstParams} 
in Algorithm~\ref{algo:2params}, 
we incorporated it to 
our proposed $K$-means clustering algorithm ES-ICP in Section~\ref{sec:prop} 
and applied ES-ICP with $K\!=\!80\,000$ to the 8.2M-sized PubMed data set 
in Section~\ref{sec:exp}.
We executed {\em EstParams} twice, 
at the first and the second iteration, in ES-ICP.
The purpose of its execution at the first iteration is only to reduce the elapsed time 
in the second iteration.
The clustering result at the first iteration strongly depends on its initial setting and 
some centroids often change their positions significantly.
At the second iteration, {\em EstParams} determines the parameter values that are utilized 
at the successive iterations.

\begin{figure}[t]
\begin{center}
	\psfrag{X}[c][c][0.88]{
		\begin{picture}(0,0)
			\put(0,0){\makebox(0,-8)[c]{Threshold on mean-feature value ($\times 10^{-2}$)}}
		\end{picture}
	}
	\psfrag{Y}[c][c][0.88]{
		\begin{picture}(0,0)
			\put(0,0){\makebox(0,16)[c]{\# multiplications ($\times 10^{11}$)}}
		\end{picture}
	}
	\psfrag{P}[c][c][0.9]{$2$}
	\psfrag{Q}[c][c][0.9]{$3$}
	\psfrag{R}[c][c][0.9]{$4$}
	\psfrag{S}[c][c][0.9]{$5$}
	\psfrag{T}[c][c][0.9]{$6$}
	\psfrag{A}[r][r][0.9]{$4$}
	\psfrag{B}[r][r][0.9]{$5$}
	\psfrag{C}[r][r][0.9]{$6$}
	\psfrag{D}[r][r][0.9]{$7$}
	\psfrag{E}[r][r][0.9]{$8$}
	\psfrag{F}[r][r][0.9]{$9$}
	\psfrag{J}[r][r][0.8]{Actual}
	\psfrag{K}[r][r][0.8]{Approximate}
	\includegraphics[width=52mm]{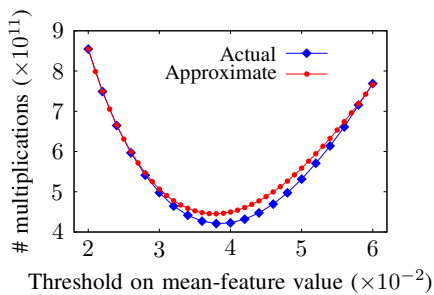}
\end{center}
\caption{\small
Comparison of the approximate and actual number of the multiplications 
along threshold $v_h^{[th]}$ on the mean-feature value 
for 8.2M-sized PubMed with K=80\,000. 
The other threshold $s'$ on term ID at each point differs from others in general.
}
\label{fig:comp_multvth}
\end{figure}

\begin{figure}[t]
\begin{center}
	\psfrag{X}[c][c][0.88]{
		\begin{picture}(0,0)
			\put(0,0){\makebox(0,-8)[c]{Threshold on mean-feature value ($\times 10^{-2}$)}}
		\end{picture}
	}
	\psfrag{Y}[c][c][0.88]{
		\begin{picture}(0,0)
			\put(0,0){\makebox(0,16)[c]{\# multiplications ($\times 10^{11}$)}}
		\end{picture}
	}
	\psfrag{P}[c][c][0.9]{$2$}
	\psfrag{Q}[c][c][0.9]{$3$}
	\psfrag{R}[c][c][0.9]{$4$}
	\psfrag{S}[c][c][0.9]{$5$}
	\psfrag{A}[r][r][0.9]{$4$}
	\psfrag{B}[r][r][0.9]{$6$}
	\psfrag{C}[r][r][0.9]{$8$}
	\psfrag{D}[r][r][0.9]{$10$}
	\psfrag{E}[r][r][0.9]{$12$}
	\psfrag{F}[r][r][0.9]{$14$}
	\psfrag{J}[r][r][0.8]{$t^{[th]}\!\times\!10^{-5}=1.22$}
	\psfrag{K}[r][r][0.8]{$1.24$}
	\psfrag{L}[r][r][0.8]{$1.26$}
	\psfrag{M}[r][r][0.8]{$1.28$}
	\psfrag{N}[r][r][0.8]{$1.30$}
	\psfrag{O}[r][r][0.8]{$1.32$}
	\includegraphics[width=52mm]{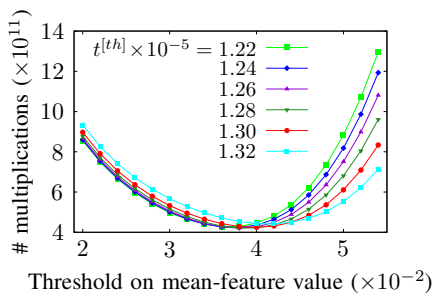}
\end{center}
\caption{\small
Actual number of the multiplications 
along threshold on the mean-feature value with various $t^{[th]}$
for 8.2M-sized PubMed with K=80\,000. 
}
\label{fig:act_multvth}
\end{figure}

We compared the approximate number of the multiplications 
obtained by {\em EstParams} 
with the corresponding actual number calculated by ES-ICP, 
where $t^{[th]}$ and $v^{[th]}$ were set at the values estimated at the second iteration.
Figure~\ref{fig:comp_multvth} shows the comparison with the approximate and actual number 
of the multiplications along $v_h^{[th]}$ at which $t_h^{[th]}$ 
at line~\ref{algo:app_est_outerend} in Algorithm~\ref{algo:2params} 
was a different value in most of the $v_h^{[th]}$ ranges 
except in the several small $v_h^{[th]}$ values.
The parameters in {\em EstParams} were set as 
$s_{(min)}\!=\! 1.22\!\times\! 10^5$ and 
$0.020\leq v_h^{[th]}\leq 0.060$ by $0.001$ step.
We observed that the approximate number of the multiplications agreed with 
the actual number in all the range and each value of $v_h^{[th]}$.
The minimum number of the multiplications in both the estimated and the actual ones
was observed at the identical value of $0.038$.
Figure~\ref{fig:act_multvth} shows the actual number of the multiplications 
when $t^{[th]}$ was set at the various fixed values 
from $1.22\!\times\! 10^5$ to $1.32\!\times\! 10^5$.
Comparing the approximate number of the multiplications in Fig.~\ref{fig:comp_multvth}
with the actual ones in Fig.~\ref{fig:act_multvth}, 
we notice that the approximate number corresponds to the lowest envelop curve of 
the actual numbers. 
Thus, our parameter-estimation algorithm {\em EstParams} can find appropriate values 
for $t^{[th]}$ and $v^{[th]}$ simultaneously.

\section{Ablation Study}\label{app:ablation}
We analyze the contributions of components in our proposed algorithm ES-ICP
to its performance.
We focus on the main filter ES since 
the performance of only the auxiliary filter ICP is evaluated in Section~\ref{ssec:perfcomp}.
The ES filter exploits the tight upper bounds on the similarities 
obtained by using the three regions 
divided by the two parameters of $t^{[th]}$ and $v^{[th]}$.
Depending on whether or not the two parameters are used, 
we prepared three algorithms without ICP in addition to the baseline MIVI. 

The first algorithm was ES that employed both the two parameters.
The second was ThV that utilized only the parameter of $v^{[th]}$ 
that is estimated by $t^{[th]}\!=\!0$
\footnote{
ThV's $v^{[th]}$ was 0.032 while ES's 0.038
for 8.2M-sized PubMed with $K\!=\!80\,000$.
}.
ThV calculated the upper-bound similarities based on the $v^{[th]}$ in the range of all the term IDs.
The upper bounds were looser than those of ES 
since the exact partial similarities in Region 1 of ES were replaced with
the corresponding upper bounds.
For calculating the exact similarities of unpruned centroids, 
ThV needed mean-inverted index $\breve{M}^{p[r]}$ 
of the memory size of $\{ D\times K\times \mbox{sizeof(double)} \}$ bytes.
Note that ThV's memory size differs from that of ES-ICP, 
which is $\{ (D\!-\!t^{[th]}\!+\!1)\times K\times \mbox{sizeof(double)} \}$ bytes.
The last was ThT that utilized the parameter of $t^{[th]}$ that is estimated by $v^{[th]}\!=\!1.0$ 
\footnote{
ThT's $t^{[th]}$ was 140\,904 while ES's 128\,090
for 8.2M-sized PubMed with $K\!=\!80\,000$.
}.
The upper bounds  $\rho_{(j;i)}^{[ub]'}$ corresponding to those in Eq.~(\ref{eq:ub})
are expressed by
$\rho_{(j;i)}^{[ub]'}\!=\!(\rho 1')_{(j;i)} +\|{\bm x}_i^{p'}\|_1$,
where $\|{\bm x}_i^{p'} \|$ and $(\rho 1')_{(j;i)}$ respectively denote 
the partial $L_1$ norm of the $i$th object 
and the exact partial similarity of the object to the $j$th centroid based on ThT's $t^{[th]}$.

\begin{figure}[t]
\small
\begin{center}
	\subfloat[{\normalsize Number of multiplications}]{ 
		\hspace*{2mm}
		\psfrag{X}[c][c][0.95]{
			\begin{picture}(0,0)
				\put(0,0){\makebox(0,-6)[c]{Iterations}}
			\end{picture}
		}
		\psfrag{Y}[c][c][0.95]{
			\begin{picture}(0,0)
				\put(0,0){\makebox(0,26)[c]{\# multiplications}}
			\end{picture}
		}
		\psfrag{A}[c][c][0.85]{$0$}
		\psfrag{B}[c][c][0.85]{$20$}
		\psfrag{C}[c][c][0.85]{$40$}
		\psfrag{D}[c][c][0.85]{$60$}
		\psfrag{H}[r][r][0.85]{$10^6$}
		\psfrag{I}[r][r][0.85]{$10^{8}$}
		\psfrag{J}[r][r][0.85]{$10^{10}$}
		\psfrag{K}[r][r][0.85]{$10^{12}$}
		\psfrag{L}[r][r][0.85]{$10^{14}$}
		\psfrag{P}[r][r][0.72]{MIVI}
		\psfrag{Q}[r][r][0.72]{ICP}
		\psfrag{R}[r][r][0.72]{ES}
		\psfrag{S}[r][r][0.75]{ThV}
		\psfrag{T}[r][r][0.75]{ThT}
		\psfrag{U}[r][r][0.70]{ES-ICP}
		\includegraphics[width=41mm]{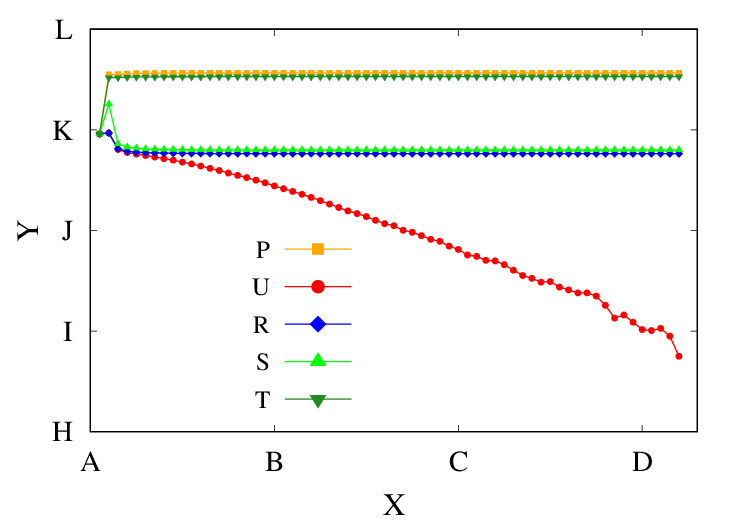}
	}\hspace*{1.0mm}
	\subfloat[{\normalsize Compl. pruning rate}]{
		\psfrag{X}[c][c][0.98]{
			\begin{picture}(0,0)
				\put(0,0){\makebox(0,-6)[c]{Iterations}}
			\end{picture}
		}
		\psfrag{Y}[c][c][0.90]{
			\begin{picture}(0,0)
				\put(0,0){\makebox(0,30){Compl. prunig rate}}
			\end{picture}
		}
		\psfrag{A}[c][c][0.85]{$0$}
		\psfrag{B}[c][c][0.85]{$20$}
		\psfrag{C}[c][c][0.85]{$40$}
		\psfrag{D}[c][c][0.85]{$60$}
		\psfrag{H}[r][r][0.85]{$10^{-8}$}
		\psfrag{I}[r][r][0.85]{$10^{-6}$}
		\psfrag{J}[r][r][0.85]{$10^{-4}$}
		\psfrag{K}[r][r][0.85]{$10^{-2}$}
		\psfrag{L}[r][r][0.85]{$1$}
		\psfrag{P}[r][r][0.68]{ES}
		\psfrag{Q}[r][r][0.68]{ThV}
		\psfrag{R}[r][r][0.68]{ThT}
		\psfrag{S}[r][r][0.64]{ES-ICP}
		\includegraphics[width=41mm]{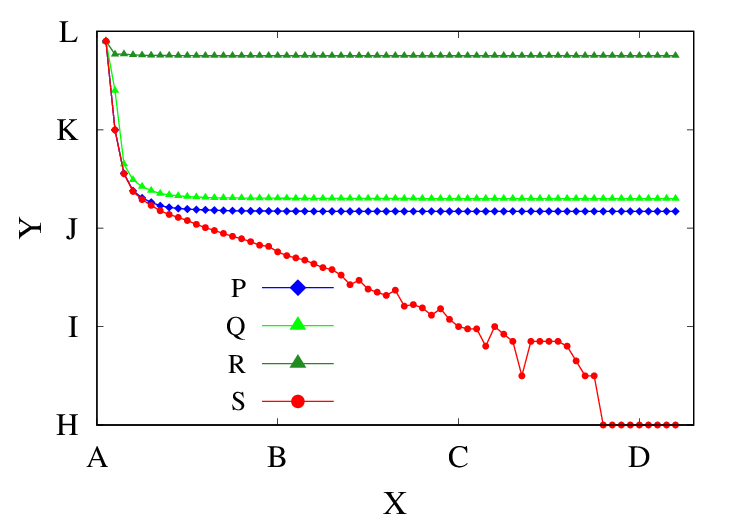}
	}
\end{center}
\caption{\small
Ablation study in 8.2M-sized PubMed data set with K=80\,000:
(a) Number of multiplications and 
(b) Complementary pruning rate
along iterations until convergence.
}
\label{fig:abl}
\end{figure}

Figures~\ref{fig:abl}(a) and (b) respectively show the number of the multiplications (Mult) 
and the complementary pruning rate (CPR)
\footnote{The values of CPR were bounded below by $1\times 10^{-8}$.}
along iterations until convergence 
when the algorithms were applied to the 8.2M-sized PubMed with $K\!=\! 80\,000$.
ES and ThV successfully reduced both the multiplications and the CPR 
at the early stage in the iterations while ThT barely did.
This observation shows that ThT had much looser upper bounds than ES and ThV.
We know that the ES's pruning performance was mainly supported by the upper bounds 
based on the parameter $v^{[th]}$.

\begin{figure}[t]
\begin{center}
		\psfrag{X}[c][c][0.95]{
			\begin{picture}(0,0)
				\put(0,0){\makebox(0,-8)[c]{Iterations}}
			\end{picture}
		}
		\psfrag{Y}[c][c][0.90]{
			\begin{picture}(0,0)
				\put(0,0){\makebox(-2,12)[c]{Elapsed time ($\times 10^3$ sec)}}
			\end{picture}
		}
		\psfrag{A}[c][c][0.85]{$0$}
		\psfrag{B}[c][c][0.85]{$20$}
		\psfrag{C}[c][c][0.85]{$40$}
		\psfrag{D}[c][c][0.85]{$60$}
		\psfrag{H}[r][r][0.85]{$0$}
		\psfrag{I}[r][r][0.85]{$1$}
		\psfrag{J}[r][r][0.85]{$2$}
		\psfrag{K}[r][r][0.85]{$3$}
		\psfrag{L}[r][r][0.85]{$4$}
		\psfrag{PPPPPPPP}[r][r][0.8]{MIVI}
		\psfrag{QQQQQQQQ}[r][r][0.8]{ThT}
		\psfrag{RRRRRRRR}[r][r][0.8]{ThV}
		\psfrag{SSSSSSSS}[r][r][0.8]{ES}
		\psfrag{TTTTTTTT}[r][r][0.8]{ES-ICP}
		\includegraphics[width=70mm]{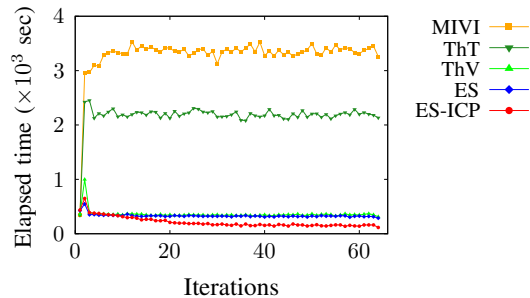}
\end{center}
\caption{\small
Elapsed time along iterations until convergence 
in 8.2M-sized PubMed data set with K=80\,000.
}
\label{fig:time_abl}
\end{figure}

\begin{table}[t]
\small
\caption{\small
Ablation study of proposed algorithm
}
\label{table:abl}
\centering
\begin{tabular}{|c|c|c||c|c|c||c|}\hline
\multirow{2}{*}{Algo.} 
& Avg & Avg  &\multirow{2}{*}{Inst}&\multirow{2}{*}{BM}&\multirow{2}{*}{LLCM}&Max \\
& Mult& time &                     &                   &                     &MEM \\ \hline
ES         & 3.793  & 1.546 & 1.782 & 1.658 & 3.205 & 0.998 \\ \hline
ThV  & 4.832  & 1.676 & 1.925 & 1.956 & 3.461 & 5.777 \\ \hline
ThT  & 119.4  & 10.16 & 13.39 & 7.173 & 11.54 & 0.5094 \\ \hline
\end{tabular}
\end{table}

Figure~\ref{fig:time_abl} shows the elapsed time that each algorithm required in the same setting 
in Fig.~\ref{fig:abl}.
Table \ref{table:abl} shows the corresponding performance rates of the compared algorithms to ES-ICP.
ES and ThV had the similar characteristics to each other in terms of
the elapsed time and the performance-degradation factors of Inst, BM, and LLCM.
ThV needed the high memory capacity for its partial mean-inverted index $\breve{M}^{p[r]}$.
We consider that the difference between these two algorithms and ES-ICP comes from 
the number of unpruned centroids in Fig.~\ref{fig:abl}.
ThT had the inferior performance to the others except the maximum memory usage.
This shows that the parameter $t^{[th]}$ contributed to not the pruning performance
but rather the low memory usage for the partial mean-inverted index $\breve{\mathcal M}^{p[r]}$
of $\{ (D -t^{[th]} +1)\times K\times \mbox{sizeof(double)} \}$-byte capacity.
Compared with MIVI in Table~\ref{table:comp}, 
ThT slightly decreased Mult and Inst with the aid of its filter 
while it increased BM and LLCM.
This was caused by ThT's poor filter passing many centroids in Fig.~\ref{fig:abl}(b).
Then, many branch mispredictions occurred in  
the conditional branch at the gathering phase, which judges whether or not $\rho_{(j;i)}^{[ub]}$
is larger than $\rho_{(max)}$
and mean-feature values in the partial mean-inverted index were loaded many times
for the exact similarity calculations.

In reference, the comparisons of actual performance of 
ES-ICP, ES, ThV, and ThT are shown in Tables~\ref{table:app_abl}, \ref{table:app_perf_abl}, 
\ref{table:app_abl_nyt}, and ~\ref{table:app_perf_abl_nyt}.
Tables~\ref{table:app_abl} and  \ref{table:app_perf_abl} correspond to those 
when the algorithms were applied to the 8.2M-sized PubMed with $K\!=\!80\,000$
and Tables~\ref{table:app_abl_nyt} and ~\ref{table:app_perf_abl_nyt} correspond to those 
when the algorithms were applied to the 1M-sized NYT data set with $K\!=\!10\,000$.

Consequently, 
the parameters of $v^{[th]}$ and $t^{[th]}$ mainly contribute to the high pruning performance 
and the low memory usage, respectively.
The proposed algorithm ES-ICP with the two filters of ES and ICP efficiently worked 
at high speed and with low memory consumption from the early to the last stage in the iterations.

\section{Algorithm Performance in Section~\ref{sec:arch}}\label{app:arch}
The algorithm performance of MIVI, DIVI and Ding$^{+}$ in Section~\ref{sec:arch}
are shown as the rates to the baseline algorithm MIVI.
This section shows the actual values of their performance.

Tables~\ref{table:app_ivfd-ding} shows  
the average values per iteration of the number of multiplications and the elapsed time (sec)
when the algorithms were executed at an identical initial state
in the 8.2M-sized PubMed with $K\!=\!80\,000$ 
until the convergence where they needed 64 iterations.
Table~\ref{table:app_perf_ivfd-ding} shows their perf results (total amounts),
where the numbers of branches and last-level-cache (LLC) loads are added 
in addition to the evaluation items in Table~\ref{table:ivfd} in Section~\ref{sec:arch}.
We notice two facts in the perf results.
One is that Ding$^+$ and DIVI needed many LLC-loads and 
respectively caused LLC-load misses of around 99\% and 80\% of the loads.
The other is that Ding$^+$ failed around 10\% of conditional-branch predictions
while MIVI did only 0.04\%.
These caused the increase of elapsed time of Ding$^+$ and DIVI.

\section{Compared Algorithms in Section~\ref{ssec:perfcomp}}\label{app:compalgo}
This section details TA-ICP and CS-ICP in Section~\ref{ssec:perfcomp}.
Both the algorithms employ the mean-inverted index that is partitioned 
by the structural parameter $t^{[th]}$ (threshold on term IDs) into two regions
like the proposed algorithm ES-ICP.
The differences from ES-ICP are in their main upper-bound-based pruning (UBP) filters
 and data structures related to the filters.

\subsection{TA-ICP}
TA-ICP is characterized by using 
(1) the UBP filter every an individual object and 
(2) two mean-inverted indexes for moving centroids and all the centroids.
Each array of the mean-inverted index is sorted in descending order of the mean-feature values.
Remind that ES-ICP's array is not sorted but classified in Section~\ref{ssec:frame}.
The UBP filter of TA-ICP is designed, inspired by the threshold algorithm (TA)
in Fagin$^+$ and Li$^+$ algorithms.
This filter utilizes a threshold ($v_{(ta)i}^{[th]}$) on mean-feature values, 
which is defined every the $i$th object by Eq.~(\ref{eq:ta_th}) as follows.
\begin{equation}
v_{(ta)i}^{[th]} = \rho_{(max)}/\|{\bm x}_i \|_1 \, ,
\nonumber
\end{equation}
where $\|{\bm x}_i\|_1$ and $\rho_{(max)}$ denote 
the $L_1$-norm of the $i$th object-feature vector and 
the similarity of $i$th object to the centroid to which the object belongs 
at the last iteration. 
Due to the threshold for the individual object,
it is difficult to structure the mean-inverted index when incorporating 
the auxiliary ICP filter unlike ES-ICP.
TA-ICP employs the two mean-inverted indexes for moving and all the centroids. 

\begin{algorithm}[t] 
\small
\newcommand{\algto}{\textbf{to}}
\algnewcommand{\LineComment}[1]{\Statex \(\hskip.8em\triangleright\) #1}
\algnewcommand{\LineCommentConti}[1]{\Statex \hskip1.8em #1}

  \caption{\hskip.8em Assignment step in TA-ICP} 
  \label{algo:app_taicp}
  \begin{algorithmic}[1]
    \State{\textbf{Input:}\hskip.8em $\hat{\mathcal{X}}$,\,\,$\breve{\mathcal{M}}^{[r-1]}$,\,\,
			$\breve{\mathcal{M}}^{p[r-1]}$,\,\,$\left\{\rho_{a(i)}^{[r-1]}\right\}_{i=1}^N$,\,\,
			$t^{[th]}$,\par
			\hskip1.5em $\breve{\mathcal{M}}_{mv}^{[r-1]}$
			}
	\State{\textbf{Output:}\hskip.8em $\mathcal{C}^{[r]}\!=\!\left\{ C_j^{[r]} \right\}_{j=1}^K$}

	\State{$C_{j}^{[r]}\leftarrow\emptyset$~,~~$j=1,2,\cdots,K$~}

	\LineComment{Calculate similarities in parallel w.\!r.\!t $\hat{\mathcal{X}}$}
	\ForAll{~$\hat{\bm{x}}_i\!=\![(t_{(i,p)},u_{t_{(i,p)}})]_{p=1}^{(nt)_i}\in \hat{\mathcal{X}}$~~}
	\label{taicp:out_start}
		\State{${\mathcal{Z}}_i\!\gets\!\emptyset$\hskip.1em,\hskip.2em
				$\left\{\rho_j \right\}_{j=1}^K \!\gets\! 0$\hskip.1em,\hskip.2em
				$\rho_{(max)}\!\gets\! \rho_{a(i)}^{[r-1]}$, \par
				$\left\{ y_{(i,j)}\right\}_{j=1}^K \!\gets\! 
					\sum_{t_{(i,p)}\geq t^{[th]}}u_{t_{(i,p)}}$}
				\label{taicp:init}

		\State{\hskip.3em\framebox[35mm][l]{
				$v_{(ta)i}^{[th]} \gets \rho_{(max)}/\| {\bm x}_i\|_1$}
				\hskip.3em}
			\label{taicp:thv}
		\Comment{Individual threshold}

		\LineComment{Gathering phase}
		\If{$xState = 1$} 
		\label{icp_switch:gta1_start}
		\Comment{Object satisties Eq.~(\ref{eq:objstate}).}
			\State{args\hskip.3em$\gets \mathcal{Z}_i,
				\,\{\rho_j,y_{(i,j)}\}_{j=1}^K,\,\rho_{(max)},\,v_{(ta)i}^{[th]}$}
			\State{$\left( \mathcal{Z}_i,\,\{\rho_j\}_{j=1}^K \right) = G_{(ta)1}(\mbox{args})$}
			\label{icp_switch:gta1_end}
		\Else
		\label{icp_switch:gta0_start}
			\State{$\left( \mathcal{Z}_i,\,\{\rho_j\}_{j=1}^K \right) = G_{(ta)0} (\mbox{args})$}
			\label{icp_switch:gta0_end}
		\EndIf

		\LineComment{Exact-similarity calculation for unpruned centroids}
		\For{$(s\gets t_{(i,p)}) \geq t^{[th]}$}
		\label{taicp:calc_start}
			\ForAll{$j\in \mathcal{Z}_i$}
				\If{\framebox[22mm][l]{\hskip.2em $w'_{(s,j)}< v_{(ta)i}^{[th]}$}\hskip.2em}
					\label{taicp:branch}
					\State{$\rho_j\gets \rho_j+u_s\cdot w'_{(s,j)}$}
					\label{taicp:calc_end}
				\EndIf
			\EndFor
		\EndFor

		\LineComment{Verification phase: The same as that
			in Algorithm~\ref{algo:app_assign}.}

		\State{$C_{a(i)}^{[r]}\leftarrow C_{a(i)}^{[r]}\cup\{\hat{\bm{x}}_i\}$}
		\label{taicp:out_end}

	\EndFor
  \end{algorithmic}
\end{algorithm}

Algorithms~\ref{algo:app_taicp} and \ref{app:taicp_gather0} show the pseudocodes 
of the assignment step in TA-ICP.
At line~\ref{taicp:thv} in Algorithm~\ref{algo:app_taicp},
the individual threshold is set.
In the gathering phase from lines~\ref{icp_switch:gta1_start} to \ref{icp_switch:gta0_end}, 
the candidate-centroid-ID set ${\mathcal Z}_i$ for exact similarity calculations is made 
by the UBP and ICP filters.
The exact similarity calculations for the centroids with IDs in ${\mathcal Z}_i$ are performed 
at lines~\ref{taicp:calc_start} to \ref{taicp:calc_end}.
$w'_{(s,j)}$ at line~\ref{taicp:branch} denotes the mean-feature value of the $j$th centroid 
in the $s$th array in the partial mean-inverted index that 
is different from the ES-ICP's counter part in including all the mean-feature values.

\begin{algorithm}[thpb]
\small
\newcommand{\mf}{\mbox{($mf$\hskip.1em)}}
\newcommand{\mfH}{\mbox{($mfH$\hskip.1em)}}
\newcommand{\algto}{\textbf{to}}
\algnewcommand{\LineComment}[1]{\Statex \(\hskip.8em\triangleright\) #1}

  \caption{\hskip.8em Candidate-gathering function: $G_{(ta)0}$} 
  \label{app:taicp_gather0}
  \begin{algorithmic}[1]
    \Statex{\textbf{Input:}\hskip.8em $\hat{\bm{x}}_i$,\,\, $\breve{\mathcal{M}}^{[r-1]}$,\,\,
					$t^{[th]}$,\par
					\hskip1.5em $\mathcal{Z}_i$,\,\,$\{\rho_j,y_{(i,j)}\}_{j=1}^K$,\,\,
					$\rho_{(max)}$,\,\,$v_{(ta)i}^{[th]}$}
	\Statex{\textbf{Output:}\hskip.8em $\mathcal{Z}_i$,\,\,$\{ \rho_j \}_{j=1}^K$}

	\LineComment{Exact partial similarity calculation}
	\For{$(s \leftarrow t_{(i,p)}) < t^{[th]}$ for $p$ in $\hat{\bm x}_i$\hskip.3em}
	\Comment{Region 1}
	\label{taicp_g0:r1_start} 
		\For{$1\leq q\leq \mf_s$}\label{taicp_g0:innermost1}
			\State{$\rho_{c_{(s,q)}}\gets \rho_{c_{(s,q)}}+u_s\cdot v_{c_{(s,q)}}$}
			\label{taicp_g0:r1_end}
		\EndFor	
	\EndFor

	\For{$(s \leftarrow t_{(i,p)}) \geq t^{[th]}$ for $p$ in $\hat{\bm x}_i$\hskip.3em}
	\Comment{Region 2}
	\label{taicp_g0:r2_start}
		\For{$1\leq q\leq \mf_s$} \label{taicp_g0:innermost2}
			\State{\framebox[45mm][l]{\hskip.2em{\bf If}\hskip.5em$v_{c(s,q)} < v_{(ta)i}^{[th]}$ 
				\hskip.5em{\bf then} \hskip.5em break\hskip.3em}}
			\label{taicp_g0:branch}

			\State{$\rho_{c_{(s,q)}}\gets \rho_{c_{(s,q)}}+u_{s}\cdot v_{c_{(s,q)}}$}
			\State{$y_{(i,c_{(s,q)})}\gets y_{(i,c_{(s,q)})} -u_s$}
			\label{taicp_g0:r2_end}
		\EndFor	
	\EndFor

	\LineComment{Gathering phase}
	\For{$1\leq j \leq K$}
	\label{taicp_g0:gather_start}
		\State{\framebox[38mm][l]{\hskip.2em{\bf If}\hskip.5em$\rho_j = 0$ 
				\hskip.5em{\bf then} \hskip.5em continue\hskip.3em}}
		\State{$\rho_{j}^{[ub]}\leftarrow \rho_{j} +\,v_{(ta)i}^{[th]}\cdot y_{(i,j)}$}
		\Comment{UB calculation}
		\If{$\rho_{j}^{[ub]} > \rho_{(max)}$\hskip.3em} \label{taicp_g0:ubp}
			\Comment{UBP filter}
			\State{$\mathcal{Z}_i \gets \mathcal{Z}_i \cup \{j\}$}
			\label{taicp_g0:gather_end}
		\EndIf
	\EndFor

  \end{algorithmic}
\end{algorithm}

Algorithm~\ref{app:taicp_gather0} shows 
the pseudocode of the candidate-gathering function $G_{(ta)0}$ 
at line~\ref{icp_switch:gta0_end} in Algorithm~\ref{algo:app_taicp}. 
$G_{(ta)0}$ is used for the object that does not satisfy the condition in Eq.~(\ref{eq:objstate}).
When calculating the exact partial similarity of the $i$th object to the centroids 
whose ID is $c_{(s,q)}$,
$G_{(ta)0}$ judges whether its feature value satisfies the condition of $v_{c_{(s,q)}}\geq v_{(ta)i}^{[th]}$
or not at line~\ref{taicp_g0:branch}.
After the exact partial-similarity calculations in Regions 1 and 2 (defined by $v_{(ta)i}^{[th]}$),
the upper bound on the similarity to the centroid whose exact partial-similarity is not zero 
is calculated.
If the exact partial-similarity is zero, the upper bound is smaller than or equal to $\rho_{(max)}$
from the definition of $v_{(ta)i}^{[th]}\!=\!\rho_{(max)}/\|{\bm x}_i\|_1$.
In the gathering phase, the centroid-IDs of the centroids passing through the TA filter 
are collected in ${\mathcal Z}_i$.

The function $G_{(ta)1}$ at line~\ref{icp_switch:gta1_end} in Algorithm~\ref{algo:app_taicp} 
has the identical structure to $G_{(ta)0}$.
The difference is that  $G_{(ta)1}$ uses the moving-centroid mean-inverted index 
instead of $\breve{\mathcal{M}}^{[r-1]}$.
For this difference,
$(mf)_s$ at lines \ref{taicp_g0:innermost1} and \ref{taicp_g0:innermost2} in $G_{(ta)0}$ 
is replaced with $(mfM)_s$ in $G_{(ta)1}$. 

\subsection{CS-ICP}
CS-ICP is characterized by using 
(1) the Cauchy-Schwarz inequality that is applied to mean-feature vectors in a subspace 
of an individual object-feature vector, 
(2) an additional mean-inverted index that stores squared mean-feature values,
and (3) no threshold on the mean-feature values unlike ES-ICP and TA-ICP.

\begin{algorithm}[t] 
\small
\newcommand{\mf}{\mbox{($mf$\hskip0.1em)}}
\newcommand{\mfH}{\mbox{($mfH$\hskip0.1em)}}
\newcommand{\algto}{\textbf{to}}
\algnewcommand{\LineComment}[1]{\Statex \(\hskip.8em\triangleright\) #1}
\algnewcommand{\LineCommentConti}[1]{\Statex \hskip1.8em #1}

  \caption{\hskip.8em Assignment step in CS-ICP} 
  \label{algo:app_csicp}
  \begin{algorithmic}[1]
    \Statex{\textbf{Input:}\hskip.8em $\hat{\mathcal{X}}$,~~$\breve{\mathcal{M}}^{[r-1]}$,~~
			$\breve{\mathcal{M}}^{p[r-1]}$,
			$\left\{\rho_{a(i)}^{[r-1]}\right\}_{i=1}^N$,
			$\left\{\| {\bm x}_i^p\|_2 \right\}_{i=1}^N$,\par
			\hskip1.5em $\breve{\mathcal{M}}_{sq}^{p[r-1]}$,
			$t^{[th]}$
			}
	\Statex{\textbf{Output:}\hskip.8em $\mathcal{C}^{[r]}\!=\!\left\{ C_j^{[r]} \right\}_{j=1}^K$}

	\State{$C_{j}^{[r]}\leftarrow\emptyset$~,~~$j=1,2,\cdots,K$~}

	\LineComment{Calculate similarities in parallel w.\!r.\!t $\hat{\mathcal{X}}$}
	\ForAll{~$\hat{\bm{x}}_i\!=\![(t_{(i,p)},u_{t_{(i,p)}})]_{p=1}^{(nt)_i}\in \hat{\mathcal{X}}$~~}
	\label{csicp:out_start}
		\State{$\left\{ \rho_j \right\}_{j=1}^K \gets\! 0$,\hskip.5em
				$\mathcal{Z}_i\!\gets\!\emptyset$\hskip.1em, \hskip.5em
				$\rho_{(max)}\!\gets\! \rho_{a(i)}^{[r-1]}$} \label{csicp:init}

		\LineComment{Gathering phase}

		\If{$xState = 1$} 
		\Comment{Object satisties Eq.~(\ref{eq:objstate}).}
		\label{icp_switch:gcs1_start}
			\State{$\left( \mathcal{Z}_i,\hskip.5em \{\rho_j\}_{j=1}^K \right) = 
				G_{(cs)1} (\mathcal{Z}_i,\{\rho_j\}_{j=1}^K,\,\rho_{(max)})$}
			\label{icp_switch:gcs1_end}
		\Else
		\label{icp_switch:gcs0_start}
			\State{$\left( \mathcal{Z}_i,\hskip.5em \{\rho_j\}_{j=1}^K \right) = 
				G_{(cs)0} (\mathcal{Z}_i,\{\rho_j\}_{j=1}^K,\,\rho_{(max)})$}
			\label{icp_switch:gcs0_end}
		\EndIf

		\LineComment{Exact-similarity calculation for unpruned centroids:\par
			The same algorithm structure as that in Algorithm~\ref{algo:app_assign}
		}

		\LineComment{Verification phase: The same as that in Algorithm~\ref{algo:app_assign}}

		\State{$C_{a(i)}^{[r]}\leftarrow C_{a(i)}^{[r]}\cup\{\hat{\bm{x}}_i\}$}
		\label{csicp:out_end}

	\EndFor
  \end{algorithmic}
\end{algorithm}

\begin{algorithm}[t] 
\small
\newcommand{\mf}{\mbox{($mf$\hskip0.1em)}}
\newcommand{\mfH}{\mbox{($mfH$\hskip0.1em)}}
\newcommand{\algto}{\textbf{to}}
\algnewcommand{\LineComment}[1]{\Statex \(\hskip.8em\triangleright\) #1}

  \caption{\hskip.8em Candidate-gathering function: $G_{(cs)0}$} 
  \label{app:csicp_gather0}
  \begin{algorithmic}[1]
    \Statex{\textbf{Input:}\hskip.8em $\hat{\bm{x}}_i$,\,\, $\breve{\mathcal{M}}^{[r-1]}$,\,\,
					$\mathcal{Z}_i$,\,\,$\{\rho_j\}_{j=1}^K$,\,\,$\rho_{(max)}$,\,\,
					$t^{[th]}$,\par
					\hskip1.2em $\| {\bm x}_i^p\|_2$,\,\,
					$\breve{\mathcal{M}}_{sq}^{p[r-1]}$}
	\Statex{\textbf{Output:}\hskip.8em $\mathcal{Z}_i$,~~$\{ \rho_j \}_{j=1}^K$}

	\State{$\{ \|{\bm \mu}_j^p\|_2^2 \}_{j=1}^K \gets 0$}
		\Comment{Initialization}

	\LineComment{Exact partial similarity calculation}
	\For{$(s \leftarrow t_{(i,p)}) < t^{[th]}$ in all term IDs in $\hat{\bm x}_i$\hskip.3em}
	\label{csicp_g0:r1_start} 
		\For{$1\leq q\leq \mf_s$}\label{csicp_g0:innermost1}
			\State{$\rho_{c_{(s,q)}}\gets \rho_{c_{(s,q)}}+u_{s}\cdot v_{c_{(s,q)}}$}
			\label{csicp_g0:r1_end}
		\EndFor	
	\EndFor

	\LineComment{Calculate squared mean-L2-norm in object-subspace}
	\For{$(s \leftarrow t_{(i,p)}) \geq t^{[th]}$ in all term IDs in $\hat{\bm x}_i$\hskip.3em}
	\label{csicp_g0:sq_start}
		\For{$1\leq q\leq \mf_s$}\label{csicp_g0:innermost2}
			\State{\framebox[48mm][l]{\hskip.3em$\|{\bm \mu}_{c(s,q)}^p\|_2^2 \gets \|{\bm \mu}_{c(s,q)}^p\|_2^2 
					+v_{c(s,q)}^2$}}
			\label{csicp_g0:sq_end}
		\EndFor	
	\EndFor

	\LineComment{Gathering phase}
	\For{$1\leq j' \leq (nMv)$} \Comment{nMv: \#moving centroids}
	\label{csicp_g0:gather_start}
		\State{$j'$ is transformed to $j$.}
		\State{\hskip.3em\framebox[45mm][l]{\hskip.3em$\rho_{j}^{[ub]}\leftarrow \rho_{j} +\|{\bm x}_i^p\|_2\!\times\! 
			\sqrt{\|{\bm \mu}_j^p\|_2^2}$}}
			\label{csicp_g0:ub}
		\Comment{UB calculation}
		\If{$\rho_{j}^{[ub]} > \rho_{(max)}$\hskip.3em} \label{csicp_g0:ubp}
			\Comment{UBP filter}
			\State{$\mathcal{Z}_i \gets \mathcal{Z}_i \cup \{j\}$}
			\label{csicp_g0:gather_end}
		\EndIf
	\EndFor

  \end{algorithmic}
\end{algorithm}

Algorithm~\ref{algo:app_csicp} shows the pseudocode of the assignment step in CS-ICP.
The $\|{\bm x}_i^{p}\|_2$ and $\breve{\mathcal M}_{sq}^{p[r-1]}$ in the inputs respectively 
denote the $L_2$-norm of the $i$th object's {\em partial} feature vector 
and the {\em partial} squared mean-inverted index 
in the range of the term IDs ($s$) satisfying $s\!\geq\! t^{[th]}$.
At lines~\ref{icp_switch:gcs1_end} to \ref{icp_switch:gcs0_end},
the candidate centroids for the exact similarity calculations are collected by
the candidate-gathering functions $G_{(cs)0}$ and $G_{(cs)1}$.
The other parts have the same algorithm structure as those in Algorithm~\ref{algo:app_assign}
although each array in the partial mean-feature-inverted index for CS-ICP contains 
all the mean-feature values unlike ES-ICP whose array contains mean-feature values smaller than
$v^{[th]}$ in Region 3.

Algorithm~\ref{app:csicp_gather0} shows the pseudocode of 
the candidate-gathering function $G_{(cs)0}$ 
at line~\ref{icp_switch:gcs0_end} in Algorithm~\ref{algo:app_csicp}. 
$G_{(cs)0}$ is used for the object that does not satisfy the condition in Eq.~(\ref{eq:objstate}).
At lines~\ref{csicp_g0:r1_start} to \ref{csicp_g0:r1_end}, 
the exact partial similarities $\rho_{c_{(s,q)}}$ are calculated in the range of $t_{(i,p)}\!<\!t^{[th]}$.
At lines~\ref{csicp_g0:sq_start} to \ref{csicp_g0:sq_end}, 
the squared $L_2$-norm of the partial mean-feature vector $\|{\bm \mu}_{c_{(s,q)}}^p\|_2^2$
is calculated in the subspace of the $i$th object-feature vector,
where $v_{c_{(s,q)}}^2$ denotes the squared mean-feature value in $\breve{\mathcal M}_{sq}^{p[r-1]}$.
At lines~\ref{csicp_g0:gather_start} to \ref{csicp_g0:gather_end}, 
the gathering phase is performed and the centroid-ID set ${\mathcal Z}_i$ 
for exact similarity calculations is returned.
At line~\ref{csicp_g0:ub}, the upper bound on the similarity to the $j$th centroid 
$\rho_j^{[ub]}$ is calculated by the sum of the foregoing exact partial similarity $\rho_j$ and 
the product of the pre-calculated $L_2$-norm of the $i$th object's  partial feature vector
$\|{\bm x}_i^p\|_2$ and 
$\sqrt{\|{\bm \mu}_{c_{(s,q)}}^p\|_2^2}$.
In this product calculation, the square-root operation is performed, 
which requires high computational cost.

The function $G_{(cs)1}$ at line~\ref{icp_switch:gcs1_end} in Algorithm~\ref{algo:app_csicp} 
is similar to $G_{(cs)0}$.
The difference is in the endpoint of the loops 
at lines \ref{csicp_g0:innermost1} and \ref{csicp_g0:innermost2}. 
$G_{(cs)0}$ uses $(mf)_s$ while 
$G_{(cs)1}$ does $(mfM)_s$ in the mean-inverted index consisting of the moving centroids.

\subsection{Performance Comparison}
This section shows the actual performance of 
ES-ICP, ICP, CS-ICP, and TA-ICP when they were applied to 
the 8.2M-sized PubMed data set with $K\!=\!80\,000$ 
in Tables~\ref{table:app_comp} and ~\ref{table:app_perf_comp}  and 
the 1M-sized NYT data set with $K\!=\!10\,000$ 
in Tables~\ref{table:app_comp_nyt} and ~\ref{table:app_perf_comp_nyt}.

From the both results in the two distinct data sets and settings,
in addition to the facts described in Section~\ref{ssec:perfcomp},
we note the following two facts.
(1) CS-ICP and TA-ICP required more elapsed time than the baseline ICP
although they operated using the smaller or around equal number of 
(completed) instructions or multiplications.
This is because they caused the larger numbers of branch mispredictions (BMs) and 
last-level-cache misses (LLCMs).
In particular, TA-ICP did much more branch mispredictions than the others 
while their numbers of branch instructions were not much different.
This led to its worse performance.
(2) The elapsed-time differences of the four algorithms came from 
those in the assignment step which we focused on.
The four algorithms worked in not much different average update time 
although they constructed the different data structures.
In particular, ES-ICP and CS-ICP took less update time than the baseline ICP
because they processed fewer centroids owing to their high pruning rates.

In the NYT results in Table~\ref{table:app_comp_nyt},
we note that surprisingly, ES-ICP took less elapsed time in the assignment step than 
that in the update step.
For any further acceleration in such data sets and settings, 
we will need to change our strategy 
of the acceleration in the assignment step to that in both the steps including the update.

These, in addition to the results in Section~\ref{ssec:perfcomp}, show 
that only reducing the numbers of multiplications or instructions does not always 
lead to reducing elapsed time required by an algorithm. 
Suppressing the performance-degradation factors in Section~\ref{sec:arch} 
realizes the acceleration.
In other words,
by carefully designing an algorithm so that it suppresses the numbers of instructions, 
last-level cache misses (LLCMs), and branch mispredictions (BMs), 
we can obtain an efficient algorithm that operates in the {\em architecture-friendly manner}.

We also note that 
reducing the maximum memory size required by ES-ICP, 
which is described as the remaining task in Section~\ref{ssec:perfcomp},
is not a serious problem with respect to the actual memory size and the LLCMs 
in the current data sets and settings from Tables~\ref{table:app_comp} and \ref{table:app_comp_nyt}.
The memory sizes used for the partial mean-inverted indexes $\breve{\mathcal M}^{p[r]}$ 
in the 8.2M-sized PubMed and 1M-sized NYT data sets in the foregoing settings were 
around 8.3 GB and 3.2 GB, respectively.
Even for the PubMed data set, total amount size was 16.72 GB.
A current standard computer system for scientific and technical computing equips a memory system 
whose size is much larger than the foregoing amount.
Regarding LLCMs,
the number of accesses to $\breve{\mathcal M}^{p[r]}$ were small, i.e., 
$\breve{\mathcal M}^{p[r]}$ was rarely loaded to the caches
because ES-ICP's filters reduced unpruned centroids that were targets for exact similarity calculations 
using $\breve{\mathcal M}^{p[r]}$.

\section{Main-Filter Comparison}\label{app:compubp}
There may be a doubt of the combination of the main UBP and auxiliary ICP filters 
in the compared algorithms weakens 
the main filter's positive effect on the performance, 
To clear up the doubt, 
this section compares only the main filters of ES-ICP, TA-ICP, and CS-ICP,
where algorithms with the filters are respectively called ES-MIVI, TA-MIVI, and CS-MIVI.
The algorithms were applied to 
the 8.2M-sized PubMed data set with $K\!=\!80\,000$ 
in Tables~\ref{table:app_compubp} and ~\ref{table:app_perf_compubp}  and 
the 1M-sized NYT data set with $K\!=\!10\,000$ 
in Tables~\ref{table:app_compubp_nyt} and ~\ref{table:app_perf_compubp_nyt}.

The most important fact is that no algorithm without the ICP filter improved its elapsed time
in the data sets and settings, compared with the corresponding algorithm with the ICP filter.
This is because the algorithm with the ICP filter suppressed the performance-degradation factors 
of the numbers of instructions, branch mispredictions (BMs), and last-level-cache misses (LLCMs).

ES-MIVI showed the best performance regardless of the data sets and settings.
We note that the ES filter was effective by itself and 
from the ES-ICP's results in Tables~\ref{table:app_comp} to \ref{table:app_perf_comp_nyt},
both the filters of ES and ICP worked without losing each other's advantages.
When combining plural filters like the UBP and ICP filters,
we should carefully design the combined algorithm that operates in the architecture-friendly manner.

\section{Initial-State Independence}\label{app:init}
We deal with a large-scale and high-dimensional sparse document data set
and partition it into numerous classes, i.e., use a huge $K$ value (Section~\ref{sec:intro}).
Compared with widely-used Lloyd-type $K$-means settings in the metric space, 
this setting differs in data size $N$, dimensionality $D$, and number of clusters $K$. 
It is often said that $K$-means clustering performance strongly depends on an initial state, 
i.e., seeding \citeappx{celebi}.
In our setting, however, algorithm performance is independent of initial states.

In a high-dimensional metric space,
object vectors are located far away from each other and 
an average and a variance of their distances are very large and small, respectively
\citeappx{chavez,beyer}.
Such a phenomenon has been called the curse of dimensionality \citeappx{bellman}.
Similariy, the phenomenon occurs in a high-dimensional hypersphere 
in the spherical $K$-means setting.
In this setting, we consider that 
a seeding method randomly choosing an initial state yields a similar result to 
a commonly used one in $K$-means++ \citeappx{arthur_supp}.

\begin{figure}[t]
\begin{center}
	\psfrag{X}[c][c][0.9]{
		\begin{picture}(0,0)
			\put(0,0){\makebox(0,-10)[c]{Number of clusters: $K$ (log scale)}}
		\end{picture}
	}
	\psfrag{Y}[c][c][0.9]{
		\begin{picture}(0,0)
			\put(0,0){\makebox(0,20)[c]{$\mbox{\em NMI}(\mathcal{C}_a,\mathcal{C}_b)$}}
		\end{picture}
	}
	\psfrag{Z}[c][c][0.85]{
		\begin{picture}(0,0)
			\put(0,0){\makebox(0,-16)[c]{Standard deviation ($\times 10^{-2}$)}}
		\end{picture}
	}
	\psfrag{A}[c][c][0.85]{$10$} \psfrag{B}[c][c][0.85]{$10^2$}
	\psfrag{C}[c][c][0.85]{$10^3$} \psfrag{D}[c][c][0.85]{$10^4$}
	\psfrag{E}[c][c][0.85]{$10^5$}
	\psfrag{F}[r][r][0.87]{$0.6$} \psfrag{G}[r][r][0.87]{$0.7$}
	\psfrag{H}[r][r][0.87]{$0.8$} \psfrag{I}[r][r][0.87]{$0.9$}
	\psfrag{J}[l][c][0.85]{$0$} \psfrag{K}[l][c][0.85]{$2$}
	\psfrag{L}[l][c][0.85]{$4$} \psfrag{M}[l][c][0.85]{$6$}
	\psfrag{P}[r][r][0.85]{{\em NMI}}
	\psfrag{Q}[r][r][0.85]{STD}
	\includegraphics[width=56mm]{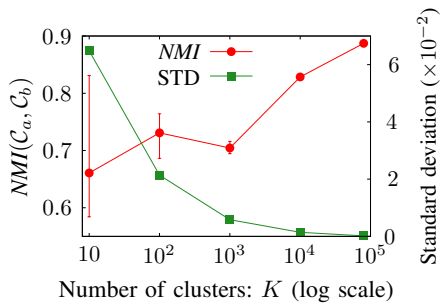}
\end{center}
\caption{\small
{\em NMI} of clustering results for 8.2M-sized PubMed.
}
\label{fig:app_nmi_pubmed}
\vspace*{-2mm}
\end{figure}

\begin{figure}[t]
\begin{center}
	\psfrag{X}[c][c][0.9]{
		\begin{picture}(0,0)
			\put(0,0){\makebox(0,-10)[c]{Number of clusters: $K$ (log scale)}}
		\end{picture}
	}
	\psfrag{Y}[c][c][0.9]{
		\begin{picture}(0,0)
			\put(0,0){\makebox(0,20)[c]{$\mbox{\em NMI}(\mathcal{C}_a,\mathcal{C}_b)$}}
		\end{picture}
	}
	\psfrag{Z}[c][c][0.85]{
		\begin{picture}(0,0)
			\put(0,0){\makebox(0,-16)[c]{Standard deviation ($\times 10^{-2}$)}}
		\end{picture}
	}
	\psfrag{A}[c][c][0.85]{$10$} \psfrag{B}[c][c][0.85]{$10^2$}
	\psfrag{C}[c][c][0.85]{$10^3$} \psfrag{D}[c][c][0.85]{$10^4$}
	\psfrag{F}[r][r][0.87]{$0.6$} \psfrag{G}[r][r][0.87]{$0.7$}
	\psfrag{H}[r][r][0.87]{$0.8$} \psfrag{I}[r][r][0.87]{$0.9$}
	\psfrag{J}[l][c][0.85]{$0$} \psfrag{K}[l][c][0.85]{$1$}
	\psfrag{L}[l][c][0.85]{$2$} \psfrag{M}[l][c][0.85]{$3$}
	\psfrag{N}[l][c][0.85]{$4$}
	\psfrag{P}[r][r][0.85]{{\em NMI}}
	\psfrag{Q}[r][r][0.85]{STD}
	\includegraphics[width=56mm]{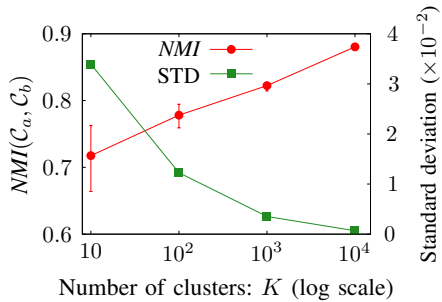}
\end{center}
\caption{\small
{\em NMI} of clustering results for 1M-sized NYT.
}
\label{fig:app_nmi_nyt}
\vspace*{-2mm}
\end{figure}

We observed in our preliminary experiments that 
when an algorithm started at different initial states, 
the resultant values of two distinct evaluation measures \citeappx{wagner} did not change so much.
One of the two measures is an objective function value at convergence, 
which is repersented by
\begin{eqnarray}
J(\mathcal{C}_h) &=& \sum_{C_j\in\mathcal{C}_h}\sum_{\bm{x}_i\in C_j}\bm{x}_i\cdot \bm{\mu}_j
\label{eq:app_objfunct}\\
J &=& \left( \frac{1}{L}\right) \sum_{1\leq h\leq L} J(\mathcal{C}_h)\, ,
\label{eq:app_avg_objfunct}
\end{eqnarray}
where $\mathcal{C}_h$ denotes the clustering result obtained by the $h$th initial states 
($h\!=\!1,2,\cdots,L$) and $L$ is the number of the prepaired initial states.
The other is normalized mutual information {\em NMI}.
Given two clustering results $\mathcal{C}_a$ and $\mathcal{C}_b$,
$\mbox{\em NMI}(\mathcal{C}_a,\mathcal{C}_b)$ is defined by
\begin{equation}
\mbox{\em NMI}(\mathcal{C}_a,\mathcal{C}_b) = \frac{\mathcal{I}(\mathcal{C}_a,\mathcal{C}_b)}
{\sqrt{\mathcal{H}(\mathcal{C}_a)\mathcal{H}(\mathcal{C}_b)}}\,,
\label{eq:app_nmi_pair}
\end{equation}
where $\mathcal{I}(\mathcal{C}_a,\mathcal{C}_b)$ and $\mathcal{H}(\mathcal{C}_a)$ denote 
the mutual infomation between $\mathcal{C}_a$ and $\mathcal{C}_b$ 
and the entropy of $\mathcal{C}_a$, respectively \citeappx{strehl}.
When the $L$ initial states are given to an algorithm,
{\em NMI} is expressed by
\begin{equation}
\mbox{\em NMI} = \left( \fracslant{1}{\dbinom{L}{2}}\right) \sum_{\substack{1\leq a,b\leq L\\ a\neq b}}
\mbox{\em NMI}(\mathcal{C}_a,\mathcal{C}_b)\, .
\label{eq:app_nmi}
\end{equation}
A larger value of $J(\mathcal{C}_h)$ means that $\mathcal{C}_h$ is a better clustering result
and if $J(\mathcal{C}_a)\!\sim\!J(\mathcal{C}_b)$, $\mathcal{C}_a$ is almost equivalent to 
$\mathcal{C}_b$ in terms of optimization.
$\mbox{\em NMI}(\mathcal{C}_a,\mathcal{C}_b)$ measures the similarity 
between the two clustering results of $\mathcal{C}_a$ and $\mathcal{C}_b$. 
A larger value of $\mbox{\em NMI}(\mathcal{C}_a,\mathcal{C}_b)$ means 
that the two clustering results are more similar.
If $\mbox{\em NMI}(\mathcal{C}_a,\mathcal{C}_b)\!=\!1$, 
$\mathcal{C}_a$ and $\mathcal{C}_b$ are completely identical.

\begin{figure}[t]
\begin{center}
	\psfrag{X}[c][c][0.9]{
		\begin{picture}(0,0)
			\put(0,0){\makebox(0,-10)[c]{Number of clusters: $K$ (log scale)}}
		\end{picture}
	}
	\psfrag{Y}[c][c][0.9]{
		\begin{picture}(0,0)
			\put(0,0){\makebox(0,16)[c]{CV of {\em NMI} ($\times 10^{-2}$)}}
		\end{picture}
	}
	\psfrag{Z}[c][c][0.9]{
		\begin{picture}(0,0)
			\put(0,0){\makebox(0,-16)[c]{CV of $J$ ($\times 10^{-4}$)}}
		\end{picture}
	}
	\psfrag{A}[c][c][0.85]{$10$} \psfrag{B}[c][c][0.85]{$10^2$}
	\psfrag{C}[c][c][0.85]{$10^3$} \psfrag{D}[c][c][0.85]{$10^4$}
	\psfrag{E}[c][c][0.85]{$10^5$}
	\psfrag{F}[r][r][0.85]{$0$} \psfrag{G}[r][r][0.85]{$2$}
	\psfrag{H}[r][r][0.85]{$4$} \psfrag{I}[r][r][0.85]{$6$}
	\psfrag{K}[r][r][0.85]{$8$} \psfrag{L}[r][r][0.85]{$10$}
	\psfrag{M}[l][c][0.85]{$0$} \psfrag{N}[l][c][0.85]{$4$}
	\psfrag{O}[l][c][0.85]{$8$} \psfrag{P}[l][c][0.85]{$12$}
	\psfrag{Q}[l][c][0.85]{$16$}
	\psfrag{J}[r][r][0.78]{$J$}
	\psfrag{U}[r][r][0.74]{{\em NMI}}
	\includegraphics[width=56mm]{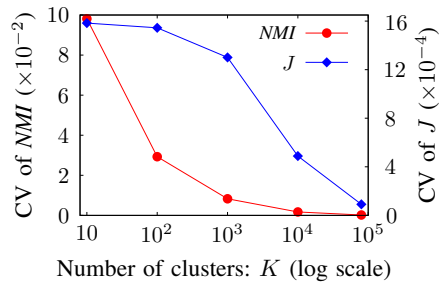}
\end{center}
\caption{\small
Coefficient of variations (CV) of objective function value $J$ and 
{\em NMI} of clustering results for 8.2M-sized PubMed.
}
\label{fig:app_cv_pubmed}
\vspace*{-2mm}
\end{figure}

\begin{figure}[t]
\begin{center}
	\psfrag{X}[c][c][0.9]{
		\begin{picture}(0,0)
			\put(0,0){\makebox(0,-10)[c]{Number of clusters: $K$ (log scale)}}
		\end{picture}
	}
	\psfrag{Y}[c][c][0.9]{
		\begin{picture}(0,0)
			\put(0,0){\makebox(0,16)[c]{CV of {\em NMI} ($\times 10^{-2}$)}}
		\end{picture}
	}
	\psfrag{Z}[c][c][0.9]{
		\begin{picture}(0,0)
			\put(0,0){\makebox(0,-16)[c]{CV of $J$ ($\times 10^{-3}$)}}
		\end{picture}
	}
	\psfrag{A}[c][c][0.85]{$10$} \psfrag{B}[c][c][0.85]{$10^2$}
	\psfrag{C}[c][c][0.85]{$10^3$} \psfrag{D}[c][c][0.85]{$10^4$}
	\psfrag{E}[c][c][0.85]{$0$} \psfrag{F}[r][r][0.85]{$1$}
	\psfrag{G}[r][r][0.85]{$2$} \psfrag{H}[r][r][0.85]{$3$}
	\psfrag{I}[r][r][0.85]{$4$} \psfrag{K}[r][r][0.85]{$5$}
	\psfrag{L}[l][c][0.85]{$0$} \psfrag{M}[l][c][0.85]{$2$}
	\psfrag{N}[l][c][0.85]{$4$} \psfrag{O}[l][c][0.85]{$6$}
	\psfrag{J}[r][r][0.78]{$J$}
	\psfrag{U}[r][r][0.74]{{\em NMI}}
	\includegraphics[width=56mm]{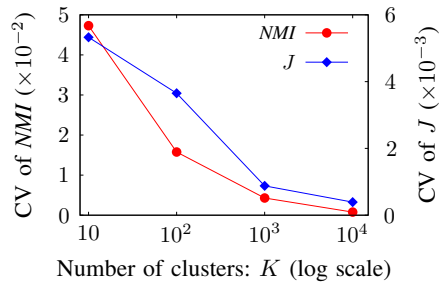}
\end{center}
\caption{\small
Coefficient of variations (CV) of objective function value $J$ and 
{\em NMI} of clustering results for 1M-sized NYT.
}
\label{fig:app_cv_nyt}
\vspace*{-2mm}
\end{figure}

To confirm experimentally that the sensitivity of an initial state is very low in our setting,
in particular, at large $K$ values, 
we evaluated the two measures, varying the $K$ values of our proposed algorithm
in both the 8.2M-sized PubMed and the 1M-sized NYT data set.
In the 8.2M-sized PubMed data set,
the five $K$ values of $10$, $100$, $1\,000$, $10\,000$, and $80\,000$ were used
while the four values except $80\,000$ were done in the 1M-sized NYT.
The proposed algorithm with each of the foregoing $K$ values started at 
the different 10 initial states chosen randomly.

Figures~\ref{fig:app_nmi_pubmed} and \ref{fig:app_nmi_nyt} show 
the values of $\mbox{\em NMI}(\mathcal{C}_a,\mathcal{C}_b)$ and 
their standard deviations in the 8.2M-sized PubMed and the 1M-sized NYT data set, respectively.
Both the results show the similar tendency that 
the {\em NMI} values increased and the standard deviations decreased with $K$.
For the huge $K$ values, the {\em NMI} values approached 0.9.
Even when $K\!=\!10$, the {\em NMI} values were around 0.7.
This shows that the clustering results obtained by randomly chosen different initial states 
were very similar in our setting.

Furthermore, we utilized as statistics of the two measures a coefficient of variation ({\em CV}) 
that is expressed as 
\begin{equation}
CV = \frac{\sigma_z}{\bar{z}}\, ,
\end{equation}
where $\sigma_z$ and $\bar{z}$ denote the standard deviation and the average of 
variable $z$, respectively, and $z$ means $J$ or {\em NMI} in our case.
A smaller {\em CV} of $z$ represents that the $z$'s variation is smaller.
Figures~\ref{fig:app_cv_pubmed} and \ref{fig:app_cv_nyt} show 
the CV's of the objective function values $J$ and the normalized mutual information
{\em NMI} in the 8.2M-sized PubMed and the 1M-sized NYT data set, respectively.
We know that the {\em CV}'s of $J$ and {\em NMI} decreased with $K$.
Thus, the clustering results are independent of the initial states
in our setting of using the large values of $N$, $D$, and $K$.

\section{Pareto-Principle-Like Phenomenon}\label{app:pareto}
We observed in our preliminary experiments 
that a {\em cumulative partial similarity} of an object to a centroid and 
a {\em normalized rank} of each partial similarity have a relationship 
like the Pareto principle \citeappx{newman_supp}.
We first define both the cumulative partial similarity and the normalized rank.
Consider that we calculate a similarity $\rho_{a(i)}$ between the $i$th object ${\bm x}_i$
and centorid ${\bm \mu}_{a(i)}$ of the cluster which ${\bm x}_i$ belongs to.
The similarity $\rho_{a(i)}$ is expressed by
\begin{equation}
\rho_{a(i)} = \sum_{p=1}^{(nt)_i} u_{t_{(i,p)}}\cdot \mu_{(a(i),t_{(i,p)})}\, ,
\label{eq:app_sim}
\end{equation}
where $\mu_{(a(i),t_{(i,p)})}$ denotes the element of ${\bm \mu_{a(i)}}$ whose term ID
is $t_{(i,p)}$, $u_{t_{(i,p)}}$ the object-feature value whose term ID is $t_{(i,p)}$,
and $(nt)_i$ the number of distinct terms which the object ${\bm x}_i$ uses 
(Table~\ref{table:nota}). 
We sort the partial similarities $[u_{t_{(i,p)}}\cdot \mu_{(a(i),t_{(i,p)})}]_{p=1}^{(nt)_i}$ 
in descending order 
and express the $h$th rank's partial similarity as $\delta\rho_{a(i)}(h)$, $h=1,2,\cdots,(nt)_i$.
Note that $h$ corresponds to the number of multiplications for the partial similarity calculation.
Then, the normalized rank $\mbox{\em NR}(i,h)$ and the cumulative partial similarity 
$\mbox{\em CPS}(i,h)$ are respectively expressed by 
\begin{eqnarray}
\mbox{\em NR}(i,h) &=& \frac{h}{(nt)_i} \label{eq:app_normrank}\\
\mbox{\em CPS}(i,h) &=& \frac{1}{\rho_{a(i)}}\sum_{h'=1}^{h} \delta\rho_{a(i)}(h')\,,
\label{eq:app_cps}
\end{eqnarray}
where $\mbox{\em CPS}(i,(nt)_i)\!=\!1$.
For statistical processing with reagrd to all the object,
we introduce ordered bin instead of $\mbox{\em NR}(i,h)$.
We rename the foregoing $\mbox{\em NR}(i,h)$ an individual normalized rank 
and the ordered bin a normalized rank again.
The normalized ranks (ordered bins) {\em NR}$(\hat{h})$ are discrete values 
aligned at regular intervals expressed as $\delta b$, 
where $\hat{h}$ denotes the bin ID of an integer from 0 to $1/\delta b$.
For instance, if $\delta b\!=\!0.01$, then $0\!\leq (\hat{h}\!\in\!Z) \leq\! 100$.
An average {\em CPS} with regard to all the objects, $\overline{\mbox{\em CPS}}(\hat{h})$, 
is defined by
\begin{eqnarray}
\mbox{\em NR}(\hat{h}) &=& \hat{h}\cdot \delta b \label{eq:app_bin} \\
\overline{\mbox{\em CPS}}(\hat{h}) &=& 
\frac{1}{N}\sum_{i=1}^N \mbox{\em CPS}(i,((nt)_i\cdot \mbox{\em NR}(\hat{h})) )\, ,
\label{eq:app_avg_cps}
\end{eqnarray}
where if $(nt)_i\!\cdot\!\mbox{\em NR}(\hat{h})\!\notin\!\{\mbox{\em NR}(i,h)\}$, 
$\mbox{\em CPS}(i,((nt)_i\!\cdot\!\mbox{\em NR}(\hat{h})) )$
is calculated with linear interporation.

\begin{figure}[t]
\begin{center}
	\psfrag{X}[c][c][0.95]{
		\begin{picture}(0,0)
			\put(0,0){\makebox(0,-10)[c]{Normalized rank}}
		\end{picture}
	}
	\psfrag{Y}[c][c][0.9]{
		\begin{picture}(0,0)
			\put(0,0){\makebox(0,48)[c]{Avg. cumulative }}
			\put(0,0){\makebox(0,22)[c]{partial similarity ({\em CPS})}}
		\end{picture}
	}
	\psfrag{Z}[c][c][0.88]{
		\begin{picture}(0,0)
			\put(0,0){\makebox(0,-20)[c]{Standard deviation ($\times 10^{-2}$)}}
		\end{picture}
	}
	\psfrag{A}[c][c][0.85]{$0$} \psfrag{B}[c][c][0.85]{$0.2$}
	\psfrag{C}[c][c][0.85]{$0.4$} \psfrag{D}[c][c][0.85]{$0.6$}
	\psfrag{E}[c][c][0.85]{$0.8$} \psfrag{F}[c][c][0.85]{$1$}
	\psfrag{G}[r][r][0.85]{$0$} \psfrag{H}[r][r][0.85]{$0.2$}
	\psfrag{I}[r][r][0.85]{$0.4$} \psfrag{J}[r][r][0.85]{$0.6$}
	\psfrag{K}[r][r][0.85]{$0.8$} \psfrag{L}[r][r][0.85]{$1$}
	\psfrag{M}[l][c][0.85]{$0$} \psfrag{N}[l][c][0.85]{$5$}
	\psfrag{O}[l][c][0.85]{$10$} \psfrag{P}[l][c][0.85]{$15$}
	\psfrag{Q}[l][c][0.85]{$20$} \psfrag{R}[l][c][0.85]{$25$}
	\psfrag{S}[r][r][0.8]{$\overline{\mbox{\em CPS}}$ at 2nd iteration}
	\psfrag{T}[r][r][0.82]{~~ at convergence}
	\psfrag{V}[r][r][0.8]{{\em STD} at 2nd iteration}
	\psfrag{W}[r][r][0.82]{~~ at convergence}
	\psfrag{U}[l][l][0.82]{(0.10,\,0.92)}
	\includegraphics[width=60mm]{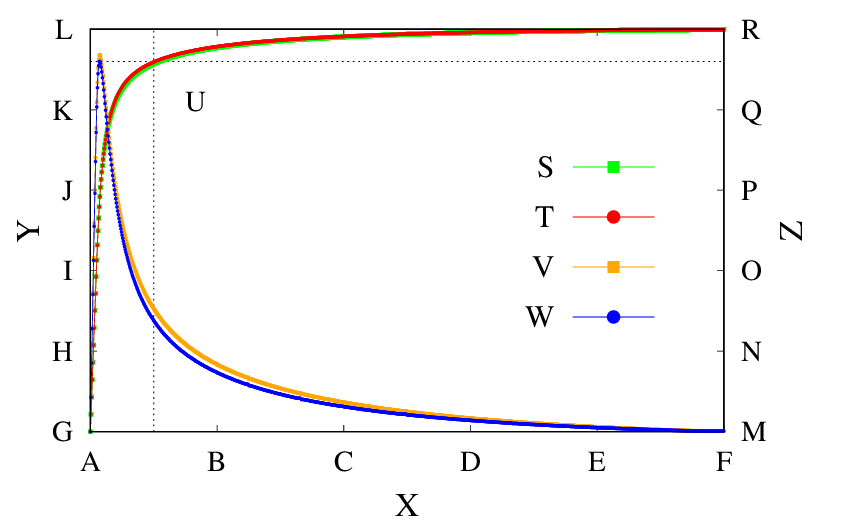}
\end{center}
\caption{\small
Average cumulative partial similarity ({\em CPS}) and its standard deviation 
({\em STD}) against normalized rank
when a spherical K-means algorithm with K=80\,000 was applied to 
the 8.2M-sized PubMed.
}
\label{fig:app_pareto_pubmed}
\vspace*{3mm}
\end{figure}

\begin{figure}[t]
\begin{center}
	\psfrag{X}[c][c][0.95]{
		\begin{picture}(0,0)
			\put(0,0){\makebox(0,-10)[c]{Normalized rank}}
		\end{picture}
	}
	\psfrag{Y}[c][c][0.9]{
		\begin{picture}(0,0)
			\put(0,0){\makebox(0,48)[c]{Avg. cumulative }}
			\put(0,0){\makebox(0,22)[c]{partial similarity ({\em CPS})}}
		\end{picture}
	}
	\psfrag{Z}[c][c][0.88]{
		\begin{picture}(0,0)
			\put(0,0){\makebox(0,-20)[c]{Standard deviation ($\times 10^{-2}$)}}
		\end{picture}
	}
	\psfrag{A}[c][c][0.85]{$0$} \psfrag{B}[c][c][0.85]{$0.2$}
	\psfrag{C}[c][c][0.85]{$0.4$} \psfrag{D}[c][c][0.85]{$0.6$}
	\psfrag{E}[c][c][0.85]{$0.8$} \psfrag{F}[c][c][0.85]{$1$}
	\psfrag{G}[r][r][0.85]{$0$} \psfrag{H}[r][r][0.85]{$0.2$}
	\psfrag{I}[r][r][0.85]{$0.4$} \psfrag{J}[r][r][0.85]{$0.6$}
	\psfrag{K}[r][r][0.85]{$0.8$} \psfrag{L}[r][r][0.85]{$1$}
	\psfrag{M}[l][c][0.85]{$0$} \psfrag{N}[l][c][0.85]{$5$}
	\psfrag{O}[l][c][0.85]{$10$} \psfrag{P}[l][c][0.85]{$15$}
	\psfrag{Q}[l][c][0.85]{$20$} \psfrag{R}[l][c][0.85]{$25$}
	\psfrag{S}[r][r][0.8]{$\overline{\mbox{\em CPS}}$ at 2nd iteration}
	\psfrag{T}[r][r][0.82]{~~ at convergence}
	\psfrag{V}[r][r][0.8]{{\em STD} at 2nd iteration}
	\psfrag{W}[r][r][0.82]{~~ at convergence}
	\psfrag{U}[l][l][0.82]{(0.10,\,0.90)}
	\includegraphics[width=60mm]{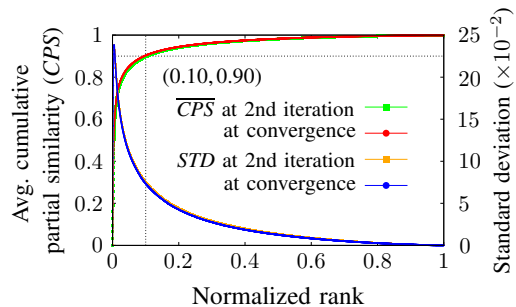}
\end{center}
\caption{\small
Average cumulative partial similarity ({\em CPS}) and its standard deviation 
({\em STD}) against normalized rank
when a spherical K-means algorithm with K=10\,000 was applied to 
the 1M-sized NYT.
}
\label{fig:app_pareto_nyt}
\vspace*{-5mm}
\end{figure}

Figures~\ref{fig:app_pareto_pubmed} and \ref{fig:app_pareto_nyt} show 
$\overline{\mbox{\em CPS}}(\hat{h})$ and its standard deviation {\em STD}$(\hat{h})$ 
against $\mbox{\em NR}(\hat{h})$ 
when a spherical $K$-means algorithm was applied to the 8.2M-sized PubMed data set 
with $K\!=\!80\,000$ and the 1M-sized NYT with $K\!=\!10\,000$, respectively.
In each figure, the two curves for each of $\overline{\mbox{\em CPS}}$ and 
{\em STD} are depicted, which correspond to those at the second iteration and 
at the convergence.
The two curves of $\overline{\mbox{\em CPS}}$ in both the figures 
showed the almost similar characteristics tha increased rapidly and reached 1.0,
despite the number of iterations (at the second or convergence).
In particular, $\overline{\mbox{\em CPS}}(0.1)\!=\!0.92,\,0.90$ in the 8.2M-sized PubMed
and the 1M-sized NYT data set, respectively.
Besides, the standard deviations at each normalized rank were small.
These characteristics seem like Pareto principle, i.e.,
a large fraction of a similarity is calculated at a very low cost.

\section{Pruning Based on Triangle Inequality}\label{app:triangle}
\begin{figure}[t]
\begin{center}
	\psfrag{X}[c][c][0.88]{${\bm x}_i$}
	\psfrag{Y}[r][r][0.88]{${\bm \mu}_{a(i)}^{[r-1]}$}
	\psfrag{A}[l][l][0.88]{${\bm \mu}_j^{[r-2]}$}
	\psfrag{B}[c][r][0.88]{${\bm \mu}_j^{[r-1]}$}
	\psfrag{P}[r][r][0.88]{${\bm \mu}_{a(i)}^{[r-2]}$}
	\psfrag{Q}[r][r][0.88]{${\bm \mu}_{a(i)}^{[r-1]}$}
	\psfrag{D}[l][l][0.88]{$\delta_j^{[r-1]}$}
	\psfrag{U}[l][l][0.85]{$d_{UB}$}
	\psfrag{L}[l][l][0.85]{$d_{LB}$}
	\includegraphics[width=53mm]{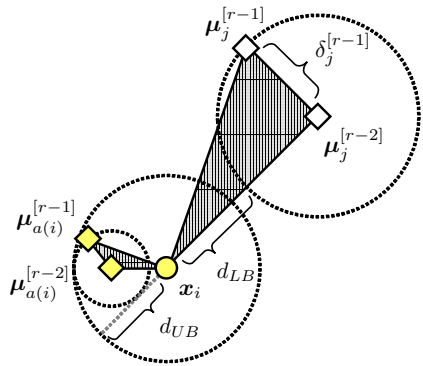}
\end{center}
\caption{Typical usage of the triangle inequality for omitting the exact distance calculation
of object ${\bm x}_i$ to centroid ${\bm \mu}_j$ in a metric space.
The distance calculation is omitted if $d_{UB}\leq d_{LB}$.}
\label{fig:app_triangleIneq}
\end{figure}

We describe a typical triangle-inequality-based pruning method 
while referring to Fig.~\ref{fig:app_triangleIneq}.
Let $d({\bm x}_i,{\bm \mu}_j^{[r-1]})$ denote the distance between 
${\bm x}_i$ and ${\bm \mu}_j^{[r-1]}$
and $\delta_j^{[r-1]}$ denote the moving distance between 
${\bm \mu}_j^{[r-1]}$ and ${\bm \mu}_j^{[r-2]}$,
where the superscript $[r\!-\! 1]$ denotes the $(r\!-\! 1)$th iteration.
$d_{LB}$ is the abbreviation of
$d_{LB}({\bm x}_i,{\bm \mu}_j^{[r-1]})$ that is the lower bound on 
$d({\bm x}_i,{\bm \mu}_j^{[r-1]})$.
Similarly, $d_{UB}$ is that of $d_{UB}({\bm x}_i,{\bm \mu}_{a(i)}^{[r-1]})$.
Note that when the distance is used as a measure like in this case, 
we focus on the lower bound on the distance between an object and a mean (centroid)
unlike in the case of a similarity measure. 

Consider which centroid of $\bm{\mu}_{a(i)}^{[r-1]}$ or $\bm{\mu}_j^{[r-1]}$ 
is closer to $\bm{x}_i$, given
the two centroids' distances to $\bm{x}_i$ at the $(r\!-\!2)$th iteration 
and the moving distances of the centroids between 
the $(r\!-\!2)$th and the $(r\!-\!1)$th iteration, i.e., 
$\delta_j^{[r-1]}$ and $\delta_{a(i)}^{[r-1]}$.
We apply the triangle inequality to the three points of 
${\bm x}_i$, ${\bm \mu}_{(*)}^{[r-2]}$, and ${\bm \mu}_{(*)}^{[r-1]}$ 
under the condition of $\delta_{(*)}^{[r-1]}$ and 
$d_{(*)}({\bm x}_i,{\bm \mu}_{(*)}^{[r-2]})$ are known,
where the subscript $(*)$ denotes either $j$ or $a(i)$.
Then $d_{LB}$ and $d_{UB}$ are expressed by 
\begin{eqnarray}
d_{LB}({\bm x}_i,{\bm \mu}_j^{[r-1]}) &=& 
|d({\bm x}_i,{\bm \mu}_j^{[r-2]})\!-\!\delta_j^{[r-1]}| \label{eq:app_lb} \\
d_{UB}({\bm x}_i,{\bm \mu}_{a(i)}^{[r-1]}) &=& 
d({\bm x}_i,{\bm \mu}_{a(i)}^{[r-2]})\!+\!\delta_{a(i)}^{[r-1]}\, .
\label{eq:app_ub}
\end{eqnarray}
If $d_{UB} < d_{LB}$, then 
we know that $\bm{\mu}_{a(i)}^{[r-1]}$ is closer to ${\bm x}_i$. 
As a result, we can omit calculations of the exact distance of ${\bm x}_i$ to ${\bm \mu}_{(*)}^{[r-1]}$.
In the pruning method, 
a key distance is the moving distance such as $\delta_{(*)}^{[r-1]}$.
Note that we can use the exact distance of 
$d({\bm x}_i,{\bm \mu}_{a(i)}^{[r-1]})$ instead of $d_{UB}$.
The lower bound tightens as the moving distance becomes smaller.
Then more centroids are pruned, causing acceleration.
This acceleration becomes effective only around the last stage  
before the convergence where most of the centroids are invariant or slightly move.
In our setting, 
it is not efficient 
as stated in Section~\ref{sec:arch}.
It is desired that 
the acceleration goes through all the iterations, 
particularly from the early to the middle stage.

\bibliographystyleappx{IEEEtran}

\newpage
\captionsetup{skip=3pt}
\begin{table*}[thb]
\centering
\small
\begin{threeparttable}
\caption{
Ablation study 
in terms of average number of multiplications and average elapsed time 
until convergence in 8.2M-sized PubMed with K=80\,000.\hspace*{7mm}\newline 
Number of iterations until convergence is 64.
}
\label{table:app_abl}
\centering
\begin{tabular}{|c|c|c||c|}\hline
\multirow{3}{*}{Algorithm} & Avg. \# multiplications & Avg. elapsed time & Maximum\\
 &  per iteration & per iteration (sec): & memory size\\
 &   & [assignment, update]\tnote{$\dagger$} & (GB) \\\hline
ES-ICP & 9.391$\times 10^{10}$ & 213.8 [185.5, 28.79]& 16.72\\\hline
ES & 3.562$\times 10^{11}$ & 330.6 [304.0, 27.00] & 16.69\\\hline
ThV & 4.538$\times 10^{11}$ & 358.4 [329.37, 29.50] & 96.59\\\hline
ThT & 1.122$\times 10^{13}$ &2172 [2149, 23.81]  &  8.52\\\hline
\end{tabular}
\begin{tablenotes}
\item[$\dagger$]{\small The average elapsed time does not exactly match the sum of 
the assignment and the update time because the algorithm terminated at the end of 
the assignment step of the last iteration.}
\end{tablenotes}
\end{threeparttable}
\end{table*}

\begin{table*}[thb]
\centering
\small
\begin{threeparttable}
\caption{
Ablation study in terms of perf results until convergence 
in 8.2M-sized PubMed with K=80\,000.
Number of iterations until convergence is 64.
}
\label{table:app_perf_abl}
\centering
\begin{tabular}{|c|c|c|c|c|c|}\hline
\multirow{2}{*}{Algorithm} & \multirow{2}{*}{\# instructions} 
& \multirow{2}{*}{\# branches} & \# branch misses 
& \multirow{2}{*}{\# LLC-loads} & \# LLC-loads\\
& & & (\%) & & misses (\%)\\
\hline
ES-ICP & 6.157$\times 10^{14}$ & 8.417$\times 10^{13}$ & 9.569$\times 10^{10}$ $(0.11)$
& 1.043$\times 10^{13}$ & 1.738$\times 10^{12}$ $(16.7)$ \\\hline
ES & 1.097$\times 10^{15}$ & 1.659$\times 10^{14}$ & 1.586$\times 10^{11}$ $(0.10)$
& 3.586$\times 10^{13}$ & 5.569$\times 10^{12}$ $(15.53)$ \\\hline
ThV & 1.185$\times 10^{15}$ & 1.725$\times 10^{14}$ & 1.872$\times 10^{11}$ $(0.11)$
& 4.218$\times 10^{13}$ & 6.015$\times 10^{12}$ $(14.26)$ \\\hline
ThT & 8.244$\times 10^{15}$ & 9.009$\times 10^{14}$ & 6.864$\times 10^{12}$ $(0.08)$
& 2.713$\times 10^{14}$ & 2.004$\times 10^{13}$ $(7.39)$ \\\hline
\end{tabular}
\end{threeparttable}
\end{table*}

\begin{table*}[thb]
\centering
\small
\begin{threeparttable}
\caption{
Ablation study 
in terms of average number of multiplications and average elapsed time 
until convergence in 1M-sized NYT with K=10\,000.\newline 
Number of iterations until convergence is 81.
}
\label{table:app_abl_nyt}
\centering
\begin{tabular}{|c|c|c||c|}\hline
\multirow{3}{*}{Algorithm} & Avg. \# multiplications & Avg. elapsed time & Maximum\\
 &  per iteration & per iteration (sec): & memory size\\
 &   & [assignment, update]\tnote{$\dagger$} & (GB) \\\hline
ES-ICP & 2.411$\times 10^{10}$ & 15.77 [5.394, 10.50]& 7.914 \\\hline
ES & 9.411$\times 10^{10}$ & 26.00 [15.72, 10.41]   & 7.907 \\\hline
ThV & 1.430$\times 10^{11}$ & 32.79 [21.99, 10.94]& 43.00 \\\hline
ThT & 2.385$\times 10^{12}$ & 238.5 [229.6, 9.016]& 4.752 \\\hline
\end{tabular}
\begin{tablenotes}
\item[$\dagger$]{\small The average elapsed time does not exactly match the sum of 
the assignment and the update time because the algorithm terminated at the end of 
the assignment step of the last iteration.}
\end{tablenotes}
\end{threeparttable}
\end{table*}

\begin{table*}[thb] 
\centering
\small
\begin{threeparttable}
\caption{
Ablation study in terms of perf results until convergence 
in 1M-sized NYT data set with K=10\,000.
Number of iterations until convergence is 81.
}
\label{table:app_perf_abl_nyt}
\centering
\begin{tabular}{|c|c|c|c|c|c|}\hline
\multirow{2}{*}{Algorithm} & \multirow{2}{*}{\# instructions} 
& \multirow{2}{*}{\# branches} & \# branch misses 
& \multirow{2}{*}{\# LLC-loads} & \# LLC-loads\\
& & & (\%) & & misses (\%)\\
\hline
ES-ICP & 7.003$\times 10^{13}$ & 1.239$\times 10^{13}$ & 4.127$\times 10^{10}$ $(0.33)$
& 8.470$\times 10^{11}$ & 1.044$\times 10^{11}$ $(12.3)$ \\\hline
ES & 1.538$\times 10^{14}$ & 1.993$\times 10^{13}$ & 7.044$\times 10^{10}$ $(0.35)$
& 3.046$\times 10^{12}$ & 2.727$\times 10^{11}$ $(8.95)$ \\\hline
ThV & 1.931$\times 10^{14}$ & 2.324$\times 10^{13}$ & 7.395$\times 10^{10}$ $(0.32)$
& 3.408$\times 10^{12}$ & 4.071$\times 10^{11}$ $(11.95)$ \\\hline
ThT & 2.017$\times 10^{15}$ & 2.091$\times 10^{14}$ & 8.155$\times 10^{10}$ $(0.04)$
& 2.508$\times 10^{13}$ & 2.215$\times 10^{12}$ $(8.83)$ \\\hline
\end{tabular}
\end{threeparttable}
\end{table*}

\captionsetup{skip=3pt}
\begin{table*}[ht]
\centering
\begin{threeparttable}
\small
\caption{
Performance of MIVI, DIVI, and Ding$^+$
in 8.2M-sized PubMed data set with K=80\,000.\hspace*{26mm}\newline
Number of iterations unitl convergence is 64.
}
\label{table:app_ivfd-ding}
\centering
\begin{tabular}{|c|c|c|}\hline
\multirow{2}{*}{Algorithm} & Avg. \# multiplications & Avg. elapsed time \\
 &  per iteration & per iteration (sec)\\\hline
MIVI & $1.326\times 10^{13}$ & $3.302\times 10^3$ \\\hline
DIVI & $1.326\times 10^{13}$ & $3.372\times 10^4$\\\hline
Ding$^+$ & $3.029\times 10^{12}$ & $9.552\times 10^3$ \\\hline
\end{tabular}
\end{threeparttable}
\end{table*}

\begin{table*}[ht]
\centering
\begin{threeparttable}
\small
\caption{
perf results of MIVI, DIVI, and Ding$^+$
in 8.2M-sized PubMed data set with K=80\,000.\newline
Number of iterations unitl convergence is 64.
}
\label{table:app_perf_ivfd-ding}
\centering
\begin{tabular}{|c|c|c|c|c|c|}\hline
\multirow{2}{*}{Algorithm} & \multirow{2}{*}{\# instructions} 
& \multirow{2}{*}{\# branches} & \# branch misses 
& \multirow{2}{*}{\# LLC-loads} & \# LLC-loads\\
& & & (\%) & & misses (\%)\\
\hline
MIVI & $1.024\times 10^{16}$ & $1.013\times 10^{15}$ & $3.929\times 10^{11}$ $(0.04)$
& $2.790\times 10^{14}$ & $1.767\times 10^{13}$ $(6.33)$ \\\hline
DIVI & $1.006\times 10^{16}$ & $1.019\times 10^{15}$ & $2.743\times 10^{12}$ $(0.27)$
& $7.983\times 10^{14}$ & $6.444\times 10^{14}$ $(80.7)$ \\\hline
Ding$^+$ & $6.691\times 10^{15}$ & $1.949\times 10^{15}$ & $1.937\times 10^{14}$ $(9.94)$
& $6.637\times 10^{14}$ & $6.577\times 10^{14}$ $(99.1)$ \\\hline
\end{tabular}
\end{threeparttable}
\end{table*}

\captionsetup{skip=3pt}
\begin{table*}[thb]
\centering
\small
\begin{threeparttable}
\caption{
Performance comparison of ES-ICP, ICP, CS-ICP and TA-ICP
in 8.2M-sized PubMed data set with K=80\,000.\hspace*{55mm}\newline 
Number of iterations until convergence is 64.
}
\label{table:app_comp}
\centering
\begin{tabular}{|c|c|c||c|}\hline
\multirow{3}{*}{Algorithm} & Avg. \# multiplications & Avg. elapsed time & Maximum\\
 &  per iteration & per iteration (sec): & memory size\\
 &   & [assignment, update]\tnote{$\dagger$} & (GB) \\\hline
ES-ICP & 9.391$\times 10^{10}$ & 204.8 [176.4, 28.86]& 16.72 \\\hline
ICP & 2.960$\times 10^{12}$ & 759.5 [729.9, 30.05]& 8.285 \\\hline
CS-ICP & 1.733$\times 10^{11}$ & 901.7 [875.2, 26.99]& 18.31 \\\hline
TA-ICP & 9.069$\times 10^{11}$ & 1042 [1006, 36.35]& 19.07 \\\hline
\end{tabular}
\begin{tablenotes}
\item[$\dagger$]{\small The average elapsed time does not exactly match the sum of 
the assignment and the update time because the algorithm terminated at the end of 
the assignment step of the last iteration.}
\end{tablenotes}
\end{threeparttable}
\end{table*}

\begin{table*}[thb]
\centering
\small
\begin{threeparttable}
\caption{
perf results of ES-ICP, ICP, CS-ICP, and TA-ICP 
in 8.2M-sized PubMed data set with K=80\,000.\newline
Number of iterations until convergence is 64.
}
\label{table:app_perf_comp}
\centering
\begin{tabular}{|c|c|c|c|c|c|}\hline
\multirow{2}{*}{Algorithm} & \multirow{2}{*}{\# instructions} 
& \multirow{2}{*}{\# branches} & \# branch misses 
& \multirow{2}{*}{\# LLC-loads} & \# LLC-loads\\
& & & (\%) & & misses (\%)\\
\hline
ES-ICP & 6.197$\times 10^{14}$ & 8.471$\times 10^{13}$ & 9.623$\times 10^{10}$ $(0.11)$
& 1.037$\times 10^{13}$ & 1.619$\times 10^{12}$ $(15.6)$ \\\hline
ICP & 2.876$\times 10^{15}$ & 2.934$\times 10^{14}$ & 2.796$\times 10^{11}$ $(0.10)$
& 8.277$\times 10^{13}$ & 4.467$\times 10^{12}$ $(5.40)$ \\\hline
CS-ICP & 2.346$\times 10^{15}$ & 2.786$\times 10^{14}$ & 3.127$\times 10^{11}$ $(0.11)$
& 9.297$\times 10^{13}$ & 8.025$\times 10^{12}$ $(8.63)$ \\\hline
TA-ICP & 1.476$\times 10^{15}$ & 2.067$\times 10^{14}$ & 1.859$\times 10^{12}$ $(0.90)$
& 4.593$\times 10^{13}$ & 2.209$\times 10^{13}$ $(48.10)$ \\\hline
\end{tabular}
\end{threeparttable}
\end{table*}

\begin{table*}[thb]
\centering
\small
\begin{threeparttable}
\caption{
Performance comparison of ES-ICP, ICP, CS-ICP and TA-ICP
in 1M-sized NYT data set with K=10\,000. \hspace*{53mm}\newline 
Number of iterations until convergence is 81.
}
\label{table:app_comp_nyt}
\centering
\begin{tabular}{|c|c|c||c|}\hline
\multirow{3}{*}{Algorithm} & Avg. \# multiplications & Avg. elapsed time & Maximum\\
 &  per iteration & per iteration (sec): & memory size\\
 &   & [assignment, update]\tnote{$\dagger$} & (GB) \\\hline
ES-ICP & 2.411$\times 10^{10}$ & 15.83 [5.466, 10.50]& 7.914 \\\hline
ICP & 3.947$\times 10^{11}$ & 68.13 [57.22, 11.05]   & 4.147 \\\hline
CS-ICP & 2.137$\times 10^{10}$ & 86.16 [76.89, 9.380]& 8.419 \\\hline
TA-ICP & 2.909$\times 10^{11}$ & 107.6 [94.44, 13.32]& 8.645 \\\hline
\end{tabular}
\begin{tablenotes}
\item[$\dagger$]{\small The average elapsed time does not exactly match the sum of 
the assignment and the update time because the algorithm terminated at the end of 
the assignment step of the last iteration.}
\end{tablenotes}
\end{threeparttable}
\end{table*}

\begin{table*}[thb]
\centering
\small
\begin{threeparttable}
\caption{
perf results of ES-ICP, ICP, CS-ICP, and TA-ICP 
in 1M-sized NYT data set with K=10\,000.\newline
Number of iterations until convergence is 81.
}
\label{table:app_perf_comp_nyt}
\centering
\begin{tabular}{|c|c|c|c|c|c|}\hline
\multirow{2}{*}{Algorithm} & \multirow{2}{*}{\# instructions} 
& \multirow{2}{*}{\# branches} & \# branch misses 
& \multirow{2}{*}{\# LLC-loads} & \# LLC-loads\\
& & & (\%) & & misses (\%)\\
\hline
ES-ICP & 7.041$\times 10^{13}$ & 1.246$\times 10^{13}$ & 4.094$\times 10^{10}$ $(0.33)$
& 8.340$\times 10^{11}$ & 1.051$\times 10^{11}$ $(12.6)$ \\\hline
ICP & 4.065$\times 10^{14}$ & 4.363$\times 10^{13}$ & 5.652$\times 10^{10}$ $(0.13)$
& 4.734$\times 10^{12}$ & 4.196$\times 10^{11}$ $(8.86)$ \\\hline
CS-ICP & 3.437$\times 10^{14}$ & 4.322$\times 10^{13}$ & 6.778$\times 10^{10}$ $(0.16)$
& 5.783$\times 10^{12}$ & 1.456$\times 10^{12}$ $(25.18)$ \\\hline
TA-ICP & 4.264$\times 10^{14}$ & 5.876$\times 10^{13}$ & 4.321$\times 10^{11}$ $(0.74)$
& 6.894$\times 10^{12}$ & 2.103$\times 10^{12}$ $(30.50)$ \\\hline
\end{tabular}
\end{threeparttable}
\end{table*}

\captionsetup{skip=3pt}
\begin{table*}[thb]
\centering
\small
\begin{threeparttable}
\caption{
Performance comparison of MIVI, ES-, CS-, and TA-MIVI
in 8.2M-sized\newline
PubMed data set with K=80\,000.\hspace*{48mm}\newline 
Number of iterations until convergence is 64.
}
\label{table:app_compubp}
\centering
\begin{tabular}{|c|c|c||c|}\hline
\multirow{3}{*}{Algorithm} & Avg. \# multiplications & Avg. elapsed time & Maximum\\
 &  per iteration & per iteration (sec): & memory size\\
 &   & [assignment, update]\tnote{$\dagger$} & (GB) \\\hline
MIVI & 1.326$\times 10^{13}$ & 3302 \,[3278, 23.81]& 8.251 \\\hline
ES-MIVI & 3.562$\times 10^{11}$ & \quad266.7 [237.8, 29.38]& 16.69 \\\hline
CS-MIVI & 7.601$\times 10^{11}$ & 2760 \,[2733, 27.53]& 18.28 \\\hline
TA-MIVI & 3.856$\times 10^{12}$ & 3380 \,[3342, 37.99]& 17.21 \\\hline
\end{tabular}
\begin{tablenotes}
\item[$\dagger$]{\small The average elapsed time does not exactly match the sum of 
the assignment and the update time because the algorithm terminated at the end of 
the assignment step of the last iteration.}
\end{tablenotes}
\end{threeparttable}
\end{table*}

\begin{table*}[thb]
\centering
\small
\begin{threeparttable}
\caption{
perf results of MIVI, ES-, CS-, and TA-MIVI
in 8.2M-sized PubMed data set with K=80\,000.\newline
Number of iterations until convergence is 64.
}
\label{table:app_perf_compubp}
\centering
\begin{tabular}{|c|c|c|c|c|c|}\hline
\multirow{2}{*}{Algorithm} & \multirow{2}{*}{\# instructions} 
& \multirow{2}{*}{\# branches} & \# branch misses 
& \multirow{2}{*}{\# LLC-loads} & \# LLC-loads\\
& & & (\%) & & misses (\%)\\
\hline
MIVI & 1.024$\times 10^{16}$ & 1.014$\times 10^{15}$ & 3.929$\times 10^{11}$ $(0.04)$
& 2.790$\times 10^{14}$ & 1.767$\times 10^{13}$ $(6.33)$ \\\hline
ES-MIVI & 1.062$\times 10^{15}$ & 1.677$\times 10^{14}$ & 1.842$\times 10^{11}$ $(0.11)$
& 3.592$\times 10^{13}$ & 4.522$\times 10^{12}$ $(12.6)$ \\\hline
CS-MIVI & 9.065$\times 10^{15}$ & 1.042$\times 10^{15}$ & 6.201$\times 10^{11}$ $(0.06)$
& 3.073$\times 10^{14}$ & 2.711$\times 10^{13}$ $(8.82)$ \\\hline
TA-MIVI & 4.824$\times 10^{15}$ & 6.895$\times 10^{14}$ & 7.428$\times 10^{12}$ $(1.08)$
& 1.754$\times 10^{14}$ & 6.626$\times 10^{13}$ $(37.77)$ \\\hline
\end{tabular}
\end{threeparttable}
\end{table*}

\begin{table*}[th]
\centering
\small
\begin{threeparttable}
\caption{
Performance comparison of MIVI, ES-, CS-, and TA-MIVI
in 1M-sized NYT data set with K=10\,000.\hspace*{67mm}\newline 
Number of iterations until convergence is 81.
}
\label{table:app_compubp_nyt}
\centering
\begin{tabular}{|c|c|c||c|}\hline
\multirow{3}{*}{Algorithm} & Avg. \# multiplications & Avg. elapsed time & Maximum\\
 &  per iteration & per iteration (sec): & memory size\\
 &   & [assignment, update]\tnote{$\dagger$} & (GB) \\\hline
MIVI & 1.955$\times 10^{12}$ & 272.4 [263.8, 8.723]   & 4.134 \\\hline
ES-MIVI & 9.411$\times 10^{10}$ & 26.06 [15.64, 10.55]& 7.907 \\\hline
CS-MIVI & 7.536$\times 10^{10}$ & 346.7 [337.2, 9.568]& 8.412 \\\hline
TA-MIVI & 1.367$\times 10^{12}$ & 280.6 [267.2, 13.56]& 8.030 \\\hline
\end{tabular}
\begin{tablenotes}
\item[$\dagger$]{\small The average elapsed time does not exactly match the sum of 
the assignment and the update time because the algorithm terminated at the end of 
the assignment step of the last iteration.}
\end{tablenotes}
\end{threeparttable}
\end{table*}

\begin{table*}[th]
\centering
\small
\begin{threeparttable}
\caption{
perf results of MIVI, ES-, CS-, and TA-MIVI
in 1M-sized NYT data set with K=10\,000.\newline
Number of iterations until convergence is 81.
}
\label{table:app_perf_compubp_nyt}
\centering
\begin{tabular}{|c|c|c|c|c|c|}\hline
\multirow{2}{*}{Algorithm} & \multirow{2}{*}{\# instructions} 
& \multirow{2}{*}{\# branches} & \# branch misses 
& \multirow{2}{*}{\# LLC-loads} & \# LLC-loads\\
& & & (\%) & & misses (\%)\\
\hline
MIVI & 1.804$\times 10^{15}$ & 1.716$\times 10^{14}$ & 7.717$\times 10^{10}$ $(0.04)$
& 2.720$\times 10^{13}$ & 2.077$\times 10^{12}$ $(7.64)$ \\\hline
ES-MIVI & 1.518$\times 10^{14}$ & 2.003$\times 10^{13}$ & 7.343$\times 10^{10}$ $(0.37)$
& 3.090$\times 10^{12}$ & 2.774$\times 10^{11}$ $(8.98)$ \\\hline
CS-MIVI & 1.540$\times 10^{15}$ & 1.766$\times 10^{14}$ & 1.287$\times 10^{11}$ $(0.07)$
& 2.907$\times 10^{13}$ & 6.011$\times 10^{12}$ $(20.68)$ \\\hline
TA-MIVI & 1.851$\times 10^{15}$ & 2.393$\times 10^{14}$ & 1.677$\times 10^{12}$ $(0.70)$
& 2.101$\times 10^{13}$ & 3.689$\times 10^{12}$ $(17.56)$ \\\hline
\end{tabular}
\end{threeparttable}
\end{table*}

\end{document}